\documentclass[10pt,twocolumn,letterpaper]{article}
\usepackage[T1]{fontenc}
\PassOptionsToPackage{table}{xcolor}
\usepackage[pagenumbers]{cvpr}
\usepackage{microtype}
\usepackage{mathtools}
\usepackage{multirow,tabularx,array}
\usepackage{float}
\usepackage{algorithm}
\usepackage{algpseudocode}
\usepackage{tikz}
\usetikzlibrary{arrows.meta,backgrounds,fit,positioning,shapes.geometric}
\usepackage{placeins}

\definecolor{T2IBlue}{HTML}{24557A}
\definecolor{T2ITeal}{HTML}{1F7A74}
\definecolor{T2IGold}{HTML}{C58B24}
\definecolor{T2IRed}{HTML}{A84448}
\definecolor{T2IGray}{HTML}{F1F4F6}
\definecolor{T2IDark}{HTML}{23313D}
\newcommand{\tTwoISpec}{\textsc{T2ISpec}\xspace}
\newcommand{\topicRoute}{\textsc{TopicRoute}\xspace}

\definecolor{cvprblue}{rgb}{0.21,0.49,0.74}
\usepackage[pagebackref,breaklinks,colorlinks,allcolors=cvprblue]{hyperref}

\title{\tTwoISpec: Benchmarking Speculative Decoding for Text-to-Image Generation}
\author{
Yifan Zhang$^{1,*}$\quad
Shatong Zhu$^{2,*}$\quad
Hui Shen$^{1,*}$\quad
Yunta Hsieh$^{3}$\\
Jianing Ma$^{6}$\quad
Muxin Fu$^{5}$\quad
Xin Wang$^{4}$\quad
Ping Zhang$^{4}$\quad
Qian Chen$^{6}$\quad
Chaofan Tao$^{6,\dagger}$\\[1.5ex]
{\normalsize $^{1}$University of Michigan -- Ann Arbor\quad
$^{2}$Stanford University\quad
$^{3}$University of Pennsylvania}\\
{\normalsize $^{4}$The Ohio State University\quad
$^{5}$Nanyang Technological University\quad
$^{6}$Independent}\\[1ex]
{\small $^{*}$Equal contribution\quad $^{\dagger}$Corresponding author}
}

\begin{document}
\maketitle
\begin{abstract}
Text-to-image (T2I) generators produce detailed images but remain slow at
inference. Speculative decoding can reduce this cost, yet its behavior
across the rendering requirements of T2I prompts remains insufficiently
understood. We introduce \tTwoISpec, a benchmark of 1,400 prompts balanced
across seven rendering topics and two difficulty levels. It evaluates
autoregressive and diffusion/flow accelerators by latency, work performed by
the target model, and image quality.
Our study reveals three key findings: (1) aggregate results hide large
differences between topics, (2) alignment and preference scores can disagree
with prompt constraint satisfaction, and (3) large reductions in target work
can yield limited latency gains. Motivated by these findings, we propose \topicRoute, a
plugin that combines prompt routing with speculative and cache controls
without retraining the generator. Across six model--method pairs, it achieves
a geometric mean speedup of $1.0308\times$ over the native methods, and the
speedup carries over to external prompts with the same settings, while quality
gains and losses vary across configurations.
Code and data are available at \url{https://github.com/zyfzs0/T2ISPEC}.
\end{abstract}

\begin{figure*}[t]
\centering
\includegraphics[width=\textwidth]{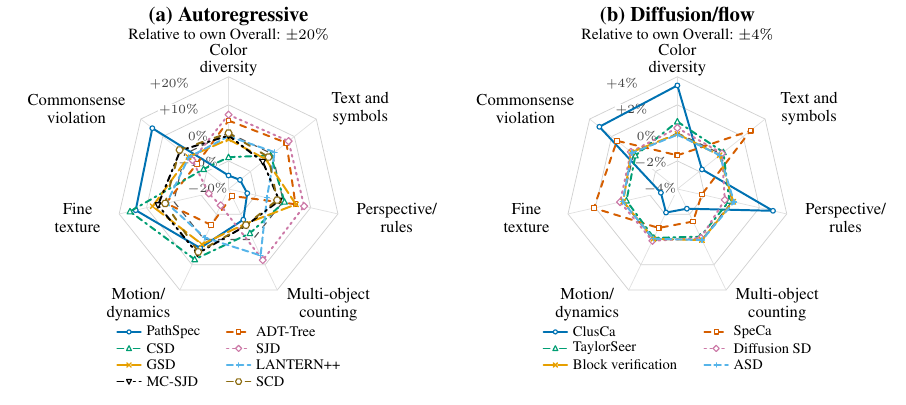}
\caption{Topic variation within each method: $100(S_g/S_{\mathrm{Overall}}-1)\%$,
where $S$ is the ratio of mean target/method times. The dashed zero line
marks each method's Overall ratio. Linear radial ranges are $\pm20\%$ for AR
and $\pm4\%$ for diffusion/flow. Appendix Table~\ref{tab:topic-spread-selected}
reports absolute ratios; settings follow Section~\ref{sec:setup}.}
\label{fig:topic-speed-radar}
\end{figure*}

\section{Introduction}
\label{sec:introduction}

Text-to-image (T2I) generators are slow at inference: autoregressive (AR)
models decode long sequences of visual tokens, and diffusion and flow models
evaluate large networks at every denoising step. Speculative decoding drafts
cheaply and verifies with the target model in parallel, and suitable
verification and correction preserve the target
distribution~\citep{leviathan2023fast,chen2023accelerating}. Many T2I
accelerators, however, accept approximate outputs and require explicit
quality evaluation.

Despite this progress, speculative decoding for T2I generation remains
insufficiently understood. Alignment and preference aggregates can conceal text and counting
failures, drafting, verification, and recovery costs can offset the savings
in target work, the forward evaluations of the target model, and prior works
differ in models, prompts, and timing boundaries.

To address this gap, we present \tTwoISpec, a benchmark with 1,400 prompts:
100 easy and 100 hard cases for each of seven rendering topics. It compares
the quality, latency, and target work of AR and diffusion accelerators using
fixed quality references and matched timing pairs.

Using \tTwoISpec, we obtain three key findings. First, aggregate results hide
differences between topics in speed (Figure~\ref{fig:topic-speed-radar}) and
quality: LANTERN on LlamaGen at $256^2$ improves the
Overall criterion by 0.0499 but loses 0.0933 on text and symbols. Second,
metrics can disagree: for the Lumina-mGPT target-only reference, counting
images rank second in mean ImageReward but second from last in criterion
score. Third, large reductions in target work can yield limited latency gains:
ADT-Tree cuts target work by 42.0\% but achieves only a 1.0889$\times$ speedup in mean
request time.

Based on these findings, we develop \topicRoute, a plugin for existing
accelerators that does not retrain the generator. A classifier predicts the
rendering topic from the prompt text and selects a frozen policy for the
backend, so prompts from different topics can receive different horizons or
cache settings; where supported, online cost feedback chooses horizons from
measured progress and cost. At inference, \topicRoute reads only the prompt
and runtime measurements.

Our contributions are:
\begin{itemize}
  \item \textbf{\tTwoISpec}, a benchmark balanced across seven rendering topics
  and two difficulty levels, with reproducible prompt identities and scene
  relationships.
  \item \textbf{A unified evaluation protocol} that reports latency, alignment,
  preference, and constraint satisfaction per topic, together with target
  work, within stated comparison groups.
  \item \textbf{A systematic empirical study} that identifies regressions in
  individual topics, disagreement between metrics, and gaps between work and
  latency.
  \item \textbf{\topicRoute}, a plugin informed by these findings, evaluated on
  six model--method pairs.
\end{itemize}

\section{Related Work}
\label{sec:related}

\subsection{Efficient T2I Models}

\paragraph{Autoregressive models.}
Autoregressive T2I models generate discrete visual tokens from
text. Parti uses a sequence-to-sequence
formulation~\citep{yu2022scaling}, and LlamaGen
applies the Llama architecture to image tokens~\citep{sun2024autoregressive}.
Lumina-mGPT supports flexible multimodal generation~\citep{liu2024lumina},
and Janus-Pro decouples visual encoding for understanding and
generation~\citep{chen2025janus}.

\paragraph{Diffusion and flow models.}
Diffusion models refine continuous representations; latent diffusion operates
in a compressed image space~\citep{rombach2022high}.
Flow models such as FLUX also apply successive latent
updates~\citep{flux2024}, each of which runs the full network. Training-free
feature reuse and forecasting, such as TaylorSeer~\citep{liu2025reusing},
SpeCa~\citep{liu2025speca}, and ClusCa~\citep{zheng2025compute}, reduce this
repeated computation.

\subsection{Speculative Decoding}

Speculative decoding verifies inexpensive proposals to advance several
generation units per iteration. Suitable discrete acceptance and correction
preserve the target distribution~\citep{leviathan2023fast,chen2023accelerating}.
Proposals come from learned auxiliary components, such as drafters that
propose static or adaptive draft trees~\citep{park2025lantern++,lei2025fast},
or from training-free procedures that use the existing model and its history,
such as SJD~\citep{teng2024accelerating} and SJD++, which retains confident
proposals across iterations~\citep{teng2025sjd++}.

LANTERN relaxes acceptance using the latent proximity of visual tokens, so
it no longer samples exactly from the target~\citep{jang2025lantern}.
Autospeculative decoding and diffusion speculative sampling instead propose
continuous states and preserve the output distribution of the target's
stochastic sampler~\citep{hu2025diffusion,de2025accelerated}. Feature
reuse and forecasting are related efficiency baselines with no such guarantee.

MMSpec connects multimodal workload evaluation to adaptive
drafting~\citep{shen2026mmspec};
\tTwoISpec applies this view to generated images, relating quality and
latency in each topic to its rendering requirements.

\section{The \tTwoISpec Benchmark}
\label{sec:benchmark}

\subsection{Dataset Construction}
\label{sec:dataset}

\begin{figure*}[t]
\centering
\begin{tikzpicture}[x=0.01\textwidth,y=-0.01\textwidth]
\node[anchor=north west,inner sep=0] at (0,0)
  {\includegraphics[width=\textwidth]{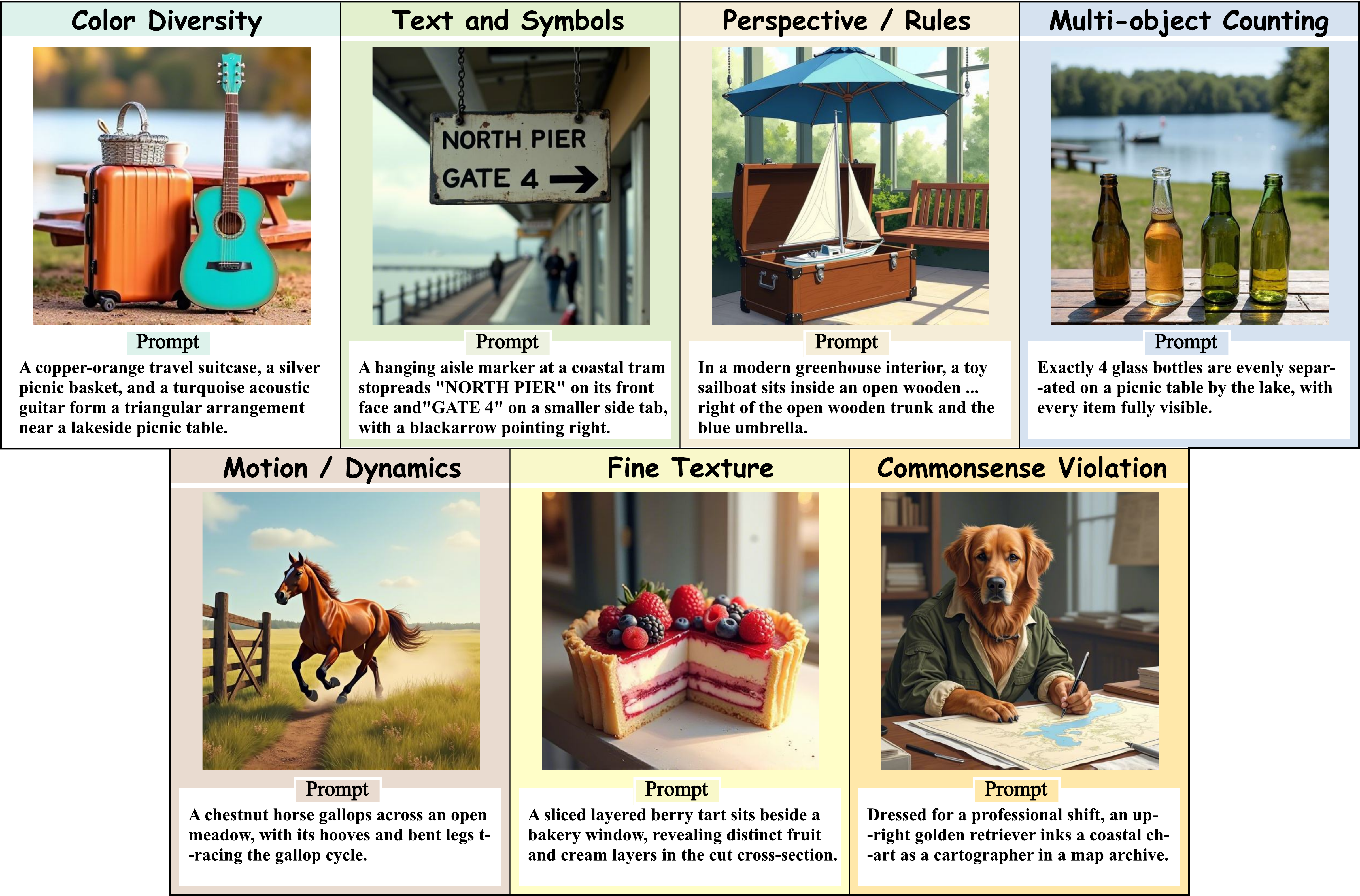}};
\foreach \left/\right in {0.6/24.4,25.6/49.4,50.6/74.4,75.6/99.4}
  \fill[white] (\left,26.35) rectangle (\right,32.6);
\foreach \left/\right in {13.3/36.8,38.3/61.8,63.3/86.8}
  \fill[white] (\left,59.2) rectangle (\right,65.2);
\tikzset{case prompt/.style={anchor=north west,inner sep=0,
  text width=0.225\textwidth,align=left,
  font=\fontsize{5.8}{6.3}\selectfont\bfseries}}
\node[case prompt] at (1.25,26.6) {A copper-orange travel suitcase, a silver
  picnic basket, and a turquoise acoustic guitar form a triangular arrangement
  near a lakeside picnic table.};
\node[case prompt] at (26.25,26.6) {A hanging aisle marker at a coastal tram
  stop reads ``NORTH PIER'' on its front face and ``GATE 4'' on a smaller side
  tab, with a black arrow pointing right.};
\node[case prompt] at (51.25,26.6) {\ldots\ a toy sailboat sits inside an
  open wooden trunk, a blue umbrella appears directly above the open wooden
  trunk, and a wooden bench stands behind and to the right \ldots};
\node[case prompt] at (76.25,26.6) {Exactly 4 glass bottles are evenly
  separated on a picnic table by the lake, with every item fully visible.};
\node[case prompt] at (13.75,59.45) {A chestnut horse gallops across an open
  meadow, with its hooves and bent legs tracing the gallop cycle.};
\node[case prompt] at (38.75,59.45) {A sliced layered berry tart sits beside
  a bakery window, revealing distinct fruit and cream layers in the cut
  cross-section.};
\node[case prompt] at (63.75,59.45) {Dressed for a professional shift, an
  upright golden retriever inks a coastal chart as a cartographer in a map
  archive.};
\end{tikzpicture}
\caption{Example prompts and generated images from the seven \tTwoISpec
topics. Prompt excerpts highlight the rendering requirement of each topic.
The images are selected from the better outputs generated in our
experiments.}
\label{fig:dataset-cases}
\end{figure*}
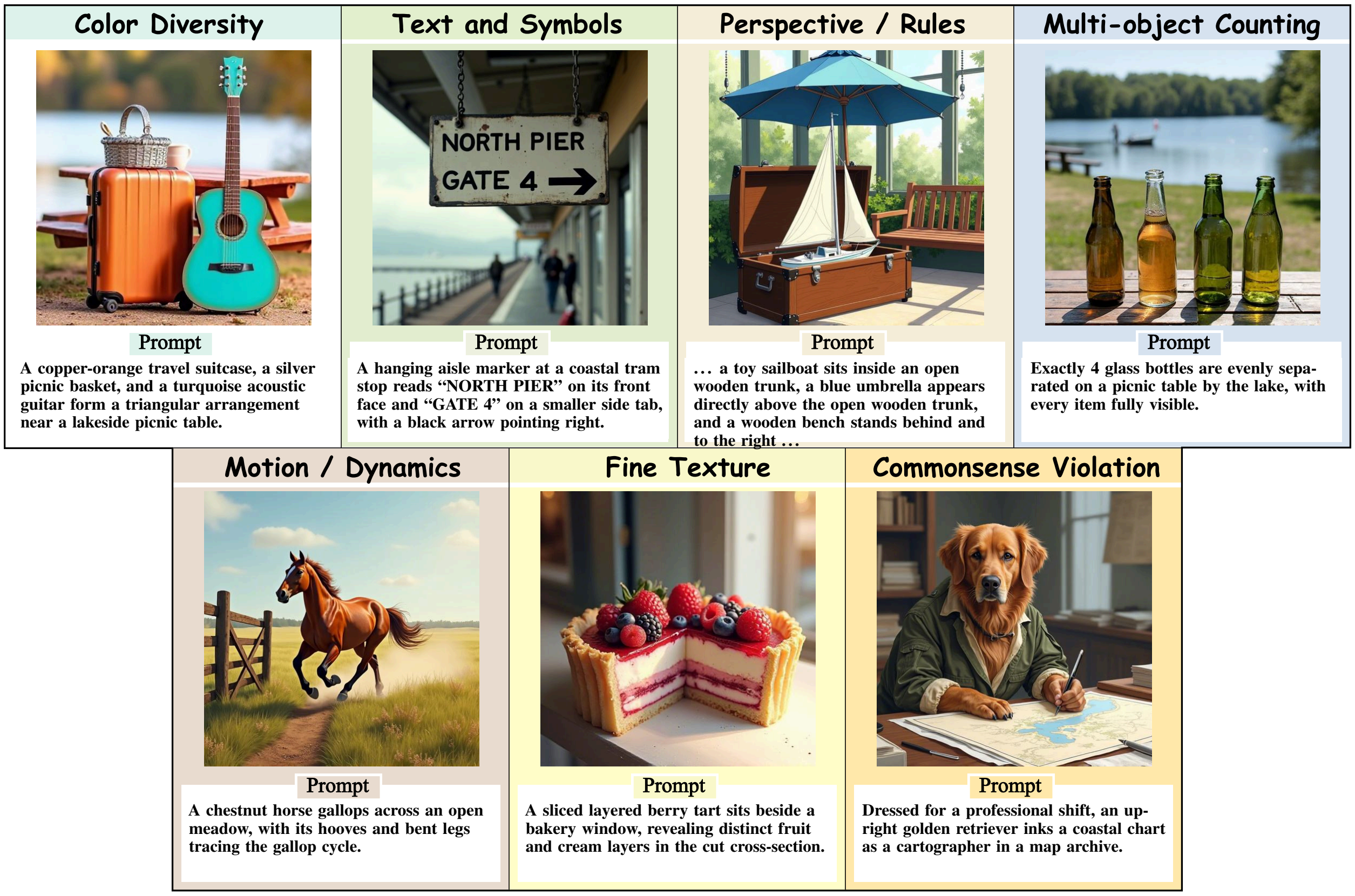

\paragraph{Source datasets and exploratory testing.}
We draw references from four source groups: \textbf{MS-COCO}~\citep{lin2014microsoft},
\textbf{PartiPrompts}~\citep{yu2022scaling},
\textbf{GenEval}~\citep{ghosh2023geneval},
and \textbf{T2I-CompBench/++}~\citep{huang2025t2i}; the two
CompBench versions form one group.

Exploratory tests of quality and efficiency for LANTERN and speculative
Jacobi decoding (SJD) across prompt content, together with the source
taxonomies, guide reference selection and motivate seven topics: \emph{color
diversity, text and symbols, perspective/rules, multi-object counting,
motion/dynamics, fine texture, and commonsense violation}.
Figure~\ref{fig:dataset-cases} illustrates each topic with a prompt excerpt
and a generated image.

\paragraph{Selection and scene synthesis.}
The reference pool contains 700 prompts, 100 per topic: 181 from MS-COCO,
169 from PartiPrompts, 139 from GenEval, and 211 from the CompBench group.
References serve only as guides for new scenes. A controlled
semantic composer varies objects, attributes, layouts, and interactions to
create 3,360 candidates. The selection step uses a fixed pretrained
sentence encoder to score semantic relevance and diversity, subject to
originality and deduplication constraints. The resulting 1,400 prompts specify
explicit visual requirements and keep links to their source references and
scene families.

\paragraph{Topic and difficulty balance.}
Each topic contains 100 easy and 100 hard prompts. Easy prompts
emphasize one primary rendering requirement; hard prompts combine interacting
constraints, such as exact counts under occlusion, local colors under
complex lighting, or text with orientation and layout requirements.
Section~\ref{sec:sensitivity} evaluates these construction labels.
All methods receive identical prompt text apart from the formatting each target requires.
Appendix~\ref{app:dataset-details} provides the selection rules, scene
relationships, prompt lengths, and examples.

\subsection{Speculative Decoding Algorithms}
\label{sec:algorithms}

\begin{figure*}[t]
\centering
\includegraphics[width=0.9\textwidth]{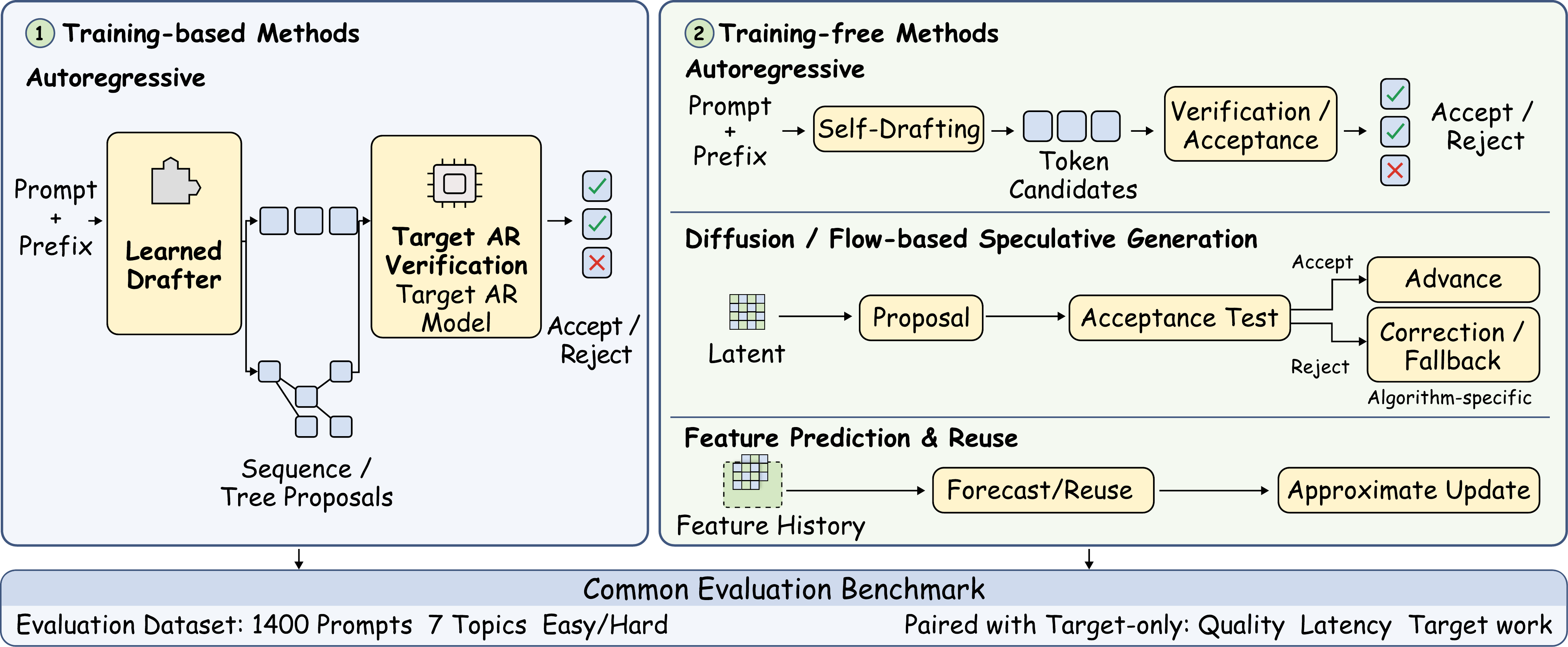}
\caption{Overview of acceleration mechanisms evaluated in \tTwoISpec.
Training categories refer to the evaluated implementations.}
\label{fig:overview}
\end{figure*}

\paragraph{Taxonomy.}
Acceleration is \emph{training-based} when it requires a learned auxiliary
drafter or predictor, and \emph{training-free} otherwise. The distinction
concerns accelerator training rather than target pretraining; an
implementation with a learned drafter remains training-based even when its
acceptance rule is training-free. Figure~\ref{fig:overview} summarizes learned
drafting, AR self-drafting, diffusion/flow speculative generation, and feature
prediction and reuse; appendix Table~\ref{tab:method-taxonomy} details each
method.

\paragraph{Evaluated methods.}
The training-based AR methods are LANTERN~\citep{jang2025lantern} and
LANTERN++~\citep{park2025lantern++} on LlamaGen, which share trained drafters
with the EAGLE architecture~\citep{li2024eagle}; ADT-Tree~\citep{lei2025fast} with a drafter
trained for Anole~\citep{chern2024anole} and LANTERN relaxation; and VVS~\citep{dong2026vvs} on
LlamaGen. The training-free AR methods are SJD~\citep{teng2024accelerating},
SJD-PAC~\citep{kang2026sjd}, grouped speculative decoding
(GSD)~\citep{so2025grouped}, speculative coupled decoding (SCD) with Gumbel
coupling and its maximal coupling configuration MC-SJD~\citep{so2025speculative},
content-aware speculative decoding (CSD)~\citep{wang2026csd}, and PathSpec, the
evaluated configuration from the PathRelax study~\citep{lei2026pathrelax}.
The evaluated ARsample filtering configuration is a sampling baseline guided
by entropy~\citep{ma2026towards}; it does not draft and verify. For diffusion and flow models, we
evaluate TaylorSeer~\citep{liu2025reusing}, SpeCa~\citep{liu2025speca}, and
ClusCa~\citep{zheng2025compute}, and adapt autospeculative decoding
(ASD)~\citep{hu2025diffusion}, diffusion speculative sampling (Diffusion
SD)~\citep{de2025accelerated}, and block
verification~\citep{sun2025block,soen2026accelerating} to FLUX.1-dev with
stochastic differential equation (SDE) samplers; Diffusion SD and block
verification are classified by whether their proposals require a learned
auxiliary model.

\section{Experiments}
\label{sec:baseline-results}

\subsection{Experiment Setup}
\label{sec:setup}
\label{sec:tracks}
\label{sec:measurements}

All benchmark experiments use NVIDIA H200 GPUs and equal batch sizes for
methods and target-only references within each comparison group.

\paragraph{Coverage and references.}
The core AR study covers 13 arms (four targets and nine methods) and 18,200
outputs on Janus-Pro-7B at $384^2$, LlamaGen Stage~I at $256^2$ and
Stage~II at $512^2$, and Lumina-mGPT at $768^2$. A further 15 paired
configurations cover nine AR and six diffusion methods, each with 1,400
target/method pairs (21,000 total).
Quality and speed are evaluated in eleven comparison groups, one per target
model and method category, with shared prompts, topics, and metrics. Every
method in a group uses the generation configuration of its target-only
reference (Appendix~\ref{app:evaluation-protocol}).
Ten parameter settings of five diffusion methods cover 14,000 pairs;
three of them also supply the SDE rows of the main comparisons.
FLUX.1-dev uses $512^2$ resolution, 50 target steps, and guidance 3.5.

In the main comparisons, all three SDE adaptations use $h=8$, the longest
evaluated horizon;
Section~\ref{sec:sensitivity} and appendix Table~\ref{tab:parameter-sweeps-overall}
report results for $h=2$. Each SDE row takes
its timing and method quality from the same run, and the three adaptations
use one quality target with a uniform schedule.
Appendix~\ref{app:latency-details} gives the timing details.

\paragraph{Quality and timing.}
The primary score $q_i$ of image $i$ averages three criterion scores in
$\{0,0.5,1\}$ that Qwen2.5-VL-7B-Instruct~\citep{Qwen2.5-VL} assigns to the explicit requirements
of its prompt and topic, including text and counting; the counting rubric covers exact count, count error, and group layout.
For a prompt group $I_g$, such as the 200 prompts
of a topic or all 1,400 prompts for Overall, $Q$ is the mean of $q_i$ over
$I_g$, $Q_2$ is the same mean for the scores of InternVL3-8B~\citep{zhu2025internvl3}, and $\Delta Q$
subtracts the target mean from the method mean.
CLIPScore~\citep{hessel2021clipscore}, ImageReward~\citep{xu2023imagereward},
and the human preference score (HPS)~\citep{wu2023human} measure
alignment and preference. All
comparisons on the 1,400 benchmark prompts use these five metrics plus an
aesthetic score, SigLIP alignment, and safety status;
Appendix~\ref{app:evaluation-protocol} gives the rubric,
evaluator settings, and data integrity checks.

The main tables report ratios of mean target/method times. In the
auxiliary analyses, core AR rows use median paired latency ratios, and the
other paired rows use ratios of mean target/method times:
\begin{equation}
 \begin{aligned}
 S_g^{\mathrm{core}}&=\operatorname{median}_{i\in I_g}
 \frac{t_i^{\mathrm{target}}}{t_i^{\mathrm{method}}},\\
 S_g^{\mathrm{paired}}&=\frac{\sum_{i\in I_g}t_i^{\mathrm{target}}}
 {\sum_{i\in I_g}t_i^{\mathrm{method}}}.
 \end{aligned}
 \label{eq:observed-ratio}
\end{equation}
Ratios above one indicate lower method latency. Overall ratios pool all
1,400 samples, and request latency, warm time after warmup runs, and
elapsed runner time are kept as separate timing scopes. For the
core AR rows, Appendix~\ref{app:resampling-details} gives 95\% bootstrap
intervals over prompt groups.

\subsection{Overall Comparison}
\label{sec:overall-comparison}

\paragraph{Autoregressive generation.}
Table~\ref{tab:overall-ar} reports AR results by topic and Overall.
On LlamaGen, LANTERN++ at $512^2$ records a latency ratio of
3.2163, and LANTERN at $256^2$ records 1.1063; their
criterion changes are $-0.0787$ and $+0.0499$, respectively.

\begin{table*}[!tp]
\centering
\caption{Autoregressive comparison across seven topics and Overall (1,400 prompts; 200/topic). $Q$ uses fixed quality references; $S$ divides mean target time by mean method time. Each group's target appears once. Black bold marks each method row's minimum/maximum topic speed. \textcolor{T2IBlue}{\textbf{Blue bold}}/\textcolor{T2IRed}{\underline{red underlined}} mark the highest and second highest Overall method values within each model block, respectively. Evaluation settings follow Section~\ref{sec:setup}.}
\label{tab:overall-ar}
\scriptsize
\setlength{\tabcolsep}{1.6pt}
\renewcommand{\arraystretch}{1.15}
\resizebox{\linewidth}{!}{%
\begin{tabular}{@{}ll*{8}{rr}@{}}
\toprule
\multirow{2}{*}{Model} & \multirow{2}{*}{Method} & \multicolumn{2}{c}{Color} & \multicolumn{2}{c}{Text} & \multicolumn{2}{c}{Persp.} & \multicolumn{2}{c}{Count} & \multicolumn{2}{c}{Motion} & \multicolumn{2}{c}{Texture} & \multicolumn{2}{c}{Common.} & \multicolumn{2}{c}{\textbf{Overall}} \\
\cmidrule(lr){3-4} \cmidrule(lr){5-6} \cmidrule(lr){7-8} \cmidrule(lr){9-10} \cmidrule(lr){11-12} \cmidrule(lr){13-14} \cmidrule(lr){15-16} \cmidrule(lr){17-18}
 & & $Q\uparrow$ & $S\uparrow$ & $Q\uparrow$ & $S\uparrow$ & $Q\uparrow$ & $S\uparrow$ & $Q\uparrow$ & $S\uparrow$ & $Q\uparrow$ & $S\uparrow$ & $Q\uparrow$ & $S\uparrow$ & $Q\uparrow$ & $S\uparrow$ & $Q\uparrow$ & $S\uparrow$ \\
\midrule
\multirow{4}{*}{\shortstack{Janus-Pro-7B\\$384^2$}} & \multicolumn{17}{l}{\cellcolor{T2IGray}\textit{\textbf{Training-free Methods}}} \\
 & Target-only & 0.994 & 1.0000$\times$ & 0.637 & 1.0000$\times$ & 0.990 & 1.0000$\times$ & 0.620 & 1.0000$\times$ & 0.745 & 1.0000$\times$ & 0.991 & 1.0000$\times$ & 0.893 & 1.0000$\times$ & 0.839 & 1.0000$\times$ \\
 & CSD & 0.992 & \textbf{1.5617$\times$} & 0.679 & 1.6966$\times$ & 0.993 & 1.7178$\times$ & 0.638 & 1.6677$\times$ & 0.764 & 1.8441$\times$ & 0.983 & \textbf{1.9846$\times$} & 0.910 & 1.5629$\times$ & \textcolor{T2IBlue}{\textbf{0.851}} & \textcolor{T2IBlue}{\textbf{1.7101$\times$}} \\
 & SJD & 0.925 & \textbf{1.3194$\times$} & 0.472 & 1.3643$\times$ & 0.954 & 1.3587$\times$ & 0.465 & 1.3239$\times$ & 0.559 & \textbf{1.4011$\times$} & 0.932 & 1.3654$\times$ & 0.795 & 1.3207$\times$ & \textcolor{T2IRed}{\underline{0.729}} & \textcolor{T2IRed}{\underline{1.3508$\times$}} \\
\midrule
\multirow{5}{*}{\shortstack{LlamaGen\\Stage I, $256^2$}} & \multicolumn{17}{l}{\cellcolor{T2IGray}\textit{\textbf{Training-based Methods}}} \\
 & Target-only & 0.837 & 1.0000$\times$ & 0.544 & 1.0000$\times$ & 0.685 & 1.0000$\times$ & 0.353 & 1.0000$\times$ & 0.471 & 1.0000$\times$ & 0.730 & 1.0000$\times$ & 0.578 & 1.0000$\times$ & 0.600 & 1.0000$\times$ \\
 & LANTERN & 0.870 & 1.0789$\times$ & 0.451 & \textbf{1.1719$\times$} & 0.892 & \textbf{1.0630$\times$} & 0.354 & 1.1540$\times$ & 0.564 & 1.0880$\times$ & 0.809 & 1.1223$\times$ & 0.608 & 1.0692$\times$ & \textcolor{T2IRed}{\underline{0.650}} & \textcolor{T2IRed}{\underline{1.1063$\times$}} \\
 & LANTERN++ & 0.704 & 4.4682$\times$ & 0.510 & 4.5694$\times$ & 0.833 & \textbf{4.2519$\times$} & 0.273 & \textbf{4.8069$\times$} & 0.427 & 4.4900$\times$ & 0.808 & 4.6403$\times$ & 0.528 & 4.4377$\times$ & 0.584 & \textcolor{T2IBlue}{\textbf{4.5219$\times$}} \\
 & VVS & 0.851 & \textbf{0.9686$\times$} & 0.529 & 1.0441$\times$ & 0.879 & 1.0460$\times$ & 0.388 & \textbf{1.0573$\times$} & 0.592 & 1.0391$\times$ & 0.817 & 1.0414$\times$ & 0.603 & 1.0474$\times$ & \textcolor{T2IBlue}{\textbf{0.665}} & 1.0337$\times$ \\
\midrule
\multirow{8}{*}{\shortstack{LlamaGen\\Stage II, $512^2$}} & \multicolumn{17}{l}{\cellcolor{T2IGray}\textit{\textbf{Training-based Methods}}} \\
 & Target-only & 0.940 & 1.0000$\times$ & 0.300 & 1.0000$\times$ & 0.963 & 1.0000$\times$ & 0.585 & 1.0000$\times$ & 0.652 & 1.0000$\times$ & 0.965 & 1.0000$\times$ & 0.808 & 1.0000$\times$ & 0.745 & 1.0000$\times$ \\
 & LANTERN & 0.932 & \textbf{0.8729$\times$} & 0.219 & 0.9437$\times$ & 0.959 & 0.8823$\times$ & 0.563 & 0.8843$\times$ & 0.764 & \textbf{0.9474$\times$} & 0.974 & 0.9043$\times$ & 0.788 & 0.9050$\times$ & \textcolor{T2IBlue}{\textbf{0.743}} & 0.9052$\times$ \\
 & LANTERN++ & 0.865 & 3.1827$\times$ & 0.339 & \textbf{3.4168$\times$} & 0.936 & 3.2569$\times$ & 0.369 & 3.1763$\times$ & 0.631 & \textbf{3.0980$\times$} & 0.808 & 3.2127$\times$ & 0.714 & 3.1889$\times$ & 0.666 & \textcolor{T2IBlue}{\textbf{3.2163$\times$}} \\
 & \multicolumn{17}{l}{\cellcolor{T2IGray}\textit{\textbf{Training-free Methods}}} \\
 & Target-only & 0.908 & 1.0000$\times$ & 0.220 & 1.0000$\times$ & 0.943 & 1.0000$\times$ & 0.552 & 1.0000$\times$ & 0.762 & 1.0000$\times$ & 0.979 & 1.0000$\times$ & 0.794 & 1.0000$\times$ & 0.737 & 1.0000$\times$ \\
 & ARsample filtering & 0.918 & 0.9936$\times$ & 0.243 & 0.9939$\times$ & 0.962 & 0.9954$\times$ & 0.537 & \textbf{0.9925$\times$} & 0.753 & 0.9932$\times$ & 0.975 & 0.9942$\times$ & 0.770 & \textbf{1.0018$\times$} & \textcolor{T2IRed}{\underline{0.737}} & 0.9949$\times$ \\
 & SJD & 0.943 & 1.5743$\times$ & 0.254 & 1.5920$\times$ & 0.951 & 1.5781$\times$ & 0.617 & 1.5231$\times$ & 0.638 & 1.5527$\times$ & 0.945 & \textbf{1.6435$\times$} & 0.805 & \textbf{1.5212$\times$} & 0.736 & \textcolor{T2IRed}{\underline{1.5682$\times$}} \\
\midrule
\multirow{8}{*}{\shortstack{Lumina-mGPT-7B\\$768^2$}} & \multicolumn{17}{l}{\cellcolor{T2IGray}\textit{\textbf{Training-free Methods}}} \\
 & Target-only & 0.976 & 1.0000$\times$ & 0.233 & 1.0000$\times$ & 0.986 & 1.0000$\times$ & 0.547 & 1.0000$\times$ & 0.760 & 1.0000$\times$ & 0.989 & 1.0000$\times$ & 0.853 & 1.0000$\times$ & 0.763 & 1.0000$\times$ \\
 & GSD & 0.971 & 4.9045$\times$ & 0.212 & 4.8684$\times$ & 0.983 & 5.2521$\times$ & 0.517 & \textbf{4.7079$\times$} & 0.831 & 5.1133$\times$ & 0.994 & \textbf{5.4098$\times$} & 0.833 & 4.9343$\times$ & \textcolor{T2IRed}{\underline{0.763}} & \textcolor{T2IRed}{\underline{5.0195$\times$}} \\
 & MC-SJD & 0.983 & 4.6784$\times$ & 0.267 & 4.5152$\times$ & 0.965 & 4.6543$\times$ & 0.541 & \textbf{4.4612$\times$} & 0.791 & 4.9853$\times$ & 0.994 & \textbf{4.9982$\times$} & 0.872 & 4.8180$\times$ & \textcolor{T2IBlue}{\textbf{0.773}} & 4.7234$\times$ \\
 & SCD & 0.968 & 5.2877$\times$ & 0.250 & 5.1965$\times$ & 0.973 & 5.1719$\times$ & 0.479 & \textbf{4.9875$\times$} & 0.787 & \textbf{5.5490$\times$} & 0.994 & 5.4593$\times$ & 0.823 & 5.4135$\times$ & 0.753 & \textcolor{T2IBlue}{\textbf{5.2901$\times$}} \\
 & SJD & 0.963 & 3.5779$\times$ & 0.247 & 3.5406$\times$ & 0.970 & \textbf{3.7742$\times$} & 0.551 & \textbf{3.4338$\times$} & 0.759 & 3.6028$\times$ & 0.991 & 3.6373$\times$ & 0.833 & 3.6093$\times$ & 0.759 & 3.5953$\times$ \\
 & PathSpec & 0.968 & \textbf{2.8283$\times$} & 0.243 & 2.8415$\times$ & 0.959 & 2.8957$\times$ & 0.506 & 3.0729$\times$ & 0.797 & 3.4465$\times$ & 0.993 & 3.8002$\times$ & 0.839 & \textbf{3.8270$\times$} & 0.758 & 3.3348$\times$ \\
 & SJD-PAC & 0.967 & \textbf{3.0006$\times$} & 0.224 & 3.2114$\times$ & 0.973 & 3.2752$\times$ & 0.538 & 3.1647$\times$ & 0.776 & 3.2062$\times$ & 0.984 & \textbf{3.3030$\times$} & 0.858 & 3.2057$\times$ & 0.760 & 3.2031$\times$ \\
\midrule
\multirow{6}{*}{\shortstack{Anole-7B}} & \multicolumn{17}{l}{\cellcolor{T2IGray}\textit{\textbf{Training-based Methods}}} \\
 & Target-only & 0.958 & 1.0000$\times$ & 0.203 & 1.0000$\times$ & 0.947 & 1.0000$\times$ & 0.403 & 1.0000$\times$ & 0.552 & 1.0000$\times$ & 0.960 & 1.0000$\times$ & 0.686 & 1.0000$\times$ & 0.672 & 1.0000$\times$ \\
 & ADT-Tree + LANTERN & 0.958 & 1.1368$\times$ & 0.221 & \textbf{1.1585$\times$} & 0.949 & 1.1375$\times$ & 0.353 & \textbf{0.9011$\times$} & 0.649 & 1.0264$\times$ & 0.962 & 1.1202$\times$ & 0.682 & 1.0284$\times$ & \textcolor{T2IBlue}{\textbf{0.682}} & \textcolor{T2IBlue}{\textbf{1.0889$\times$}} \\
 & \multicolumn{17}{l}{\cellcolor{T2IGray}\textit{\textbf{Training-free Methods}}} \\
 & Target-only & 0.953 & 1.0000$\times$ & 0.164 & 1.0000$\times$ & 0.818 & 1.0000$\times$ & 0.482 & 1.0000$\times$ & 0.592 & 1.0000$\times$ & 0.940 & 1.0000$\times$ & 0.634 & 1.0000$\times$ & 0.655 & 1.0000$\times$ \\
 & SJD & 0.940 & 1.0630$\times$ & 0.175 & 1.0718$\times$ & 0.820 & 1.0745$\times$ & 0.487 & \textbf{1.0794$\times$} & 0.613 & \textbf{0.8633$\times$} & 0.901 & 0.8708$\times$ & 0.652 & 0.9631$\times$ & \textcolor{T2IRed}{\underline{0.655}} & \textcolor{T2IRed}{\underline{0.9983$\times$}} \\
\midrule
\multirow{3}{*}{\shortstack{Emu3}} & \multicolumn{17}{l}{\cellcolor{T2IGray}\textit{\textbf{Training-free Methods}}} \\
 & Target-only & 0.943 & 1.0000$\times$ & 0.275 & 1.0000$\times$ & 0.922 & 1.0000$\times$ & 0.546 & 1.0000$\times$ & 0.662 & 1.0000$\times$ & 0.958 & 1.0000$\times$ & 0.853 & 1.0000$\times$ & 0.737 & 1.0000$\times$ \\
 & SJD & 0.980 & 1.0161$\times$ & 0.275 & 1.0035$\times$ & 0.912 & 1.0478$\times$ & 0.540 & \textbf{1.0531$\times$} & 0.639 & \textbf{0.9539$\times$} & 0.978 & 1.0500$\times$ & 0.866 & 1.0497$\times$ & \textcolor{T2IBlue}{\textbf{0.741}} & \textcolor{T2IBlue}{\textbf{1.0249$\times$}} \\
\bottomrule
\end{tabular}%
}
\end{table*}

Speed also varies within methods: across topics, PathSpec spans
$2.8283$--$3.8270\times$ (35.31\% relative spread), and CSD spans
$1.5617$--$1.9846\times$ (27.08\%). ADT-Tree (latency ratio
$1.0889$) changes the Overall criterion by $+0.0096$, and ARsample's Overall
criterion change (0.0000, latency ratio $0.9949$) masks a text gain of
0.0233 and a commonsense loss of 0.0242.

\paragraph{Diffusion generation.}
Table~\ref{tab:overall-diffusion} reports six diffusion/flow configurations.
SpeCa and ClusCa achieve speedups of $4.6825\times$ and $3.5739\times$ on
FLUX.1-dev, respectively, and TaylorSeer achieves $1.6149\times$ on
FLUX.1-schnell. For the three SDE adaptations at $h=8$, target/method latency
ratios of $0.1460$--$0.2522$ indicate slower execution than their matched
timing references.

\begin{table*}[!tp]
\centering
\caption{Diffusion/flow comparison across seven topics and Overall (1,400 prompts; 200/topic). $Q$ uses fixed quality references; $S$ divides mean target time by mean method time. Each group's target appears once. Black bold marks each method row's minimum/maximum topic speed. \textcolor{T2IBlue}{\textbf{Blue bold}}/\textcolor{T2IRed}{\underline{red underlined}} mark the highest and second highest Overall method values within each model block, respectively. Evaluation settings follow Section~\ref{sec:setup}.}
\label{tab:overall-diffusion}
\scriptsize
\setlength{\tabcolsep}{1.6pt}
\renewcommand{\arraystretch}{1.15}
\resizebox{\linewidth}{!}{%
\begin{tabular}{@{}ll*{8}{rr}@{}}
\toprule
\multirow{2}{*}{Model} & \multirow{2}{*}{Method} & \multicolumn{2}{c}{Color} & \multicolumn{2}{c}{Text} & \multicolumn{2}{c}{Persp.} & \multicolumn{2}{c}{Count} & \multicolumn{2}{c}{Motion} & \multicolumn{2}{c}{Texture} & \multicolumn{2}{c}{Common.} & \multicolumn{2}{c}{\textbf{Overall}} \\
\cmidrule(lr){3-4} \cmidrule(lr){5-6} \cmidrule(lr){7-8} \cmidrule(lr){9-10} \cmidrule(lr){11-12} \cmidrule(lr){13-14} \cmidrule(lr){15-16} \cmidrule(lr){17-18}
 & & $Q\uparrow$ & $S\uparrow$ & $Q\uparrow$ & $S\uparrow$ & $Q\uparrow$ & $S\uparrow$ & $Q\uparrow$ & $S\uparrow$ & $Q\uparrow$ & $S\uparrow$ & $Q\uparrow$ & $S\uparrow$ & $Q\uparrow$ & $S\uparrow$ & $Q\uparrow$ & $S\uparrow$ \\
\midrule
\multirow{3}{*}{\shortstack{FLUX.1-schnell}} & \multicolumn{17}{l}{\cellcolor{T2IGray}\textit{\textbf{Training-free Methods}}} \\
 & Target-only & 0.997 & 1.0000$\times$ & 0.847 & 1.0000$\times$ & 0.985 & 1.0000$\times$ & 0.827 & 1.0000$\times$ & 0.661 & 1.0000$\times$ & 0.985 & 1.0000$\times$ & 0.945 & 1.0000$\times$ & 0.892 & 1.0000$\times$ \\
 & TaylorSeer & 0.995 & \textbf{1.6279$\times$} & 0.838 & 1.6181$\times$ & 0.992 & 1.6117$\times$ & 0.816 & 1.6112$\times$ & 0.759 & 1.6128$\times$ & 0.995 & \textbf{1.6105$\times$} & 0.917 & 1.6121$\times$ & \textcolor{T2IBlue}{\textbf{0.902}} & \textcolor{T2IBlue}{\textbf{1.6149$\times$}} \\
\midrule
\multirow{9}{*}{\shortstack{FLUX.1-dev}} & \multicolumn{17}{l}{\cellcolor{T2IGray}\textit{\textbf{Training-free Methods}}} \\
 & Target-only & 0.993 & 1.0000$\times$ & 0.779 & 1.0000$\times$ & 0.996 & 1.0000$\times$ & 0.757 & 1.0000$\times$ & 0.703 & 1.0000$\times$ & 0.991 & 1.0000$\times$ & 0.907 & 1.0000$\times$ & 0.875 & 1.0000$\times$ \\
 & ClusCa & 0.993 & \textbf{3.6948$\times$} & 0.748 & 3.5110$\times$ & 0.995 & 3.6808$\times$ & 0.741 & 3.4866$\times$ & 0.740 & 3.4975$\times$ & 0.990 & \textbf{3.4741$\times$} & 0.917 & 3.6852$\times$ & \textcolor{T2IRed}{\underline{0.875}} & \textcolor{T2IRed}{\underline{3.5739$\times$}} \\
 & SpeCa & 0.993 & 4.6075$\times$ & 0.772 & \textbf{4.8079$\times$} & 0.992 & \textbf{4.5786$\times$} & 0.758 & 4.6157$\times$ & 0.734 & 4.6395$\times$ & 0.987 & 4.7813$\times$ & 0.915 & 4.7533$\times$ & \textcolor{T2IBlue}{\textbf{0.879}} & \textcolor{T2IBlue}{\textbf{4.6825$\times$}} \\
 & \multicolumn{17}{l}{\cellcolor{T2IGray}\textit{\textbf{SDE adaptations}}} \\
 & Target-only & 0.036 & 1.0000$\times$ & 0.000 & 1.0000$\times$ & 0.022 & 1.0000$\times$ & 0.122 & 1.0000$\times$ & 0.016 & 1.0000$\times$ & 0.463 & 1.0000$\times$ & 0.118 & 1.0000$\times$ & 0.111 & 1.0000$\times$ \\
 & ASD & 0.035 & \textbf{0.1458$\times$} & 0.000 & 0.1458$\times$ & 0.023 & \textbf{0.1462$\times$} & 0.095 & 0.1461$\times$ & 0.014 & 0.1459$\times$ & 0.437 & 0.1460$\times$ & 0.118 & 0.1462$\times$ & 0.103 & 0.1460$\times$ \\
 & Block verification & 0.029 & 0.2520$\times$ & 0.000 & \textbf{0.2518$\times$} & 0.022 & 0.2524$\times$ & 0.087 & 0.2523$\times$ & 0.009 & 0.2520$\times$ & 0.443 & 0.2521$\times$ & 0.118 & \textbf{0.2527$\times$} & 0.101 & 0.2522$\times$ \\
 & Diffusion SD & 0.032 & \textbf{0.1480$\times$} & 0.000 & 0.1476$\times$ & 0.025 & \textbf{0.1467$\times$} & 0.100 & 0.1473$\times$ & 0.012 & 0.1476$\times$ & 0.472 & 0.1478$\times$ & 0.118 & 0.1478$\times$ & 0.108 & 0.1475$\times$ \\
\bottomrule
\end{tabular}%
}
\end{table*}

\paragraph{Quality metrics expose different tradeoffs.}
SJD-PAC increases ImageReward by 0.0368 but reduces CLIPScore by 0.0056, with
a criterion change of only $-0.0033$. SpeCa leaves the Overall criterion
nearly unchanged but lowers HPS. Preference, alignment, and explicit
constraint satisfaction are therefore not interchangeable.
Appendix~\ref{app:topic-details} reports all eight quality metrics.

\subsection{Sensitivity Study}
\label{sec:sensitivity}

\paragraph{Overall can hide regressions in individual topics.}
Counting has one of the two lowest
criterion scores for every core target, and text scores below each target's
Overall score. Acceleration changes these profiles: LANTERN at $256^2$ gains 0.0499
Overall and 0.2067 on perspective yet loses 0.0933 on text. The interval of
its text change, $[-0.1625,-0.0241]$, excludes zero despite the aggregate gain. The second evaluator reproduces this reversal ($\Delta Q_2=+0.0354$ Overall and $-0.0300$ on text).
LANTERN's mean CLIPScore, ImageReward, and HPS nevertheless rise in all seven
topics, including text.

Diffusion shows similar cancellation. ClusCa's Overall criterion change
($-0.0002$) is close to zero, combining losses of 0.0308 on text and 0.0158 on
counting with gains of 0.0375 on motion and 0.0100 on commonsense violation. SpeCa
gains only 0.0037 Overall, combining a motion gain of 0.0317 with losses of
0.0067 on text and 0.0042 each on texture and perspective. TaylorSeer's Overall
gain of 0.0093 combines a motion gain of 0.0983 with a commonsense loss of
0.0283.
Appendix~\ref{app:qualitative-comparison} shows examples across the seven topics.

\paragraph{Easy/Hard effects depend on the configuration.}
The Easy/Hard ordering varies by target
(Appendix~\ref{app:resampling-details}). The LlamaGen $256^2$ target-only
reference scores 0.0390 lower on Hard, whereas Janus and Lumina score higher
on Hard than on Easy. The labels thus mark construction constraints, such as
prompt length and scene composition, whose effect depends on the target.

ARsample changes the criterion by $+0.0129$ on Easy and $-0.0129$ on
Hard at the same latency ratio (0.969). VVS has nearly equal latency ratios
on Easy and Hard (1.030 and 1.037) but a larger criterion gain on Hard
($+0.0738$ versus $+0.0576$).

\paragraph{Longer horizons can increase work without improving quality.}
Increasing the horizon from 2 to 8 lowers the latency ratios of
diffusion speculative sampling, ASD, and block verification from 0.534,
0.531, and 0.684 to 0.147, 0.146, and 0.252, respectively, without improving
the Overall criterion.
For diffusion speculative sampling, the mean number of target drift
evaluations per image rises from 96 to 364, accepted units per image fall
from 1.5 to 1.1, and discarded units rise from 46 to 314.

\paragraph{Aggressive caching concentrates losses in individual topics.}
Raising SpeCa's base threshold from 5 to 9 increases the latency ratio
from 4.284 to 5.134, while the Overall criterion change falls from
$+0.0043$ to $-0.0095$. The text change worsens far more, from $-0.0167$
to $-0.0717$, and the ImageReward change falls from $+0.0195$ to $-0.0450$.
For ClusCa, freshness thresholds 3 and 7 yield ratios of 2.852 and 4.103,
Overall criterion changes of $+0.0052$ and $-0.0052$, and text changes
of $+0.0383$ and $-0.0517$; at threshold 7, Easy loses 0.0098 while Hard
is nearly unchanged ($-0.0007$).
CLIPScore increases at all four settings, so it alone would not reveal these
costs. Adaptive reuse therefore requires topic and difficulty analysis
alongside speed. Appendix~\ref{app:parameter-sweeps} reports all ten
evaluations against their targets, with eight metrics and topic
breakdowns.

\subsection{Latency Analysis}
\label{sec:latency}

\paragraph{Target work and latency gains can diverge.}
ADT-Tree lowers mean target evaluations from 1,024 to 594.13 (42.0\%),
but mean request time falls only from 30.846 to 28.327 seconds, a speedup of
$1.0889\times$, with lower latency than the paired target-only reference in
912 of 1,400 requests. Proposal, verification, and coordination costs can offset
the saved target work. In the core AR measurements, the interval spent generating
visual tokens accounts for 97.1--99.8\% of mean warm time, so the main
opportunity lies in token generation. Component timers overlap and are not
additive; Appendix~\ref{app:latency-details} reports their measurements and
work counts.

PathSpec outputs contain 2,357--2,506 units and match the target length of
2,357 units in six of 1,400 pairs, so its ratio compares requests of
different lengths.

\paragraph{Aggregation and timing boundaries matter.}
In the core AR study, LANTERN on LlamaGen $512^2$ has a median
paired ratio of 1.001 even though its mean latency increases from 17.456 to
19.283 seconds; a median ratio near one can mask a higher mean
latency. Appendix~\ref{app:latency-details} reports absolute times and
distinguishes warm, request, and runner timing.

\section{\topicRoute: Prompt and Cost Aware Speculative Control}
\label{sec:implications}
\label{sec:plugin}

Motivated by the findings in Section~\ref{sec:baseline-results}, we design
\topicRoute, a lightweight controller
that selects a policy for each prompt and applies native horizon or cache
controls, using online cost feedback where supported.

\begin{table*}[t]
\centering
\caption{\topicRoute on 1,400 prompts per group. Each native method is followed by the same method with \topicRoute (shaded); bold marks the better value of each pair. $S$ is the speedup over the target-only model and $D$ the topic speed disparity. $W$ counts topics faster than native, $K$ counts nondecreasing means among eight quality metrics, and Pass requires all five group criteria. IR denotes ImageReward.}
\label{tab:plugin-main}
\begingroup
\setlength{\tabcolsep}{4.5pt}
\renewcommand{\arraystretch}{1.06}
\footnotesize
\begin{tabular}{llrrrrrrccc}
\toprule
 & & \multicolumn{2}{c}{Speed} & \multicolumn{4}{c}{Quality} & \multicolumn{3}{c}{Group criteria} \\
\cmidrule(lr){3-4}\cmidrule(lr){5-8}\cmidrule(lr){9-11}
Model & Method & $S\uparrow$ & $D\downarrow$ & $Q\uparrow$ & CLIP$\uparrow$ & IR$\uparrow$ & HPS$\uparrow$ & $W$/7 & $K$/8 & Pass \\
\midrule
\multirow{2}{*}{Janus-Pro-7B $384^2$} & CSD & 2.3786 & 0.2325 & 0.8396 & 0.7310 & \textbf{0.6461} & 0.2509 &  &  &  \\
 & \cellcolor{T2IGray}w/ \topicRoute & \cellcolor{T2IGray}\textbf{2.4106} & \cellcolor{T2IGray}\textbf{0.2177} & \cellcolor{T2IGray}\textbf{0.8421} & \cellcolor{T2IGray}\textbf{0.7315} & \cellcolor{T2IGray}0.6417 & \cellcolor{T2IGray}\textbf{0.2510} & \cellcolor{T2IGray}6 & \cellcolor{T2IGray}6 & \cellcolor{T2IGray}Yes \\
\midrule
\multirow{2}{*}{LlamaGen Stage I $256^2$} & LANTERN & 0.9444 & 0.0659 & \textbf{0.6489} & \textbf{0.5202} & -1.6908 & \textbf{0.1544} &  &  &  \\
 & \cellcolor{T2IGray}w/ \topicRoute & \cellcolor{T2IGray}\textbf{1.0143} & \cellcolor{T2IGray}\textbf{0.0226} & \cellcolor{T2IGray}0.6426 & \cellcolor{T2IGray}0.5197 & \cellcolor{T2IGray}\textbf{-1.6826} & \cellcolor{T2IGray}0.1543 & \cellcolor{T2IGray}7 & \cellcolor{T2IGray}5 & \cellcolor{T2IGray}Yes \\
\midrule
\multirow{2}{*}{Lumina-mGPT $768^2$} & SJD & \textbf{1.7961} & \textbf{0.0311} & 0.7551 & \textbf{0.6879} & \textbf{0.2743} & \textbf{0.2698} &  &  &  \\
 & \cellcolor{T2IGray}w/ \topicRoute & \cellcolor{T2IGray}1.7953 & \cellcolor{T2IGray}0.0384 & \cellcolor{T2IGray}\textbf{0.7655} & \cellcolor{T2IGray}0.6864 & \cellcolor{T2IGray}0.2508 & \cellcolor{T2IGray}0.2687 & \cellcolor{T2IGray}4 & \cellcolor{T2IGray}3 & \cellcolor{T2IGray}No \\
\midrule
\multirow{2}{*}{Anole-7B} & SJD & 1.3781 & 0.0559 & 0.7351 & \textbf{0.6456} & -0.6832 & \textbf{0.2123} &  &  &  \\
 & \cellcolor{T2IGray}w/ \topicRoute & \cellcolor{T2IGray}\textbf{1.3820} & \cellcolor{T2IGray}\textbf{0.0519} & \cellcolor{T2IGray}\textbf{0.7367} & \cellcolor{T2IGray}0.6436 & \cellcolor{T2IGray}\textbf{-0.6694} & \cellcolor{T2IGray}0.2120 & \cellcolor{T2IGray}7 & \cellcolor{T2IGray}6 & \cellcolor{T2IGray}Yes \\
\midrule
\multirow{2}{*}{FLUX.1-schnell} & TaylorSeer & 1.6341 & 0.0197 & \textbf{0.8957} & 0.7621 & \textbf{1.0237} & \textbf{0.2790} &  &  &  \\
 & \cellcolor{T2IGray}w/ \topicRoute & \cellcolor{T2IGray}\textbf{1.7838} & \cellcolor{T2IGray}\textbf{0.0123} & \cellcolor{T2IGray}0.8804 & \cellcolor{T2IGray}\textbf{0.7823} & \cellcolor{T2IGray}0.9451 & \cellcolor{T2IGray}0.2607 & \cellcolor{T2IGray}7 & \cellcolor{T2IGray}2 & \cellcolor{T2IGray}No \\
\midrule
\multirow{2}{*}{FLUX.1-dev} & SpeCa & 4.6932 & 0.0228 & 0.8824 & 0.7089 & \textbf{0.8682} & \textbf{0.2841} &  &  &  \\
 & \cellcolor{T2IGray}w/ \topicRoute & \cellcolor{T2IGray}\textbf{4.7277} & \cellcolor{T2IGray}\textbf{0.0203} & \cellcolor{T2IGray}\textbf{0.8887} & \cellcolor{T2IGray}\textbf{0.7095} & \cellcolor{T2IGray}0.8681 & \cellcolor{T2IGray}0.2839 & \cellcolor{T2IGray}4 & \cellcolor{T2IGray}5 & \cellcolor{T2IGray}Yes \\
\bottomrule
\end{tabular}
\endgroup
\end{table*}

\begin{figure*}[t]
\centering
\includegraphics[width=\textwidth]{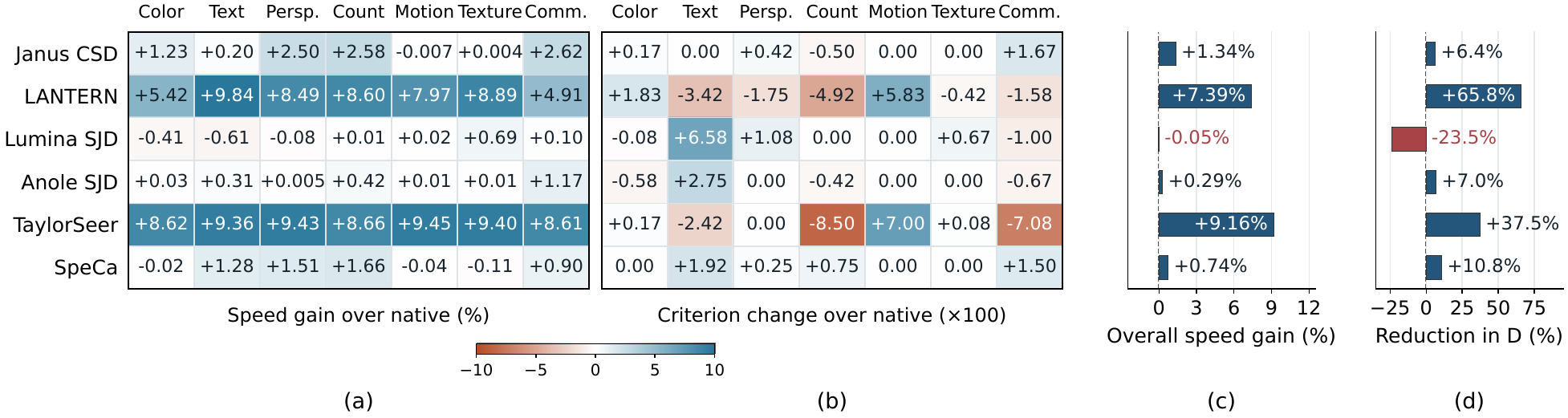}
\caption{Six native methods with and without \topicRoute; zero is the native method. (a) Topic speed gain $100(U_g-1)\%$. (b) Topic criterion change $100(Q_g^{\rm plugin}-Q_g^{\rm native})$. (c) Overall speed gain $100(U-1)\%$. (d) Disparity reduction $100(1-D_{\rm plugin}/D_{\rm native})\%$.}
\label{fig:plugin-evaluation}
\end{figure*}

\subsection{Methods}
\label{sec:plugin-methods}
\paragraph{Prompt routing.}
\topicRoute selects a policy for an existing accelerator without retraining
the generator or adding a learned drafter (Algorithm~\ref{alg:topicroute}). A fitted multinomial Naive Bayes classifier uses word unigrams and bigrams to
predict a rendering topic from the prompt; it is trained on construction
candidates disjoint from the benchmark. The predicted topic selects a frozen
policy, and Janus and Anole also use a deterministic prompt hash to limit how
often an override is enabled. LANTERN and TaylorSeer use the same policy for
all prompts. The frozen policies were set with development runs and earlier
benchmark feedback on the same prompts, and the DrawBench~\citep{saharia2022photorealistic} evaluation in
Section~\ref{sec:plugin-results} uses them without retuning.

\begin{algorithm}[t]
\caption{\topicRoute: prompt and execution controls}
\label{alg:topicroute}
\small
\begin{algorithmic}[1]
\Require Prompt $p$, native method $M$, frozen policies $\Pi$
\State Select $\pi$ from $M$, predicted topic, and optional hash
\State Initialize state and any online cost estimates
\While{generation is incomplete}
  \State Read progress, boundaries, and cache signals
  \State Form the intervention permitted by $\pi$
  \If{$\pi$ enables online gain screening}
    \State Choose a feasible horizon passing the gain screen
  \EndIf
  \State Apply fallback, horizon override, or cache action
  \State Execute $M$; record actual cost and progress
  \State Update active estimates, guards, and budget
\EndWhile
\end{algorithmic}
\end{algorithm}

\paragraph{Online control where applicable.}
The LANTERN, Lumina, and selected Anole paths can further choose a feasible
horizon from causal progress and cost observations. Heuristic lower/upper
estimates at iteration $t$ give
\begin{equation}
 G_t^L(h)=P_t^L(h)C_{T,t}^L-C_{\mathrm{attempt},t}^U(h),
 \label{eq:plugin-routing}
\end{equation}
where $P^L$, $C_T^L$, and $C_{\mathrm{attempt}}^U$ estimate progress,
reference cost per unit, and attempt cost, respectively. Remaining work,
structural boundaries, and configured risk and budget tests restrict
candidate horizons. A horizon passes the gain screen when its lower gain
estimate exceeds the configured margin.
State and estimates reset per image. A declined override returns to the
native accelerator. The selected SpeCa policy skips the gain screen.

\paragraph{Backend controls.}
Janus CSD uses a fixed window per prompt and changes it from 16 to 20 or 32
for four predicted topics. LANTERN caps the verified draft path, and Lumina
and Anole SJD alter block horizons while preserving
native proposal, verification, and cache updates. SpeCa relaxes the cache
threshold in the second half of the schedule for four predicted topics, and
TaylorSeer shortens its initial phase of full computation and partially
refreshes transformer blocks at the final step; their cache guards do not
roll back the latent trajectory. The AR policies keep the sampling and
acceptance rules of their accelerators, and Appendix~\ref{sec:method} gives
the exact policies.

\subsection{Evaluation Results}
\label{sec:plugin-results}

\paragraph{Matched comparisons of complete images.}
We compare six model--method pairs with and without \topicRoute on 1,400
prompts each. Every group includes target-only, native, plugin, and
observation-only arms, and within each group all arms share base generation
settings. The observation-only arm records decisions without
applying them. The six groups are separate runs, each with its own
target-only arm and timed by request latency. Request latency covers
generation and the routing work of each runner.
Table~\ref{tab:plugin-main} reports target/method speed $S$, the
number of faster topics $W$, and speed spread $D=\max_g S_g/\min_g S_g-1$;
the ratio of the two speeds is $U=\bar t_{\rm native}/\bar t_{\rm plugin}$,
and $\Delta Q$ subtracts the native $Q$ from the plugin $Q$.
The group criterion requires $U>1$, $S>1$,
$W\ge4$, a smaller $D$, and nondecreasing means in at least five of eight
quality metrics ($K\ge5$).

\paragraph{Speed and topic robustness.}
Four groups meet all five criteria. Across all six, the geometric mean of
$U$ is $1.0308\times$, and five groups reduce topic disparity
(Figure~\ref{fig:plugin-evaluation}). Five groups have 95\% intervals for
$U$ above one (Appendix~\ref{app:plugin-followup}). LANTERN
accelerates every topic by 4.91\%--9.84\%, with $U=1.0739$ and $D$ falling
from 0.0659 to 0.0226, a relative reduction of 65.8\%; its plugin is also faster than target-only generation
($S=1.0143$). Janus CSD, Anole SJD, and SpeCa are faster in every topic
they route to a new policy (Appendix~\ref{sec:method}), with Overall ratios of 1.0134, 1.0029, and
1.0074. Lumina SJD is faster in two of the five topics it routes but has
$U=0.9995$ and a larger $D$.

\paragraph{Quality and failure modes.}
LANTERN meets the quality criterion with five nondecreasing metrics but
loses 0.0063 in $Q$; its
motion gain of 0.0583 coexists with a counting loss of 0.0492.
Janus, Anole, SpeCa, and Lumina increase $Q$, with $K=3$ for Lumina.
TaylorSeer gives the largest speed gain
of 9.16\% ($U=1.0916$) and reduces disparity, but only two of eight metrics
are nondecreasing, with $\Delta Q=-0.0154$ and a counting loss of 0.0850.
Averaged equally across the six groups, $\Delta Q$
is $-0.0002$, and quality changes vary by backend, topic, and metric.
Appendix~\ref{app:plugin-evaluation} reports topic and difficulty
results and all eight quality metrics.

\paragraph{Ablations, reruns, and external prompts.}
In ablations of the final policies (Appendix~\ref{app:plugin-followup}),
Janus's topic routing and LANTERN's gain screen mainly improve topic balance
at some cost in speed, threshold relaxation produces SpeCa's gain, and the
hash gates of Janus and Anole add neither speed nor balance. The speed gains
recur in reruns of the final policies, but of the four passing groups only
Anole passes again in its rerun. On 200 external DrawBench prompts disjoint from the benchmark
and development sets, five frozen policies remain faster than native without
retuning, with intervals above one, and the geometric mean of $U$ across all six is $1.0375\times$.
Relative to native methods, TaylorSeer lowers five of six generic quality
metrics, while the other groups have three to five nondecreasing means.

\FloatBarrier
\section{Conclusion}

\tTwoISpec evaluates speculative decoding for T2I generation in seven
rendering topics. Topic comparisons reveal latency variation, quality
regressions, and metric disagreement, and large reductions in target work
can yield limited latency gains.
Parameter sweeps link longer horizons to wasted work and aggressive caching
to quality losses in individual topics.
These findings motivate \topicRoute, which routes prompts and controls
existing accelerators without retraining the generators: across all six pairs,
its geometric mean speedup over native methods is $1.0308\times$ and carries
over to external DrawBench prompts, while quality changes vary by backend,
topic, and metric.

\section*{Use of Generative AI Tools}
OpenAI Codex assisted in building the deterministic pipeline that derives
candidate prompts for each topic and difficulty level from existing reference
prompts (Section~\ref{sec:dataset}), in adapting interfaces to the evaluated
models and methods, and in designing \topicRoute and integrating it with these
methods. Anthropic Claude (through Claude Code) and OpenAI Codex were used to
revise manuscript text under the authors' direction, including polishing
paragraphs written by the authors, and to assist with analyzing selected
results and checking reported numbers against the experimental records. OpenAI
ChatGPT was used to identify potential issues and give feedback on the
manuscript. As part of the benchmark protocol, the vision-language models
Qwen2.5-VL-7B-Instruct and InternVL3-8B evaluate how well generated images
satisfy the prompt requirements (Appendix~\ref{app:evaluation-protocol}). The
authors take full responsibility
for the content and conclusions of this work.

\begingroup
\small
\bibliographystyle{ieeenat_fullname}
\bibliography{references}
\endgroup

\clearpage
\appendix
\maketitlesupplementary
\makeatletter
\setlength{\@fptop}{0pt}
\setlength{\@dblfptop}{0pt}
\setlength{\@fpsep}{12pt}
\setlength{\@dblfpsep}{12pt}
\makeatother
\raggedbottom
\section{Dataset Details}
\label{app:dataset-details}
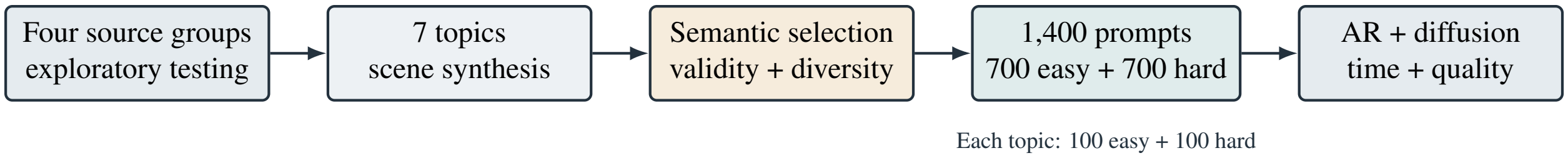
\begin{figure*}[t]
\centering
\resizebox{\linewidth}{!}{%
\begin{tikzpicture}[
  node distance=6mm,
  every node/.style={font=\small},
  box/.style={draw=T2IDark, rounded corners=2pt, thick, align=center,
    minimum height=10mm, minimum width=28mm, inner sep=4pt},
  arrow/.style={-{Latex[length=2mm]}, thick, draw=T2IDark}
]
\node[box, fill=T2IBlue!12] (ref) {Four source groups\\exploratory testing};
\node[box, fill=T2IBlue!8, right=of ref] (cand) {7 topics\\scene synthesis};
\node[box, fill=T2IGold!16, right=of cand] (select) {Semantic selection\\validity + diversity};
\node[box, fill=T2ITeal!14, right=of select] (release) {1,400 prompts\\700 easy + 700 hard};
\node[box, fill=T2IBlue!12, right=of release] (tracks) {AR + diffusion\\time + quality};
\draw[arrow] (ref) -- (cand);
\draw[arrow] (cand) -- (select);
\draw[arrow] (select) -- (release);
\draw[arrow] (release) -- (tracks);
\node[below=2mm of release, font=\scriptsize, text=T2IDark] {Each topic: 100 easy + 100 hard};
\end{tikzpicture}%
}
\caption{Benchmark construction and evaluation. Exploratory tests on MS-COCO,
PartiPrompts, GenEval, and T2I-CompBench/++ motivate seven rendering topics.
Scene synthesis guided by references and semantic selection produce
one workload of 1,400 prompts, partitioned into 700 easy and 700 hard prompts.
Model evaluation measures generation time and image quality separately.}
\label{fig:dataset-pipeline}
\end{figure*}

\subsection{Construction and Scene Relationships}

Construction (Figure~\ref{fig:dataset-pipeline}) starts from 700 curated
references, 100 for each of the seven
topics. References provide provenance, and each seeds four new candidate
scenes, two easy and two hard. Forty
references per topic also seed an easy/hard pair that shares a newly composed
scene, giving 480 candidates per topic and 3,360 candidates overall.

Selection retains one easy and one hard prompt per reference, preserves the
distribution of reference categories, and includes 20 easy/hard pairs per
topic that share a scene. Each candidate's score rewards relevance, measured
as cosine similarity to related references under the all-MiniLM-L6-v2
sentence encoder, and penalizes word overlap with the most similar reference
in its topic to favor originality. A
deterministic selection then trades this score against the diversity of the
chosen scenes. Two candidates from different scenes cannot both be selected
when their token Jaccard similarity is at least 0.80 and their character
4-gram Jaccard similarity is at least 0.86. Scene composition uses a fixed
seed, so the build is reproducible. The selected prompts contain no exact
duplicates or copied reference prefixes, and their maximum token Jaccard
similarity to a reference is 0.20.

The final dataset contains 1,400 prompts. By scene provenance, 1,120 prompts
have standalone scenes, and 280 form 140 easy/hard pairs that share a scene,
giving 1,260 scene families.

Easy prompts average 41.92 words (median 41), and hard prompts average 53.31
words (median 53). The mean character counts are 276.84 and 366.59,
respectively.

\subsection{Topics, Categories, and Examples}
\label{app:categories}

Each topic has five descriptive category definitions, such as object--color
binding, public signage, occlusion, nested counting, fluid motion, layered
food materials, and scale--context reversals. Thirty categories are nonempty, and five
are unused taxonomy labels; the nonempty categories contain 2 to 156 prompts.
Summaries by topic follow
the balanced design, whereas categories retain the reference distribution.

Table~\ref{tab:prompt-excerpts} illustrates the intended constraints. The
entries summarize the prompts for space; evaluation uses the
complete canonical prompts.

\begin{table*}[!tp]
\centering
\scriptsize
\begin{tabularx}{\linewidth}{@{}p{0.14\linewidth}XX@{}}
\toprule
Topic & Easy summary & Hard summary \\
\midrule
Color & A silver brass clock, violet wool basket, and charcoal frosted glass tray share a display. & A coral clock, turquoise pitcher, and amber fan remain locally colored under partial occlusion. \\
Text/symbols & An index card shows QUIET ZONE, $3 \times 4 = 12$, a four-row table, and DESK 12. & Every requested word, digit, and symbol keeps its sequence and orientation despite borders and wear. \\
Perspective & Square doorframes converge down a museum corridor; a ceramic pot establishes scale. & Boundaries, contacts, and depth cues must support all relations from one coherent camera. \\
Counting & Exactly ten lanterns form two straight rows of five with equal spacing. & Exactly ten apples form a $2\times5$ grid partially interrupted by a foreground object yet still countable. \\
Motion & A galloping horse remains readable while a light motion trace marks its trajectory. & Fast and slow components require distinct plausible blur, with pose, contacts, and shadow consistent. \\
Texture & Butterfly wing scales and bounded color patches follow the natural three-dimensional form. & Fine and coarse cat fur structure must remain distinct and follow folds, curves, and layers. \\
Commonsense & An upright elephant records leaf measurements as a botanist in a greenhouse. & An upright raven completes a multi-step florist task using role-specific tools among coworkers. \\
\bottomrule
\end{tabularx}
\caption{Abbreviated prompt requirements across the seven topics. Easy and
Hard are construction labels.}
\label{tab:prompt-excerpts}
\end{table*}

\section{Evaluation Protocol}
\label{app:evaluation-protocol}
\label{app:baseline-details}
\begin{table*}[t]
\centering
\small
\begin{tabularx}{\linewidth}{@{}l l X@{}}
\toprule
Method & Family & Drafting or acceleration mechanism \\
\midrule
\rowcolor{T2IGray}
\multicolumn{3}{c}{\textit{Training-based methods}} \\
LANTERN & AR & Learned auxiliary drafter; relaxed acceptance of visual tokens \\
LANTERN++ & AR & Learned drafter with static draft trees \\
ADT-Tree + LANTERN & AR & Learned drafter with trees that adapt to adjacency \\
VVS & AR & Learned visual drafter with partial verification skipping \\
\midrule
\rowcolor{T2IGray}
\multicolumn{3}{c}{\textit{Training-free methods}} \\
GSD & AR & Dynamic acceptance of token groups \\
SJD & AR & Jacobi predictions with target verification \\
SCD / MC-SJD & AR & Gumbel (SCD) or maximal (MC-SJD) coupling across speculative iterations \\
CSD & AR & Content-aware speculative control \\
SJD-PAC & AR & Proactive drafting and adaptive continuation \\
PathSpec & AR & Parallel grouped decoding paths \\
ARsample filtering & AR & Sampling guided by entropy; related efficiency baseline \\
TaylorSeer & Diffusion/flow & Forecasting of intermediate features from previous steps \\
SpeCa & Diffusion/flow & Feature caching adapted to sample complexity \\
ClusCa & Diffusion/flow & Spatially clustered feature updates \\
ASD & Diffusion & Parallel proposals using exchangeable increments \\
\midrule
\rowcolor{T2IGray}
\multicolumn{3}{c}{\textit{Methods classified by their proposals}} \\
Diffusion SD & Diffusion & Continuous speculative proposals with acceptance and correction \\
Block verification & Diffusion & Joint verification of proposed sampling blocks \\
\bottomrule
\end{tabularx}
\caption{Algorithm taxonomy by drafting mechanism. Training-based methods use
trained auxiliary weights; training-free methods use forecasting, reuse,
proposals from the target model, or sampling filters. Diffusion SD and block verification are classified by
their chosen proposals, as discussed in Section~\ref{sec:algorithms}; the result tables group them
with ASD as SDE adaptations.}
\label{tab:method-taxonomy}
\end{table*}

\paragraph{Coverage and comparison controls.}
The core AR study pairs the Janus target-only reference with CSD, LlamaGen
Stages~I and II with LANTERN and LANTERN++, and Lumina with GSD, SJD, SCD,
and MC-SJD. Each configuration has 200 outputs per topic and 1,400 Overall,
and all outputs have valid images and recorded automatic metrics. In the
15 paired configurations, each arm has 200 records per topic and 100 per
topic/difficulty stratum. Counting each method with its group target, which
serves every method in the group, gives 42,000 scored arm records.

\paragraph{Target generation settings.}
The core Lumina target uses $768^2$ resolution, classifier-free guidance
(CFG) 3.0, and top-$k$ 2,000. The LlamaGen Stage~I and Stage~II
training-based targets use $256^2$ and $512^2$, respectively, with CFG 7.5
and top-$k$ 1,000. These core targets use BF16, temperature 1.0, and top-$p$
1.0, and the core Janus target uses $384^2$, BF16, CFG 5.0, and temperature
1.0. The separate Stage~II training-free reference for ARsample and SJD uses
$512^2$, BF16, CFG 7.5, and temperature 1.4.
FLUX.1-dev uses $512^2$, BF16 model weights, 50 target steps, and guidance 3.5.
The FLUX.1-dev training-free group uses the ClusCa target, and the SDE group
uses the target of the ASD $h=8$ run, with a uniform schedule, localization
horizon 1,024, and FP32 sampler states.
The Anole, Emu3~\citep{wang2024emu3}, and FLUX.1-schnell runs
record their seeds but not their resolution, guidance, or sampling settings.

\paragraph{Controls for individual methods.}
Core CSD uses a proposal window of 16. Core LANTERN and LANTERN++ use a
neighborhood size of 1,000 and relaxation setting 0.1; LANTERN++ adds a static draft tree.
Lumina GSD and SJD use windows of 16, while SCD and MC-SJD use windows of 32
with Gumbel and maximal coupling, respectively.

\paragraph{Timing measurements.}
Of the 15 paired configurations, SJD on Anole and on Emu3 report elapsed
runner time, and the other 13 report request latency. Core warm timings use
three timed repeats after three warmups. Each reported run is timed against
its own matched target-only reference. The strict-512 evaluation denotes the
core LlamaGen Stage~II $512^2$ runs, which use seeds derived from prompt
identities and warm timing.

\paragraph{Quality references.}
Quality comparisons use one target reference per model and method category,
with training-based, training-free, and SDE groups kept separate. Every
method in a group uses the generation configuration of its target. The
tables of absolute quality and Easy/Hard results show it once per category
as Target-only.

\paragraph{Measurement identity.}
The core AR study has 13 arms, four targets and nine methods, with 1,400
outputs each (18,200 in total); each method output is paired with the timing
record of its target. Its summaries are recomputed from the individual sample
records, which cover the same prompt identities in every arm.
Prompt identities, completion flags, and target/method pairing are checked
before aggregation.

\paragraph{Automatic scores.}
The primary topic evaluator is Qwen2.5-VL-7B-Instruct (checkpoint
cc594898), and the second evaluator is InternVL3-8B (853e3a79).
In the core AR evaluation, both judges use a 512-token output cap and
greedy decoding with a single beam.
Each image receives three criterion scores for its topic in
$\{0,0.5,1\}$, whose mean is the image score $q_i$; $Q$ and $Q_2$ average the
image scores of the two evaluators, respectively, over a prompt group. For counting, the
criteria cover
the exact count, the count error, and the group layout. All criterion scores,
including those for text and counting, are rubric judgments by a
vision-language model. The judge receives the full prompt, the topic, deterministically
extracted gold cues, the allowed scores, and a response schema that requires
visible evidence for each criterion. Parsing requires the three
criterion keys and nonempty evidence; the core AR study rejects scores outside $\{0,0.5,1\}$, and the other scoring runs map them to the nearest allowed value. The stored rubric labels
0, 0.5, and 1 as absent or contradicted, partial or ambiguous, and clearly and
completely satisfied, respectively; the judge sees only the numbers. For the LlamaGen Stage~I target-only arm, the two
evaluators give mean criterion scores of 0.5998 and 0.3631, respectively.
The other six scalar metrics are
CLIPScore with OpenCLIP ViT-L/14, ImageReward, HPS v2.1, the LAION aesthetic
predictor, SigLIP alignment with so400m, and the status of the Stable
Diffusion safety checker. Appendix~\ref{app:topic-details} reports
the complementary image quality measures. The intervals in
Appendix~\ref{app:resampling-details} resample connected groups of prompts
that share a scene or creative seed, so dependent generations stay together.

\paragraph{ADT-Tree, ARsample, and PathSpec.}
Each of these configurations contains 1,400 target/method pairs.
Source IDs and prompt hashes align these evaluations with all 1,400
canonical prompts, preserving topic, difficulty, category, and scene family.
ADT-Tree uses a
dynamic draft tree with LANTERN relaxation (neighborhood size 100,
relaxation setting 0.5). ARsample uses CFG 7.5 and temperature 1.4, and
PathSpec uses a static configuration with grouped trees.
Table~\ref{tab:compact-target-work} reports their
target work and the number of pairs with lower method latency.
Quality scores and timings cover the same prompt identities.

\paragraph{Diffusion configurations.}
All three SDE adaptations use $h=8$ for both timing and quality, a uniform
sampling schedule, FP32 sampler states, and BF16 model weights.

\paragraph{Metric coverage of the paired configurations.}
The 15 paired configurations use the same eight scalar metric definitions
as the core AR study. The primary criterion averages the scores for three
frozen requirements per prompt, and the second criterion uses InternVL3-8B.
Quality scores are linked to the generation settings and canonical
prompt/image identities.

\section{Supplementary Benchmark Results}
\label{app:benchmark-results}

The main and supplementary tables report the same quality configurations,
including the Easy/Hard comparisons. Auxiliary timings retain matched
target/method pairs; the SDE rows use the $h=8$ runs of the main comparisons.
The tables appear in the same order as this
section: quality, difficulty and uncertainty, parameter sensitivity, and
absolute latency and target work.

\subsection{Complementary Quality}
\label{app:topic-details}
\label{app:shared-quality-details}
\label{app:first-round-quality-details}

Tables~\ref{tab:compact-topic-clip}--\ref{tab:compact-topic-hps} report
CLIPScore, ImageReward, and HPS in all seven topics for every quality
configuration, with AR and diffusion grouped within each table. Each group's
target appears once, above its methods. Each topic contains 200 outputs per
configuration.
Table~\ref{tab:consolidated-quality-overall} reports the Overall changes.

Table~\ref{tab:compact-ancillary} reports Overall SigLIP alignment, aesthetic
score, the criterion of the second evaluator ($Q_2$), and the automatic
safety status. Each Overall mean uses 1,400 outputs. Quality
changes quoted in the text use unrounded paired values.

The ADT-Tree and ARsample quality profiles in the main comparison also
show opposing topic changes. ADT-Tree gains $0.0975$ on motion but loses
$0.0492$ on counting, and its criterion score falls in two of seven
topics. ARsample gains $0.0233$ on text but loses $0.0242$ on commonsense
violation, so its unchanged Overall criterion conceals larger, opposing
topic effects.

\subsection{Difficulty and Uncertainty}
\label{app:resampling-details}

\paragraph{Consolidated construction strata.}
Table~\ref{tab:consolidated-difficulty} reports Easy/Hard
results by generation family, including the stated quality references;
latency ratios use each run's matched timing reference. Each configuration has 700 outputs per
stratum, and Equation~\ref{eq:observed-ratio} defines the timing
aggregations.

\paragraph{Paired resampling of dependent prompts.}
We reconstruct connected components by linking prompts that share a scene
family or the creative seed used during construction. This produces
700 groups, each containing one Easy and one Hard prompt, with 100 groups per
topic; a group therefore contains two outputs per AR configuration. The
grouping agrees with the group assignment of every output.

For each of 2,000 bootstrap replicates, we draw 100 groups with replacement
within each topic, using random seed 20260930. All rows in a selected group
and their matched target rows receive the same multiplicity. Quality changes
are means of paired differences, and latency ratios are medians of individual
matched ratios. Overall estimates pool all seven resampled topics instead of
averaging topic medians. The 2.5th and 97.5th percentiles, with linear
interpolation, define the intervals in Tables~\ref{tab:core-confidence}
and~\ref{tab:topic-confidence}. All methods and metrics use
the same resampling indices.

These pointwise intervals quantify sampling variation across
the observed groups while keeping their paired Easy/Hard prompts together.
Quality estimates use the same fixed references as the main tables, whereas
latency intervals use the auxiliary timing runs.

The main tables report topic $Q$ and speed, and the pooled Easy/Hard
comparisons here use the same quality references.

\subsection{Parameter Sensitivity}
\label{app:parameter-sweeps}

Table~\ref{tab:parameter-sweeps-overall} reports ten evaluations of five
existing diffusion methods. Each evaluation covers the same 1,400 canonical
prompts: 200 per topic, 700 per construction stratum, and 100 per
topic/stratum intersection. Generation uses FLUX.1-dev at $512^2$, BF16,
50 target steps, and guidance 3.5. The SDE adaptations use a uniform
schedule. Each SDE adaptation compares horizons 2 and 8, and its
$h=8$ run supplies the main comparisons. SpeCa varies its base threshold, and
ClusCa varies its freshness threshold.

Each row keeps the target timing of its own run, and its quality changes
subtract the scores of its group target: the ClusCa target for SpeCa and
ClusCa, and the target with a uniform schedule for the SDE adaptations.

\begin{table}[!t]
\centering
\caption{Overall quality changes over 1,400 prompts. IR denotes ImageReward, HPS the human preference score, and $Q$ the automated topic criterion. Positive changes indicate higher scores.}
\label{tab:consolidated-quality-overall}
\scriptsize
\setlength{\tabcolsep}{3pt}
\renewcommand{\arraystretch}{1.05}
\begin{tabularx}{\columnwidth}{@{}>{\raggedright\arraybackslash}Xrrrr@{}}
\toprule
Method & $\Delta$CLIP & $\Delta$IR & $\Delta$HPS & $\Delta Q$ \\
\midrule
\multicolumn{5}{l}{\cellcolor{T2IGray}\textbf{Autoregressive}} \\
\multicolumn{5}{l}{\textit{Janus-Pro-7B, $384^2$}} \\
CSD & +0.0003 & -0.0143 & -0.0014 & +0.0127 \\
SJD & -0.1406 & -1.4862 & -0.0545 & -0.1095 \\
\midrule
\multicolumn{5}{l}{\textit{LlamaGen Stage I, $256^2$}} \\
LANTERN & +0.0195 & +0.1467 & +0.0143 & +0.0499 \\
LANTERN++ & -0.0028 & -0.0276 & -0.0055 & -0.0162 \\
VVS & +0.0209 & +0.1351 & +0.0102 & +0.0657 \\
\midrule
\multicolumn{5}{l}{\textit{LlamaGen Stage II, $512^2$}} \\
\multicolumn{5}{l}{\textit{\textbf{Training-based Methods}}} \\
LANTERN & -0.0104 & -0.0830 & -0.0006 & -0.0019 \\
LANTERN++ & -0.0353 & -0.5128 & -0.0393 & -0.0787 \\
\multicolumn{5}{l}{\textit{\textbf{Training-free Methods}}} \\
ARsample filtering & -0.0059 & -0.0847 & -0.0026 & 0.0000 \\
SJD & +0.0168 & +0.0297 & +0.0007 & -0.0008 \\
\midrule
\multicolumn{5}{l}{\textit{Lumina-mGPT-7B, $768^2$}} \\
GSD & -0.0123 & -0.0221 & +0.0045 & -0.0002 \\
MC-SJD & -0.0069 & +0.0073 & +0.0056 & +0.0099 \\
SCD & -0.0072 & -0.0002 & +0.0058 & -0.0099 \\
SJD & -0.0073 & +0.0174 & +0.0056 & -0.0040 \\
PathSpec & -0.0077 & +0.0185 & +0.0057 & -0.0051 \\
SJD-PAC & -0.0056 & +0.0368 & +0.0067 & -0.0033 \\
\midrule
\multicolumn{5}{l}{\textit{Anole-7B}} \\
\multicolumn{5}{l}{\textit{\textbf{Training-based Methods}}} \\
ADT-Tree + LANTERN & -0.0079 & -0.0586 & +0.0015 & +0.0096 \\
\multicolumn{5}{l}{\textit{\textbf{Training-free Methods}}} \\
SJD & -0.0007 & -0.0060 & +0.0002 & +0.0005 \\
\midrule
\multicolumn{5}{l}{\textit{Emu3}} \\
SJD & 0.0000 & +0.0182 & 0.0000 & +0.0043 \\
\midrule
\multicolumn{5}{l}{\cellcolor{T2IGray}\textbf{Diffusion/flow}} \\
\multicolumn{5}{l}{\textit{FLUX.1-schnell}} \\
TaylorSeer & +0.0191 & +0.0286 & +0.0006 & +0.0093 \\
\midrule
\multicolumn{5}{l}{\textit{FLUX.1-dev}} \\
\multicolumn{5}{l}{\textit{\textbf{Training-free Methods}}} \\
ClusCa & +0.0048 & -0.0232 & -0.0045 & -0.0002 \\
SpeCa & +0.0056 & +0.0074 & -0.0028 & +0.0037 \\
\multicolumn{5}{l}{\textit{\textbf{SDE adaptations}}} \\
ASD & -0.0022 & +0.0001 & +0.0008 & -0.0077 \\
Block verification & +0.0025 & +0.0003 & +0.0014 & -0.0099 \\
Diffusion SD & -0.0028 & 0.0000 & +0.0008 & -0.0025 \\
\bottomrule
\end{tabularx}
\end{table}

\begin{table*}[!tp]
\centering
\caption{Auxiliary analysis. SDE rows use their $h=8$ runs. CLIPScore by topic ($\uparrow$), grouped by generation family and target model. Each row covers 200 prompts per topic; each group's target appears once. Values are recorded arm means.}
\label{tab:compact-topic-clip}
\label{tab:compact-topic-ar-clip}
\label{tab:compact-topic-diffusion-clip}
\scriptsize
\setlength{\tabcolsep}{4pt}
\renewcommand{\arraystretch}{1.10}
\begin{tabular}{@{}lrrrrrrr@{}}
\toprule
Method & Color & Text & Persp. & Count & Motion & Texture & Common. \\
\midrule
\multicolumn{8}{l}{\cellcolor{T2IGray}\textbf{Autoregressive}} \\
\multicolumn{8}{l}{\textit{Janus-Pro-7B, $384^2$}} \\
\multicolumn{8}{l}{\textit{\textbf{Training-free Methods}}} \\
Target-only & 0.7553 & 0.7483 & 0.7248 & 0.7081 & 0.6999 & 0.7058 & 0.7806 \\
CSD & 0.7571 & 0.7388 & 0.7339 & 0.7094 & 0.6994 & 0.7018 & 0.7846 \\
SJD & 0.6116 & 0.5313 & 0.5889 & 0.5970 & 0.5924 & 0.5830 & 0.6347 \\
\midrule
\multicolumn{8}{l}{\textit{LlamaGen Stage I, $256^2$}} \\
\multicolumn{8}{l}{\textit{\textbf{Training-based Methods}}} \\
Target-only & 0.5303 & 0.4809 & 0.4778 & 0.5137 & 0.5311 & 0.5015 & 0.4757 \\
LANTERN & 0.5337 & 0.4841 & 0.5275 & 0.5259 & 0.5466 & 0.5301 & 0.4998 \\
LANTERN++ & 0.5090 & 0.4683 & 0.4989 & 0.5048 & 0.5202 & 0.5263 & 0.4635 \\
VVS & 0.5332 & 0.4933 & 0.5269 & 0.5250 & 0.5516 & 0.5296 & 0.4981 \\
\midrule
\multicolumn{8}{l}{\textit{LlamaGen Stage II, $512^2$}} \\
\multicolumn{8}{l}{\textit{\textbf{Training-based Methods}}} \\
Target-only & 0.6334 & 0.5718 & 0.6390 & 0.6439 & 0.6481 & 0.6393 & 0.6887 \\
LANTERN & 0.6165 & 0.5548 & 0.6273 & 0.6311 & 0.6458 & 0.6410 & 0.6748 \\
LANTERN++ & 0.5905 & 0.5361 & 0.6032 & 0.6194 & 0.6118 & 0.6151 & 0.6410 \\
\multicolumn{8}{l}{\textit{\textbf{Training-free Methods}}} \\
Target-only & 0.6172 & 0.5367 & 0.6256 & 0.6346 & 0.6366 & 0.6298 & 0.6782 \\
ARsample filtering & 0.6027 & 0.5306 & 0.6274 & 0.6244 & 0.6253 & 0.6386 & 0.6683 \\
SJD & 0.6367 & 0.5766 & 0.6329 & 0.6378 & 0.6575 & 0.6351 & 0.6997 \\
\midrule
\multicolumn{8}{l}{\textit{Lumina-mGPT-7B, $768^2$}} \\
\multicolumn{8}{l}{\textit{\textbf{Training-free Methods}}} \\
Target-only & 0.7109 & 0.6216 & 0.6925 & 0.7053 & 0.6863 & 0.6803 & 0.7471 \\
GSD & 0.7005 & 0.5951 & 0.6911 & 0.6943 & 0.6806 & 0.6668 & 0.7296 \\
MC-SJD & 0.7071 & 0.6162 & 0.6991 & 0.7007 & 0.6703 & 0.6685 & 0.7339 \\
SCD & 0.7031 & 0.6179 & 0.6968 & 0.6925 & 0.6789 & 0.6720 & 0.7325 \\
SJD & 0.6975 & 0.6055 & 0.6956 & 0.7016 & 0.6804 & 0.6758 & 0.7367 \\
PathSpec & 0.7078 & 0.6094 & 0.6896 & 0.6959 & 0.6834 & 0.6768 & 0.7274 \\
SJD-PAC & 0.7101 & 0.6045 & 0.6935 & 0.6973 & 0.6785 & 0.6781 & 0.7429 \\
\midrule
\multicolumn{8}{l}{\textit{Anole-7B}} \\
\multicolumn{8}{l}{\textit{\textbf{Training-based Methods}}} \\
Target-only & 0.6596 & 0.5698 & 0.6452 & 0.6331 & 0.6572 & 0.6462 & 0.6321 \\
ADT-Tree + LANTERN & 0.6500 & 0.5498 & 0.6298 & 0.6257 & 0.6513 & 0.6481 & 0.6336 \\
\multicolumn{8}{l}{\textit{\textbf{Training-free Methods}}} \\
Target-only & 0.6818 & 0.5903 & 0.6700 & 0.6411 & 0.6700 & 0.6404 & 0.6366 \\
SJD & 0.6785 & 0.5863 & 0.6705 & 0.6458 & 0.6700 & 0.6372 & 0.6373 \\
\midrule
\multicolumn{8}{l}{\textit{Emu3}} \\
\multicolumn{8}{l}{\textit{\textbf{Training-free Methods}}} \\
Target-only & 0.7170 & 0.6629 & 0.6782 & 0.7104 & 0.6784 & 0.6724 & 0.7320 \\
SJD & 0.7196 & 0.6618 & 0.6789 & 0.7026 & 0.6855 & 0.6715 & 0.7310 \\
\midrule
\multicolumn{8}{l}{\cellcolor{T2IGray}\textbf{Diffusion/flow}} \\
\multicolumn{8}{l}{\textit{FLUX.1-schnell}} \\
\multicolumn{8}{l}{\textit{\textbf{Training-free Methods}}} \\
Target-only & 0.7460 & 0.8523 & 0.6895 & 0.7549 & 0.6590 & 0.6644 & 0.7999 \\
TaylorSeer & 0.7560 & 0.8717 & 0.7228 & 0.7508 & 0.7032 & 0.6819 & 0.8133 \\
\midrule
\multicolumn{8}{l}{\textit{FLUX.1-dev}} \\
\multicolumn{8}{l}{\textit{\textbf{Training-free Methods}}} \\
Target-only & 0.7164 & 0.7865 & 0.6698 & 0.7085 & 0.6519 & 0.6376 & 0.7632 \\
ClusCa & 0.7192 & 0.7841 & 0.6774 & 0.7181 & 0.6613 & 0.6374 & 0.7697 \\
SpeCa & 0.7185 & 0.7824 & 0.6818 & 0.7190 & 0.6599 & 0.6395 & 0.7720 \\
\multicolumn{8}{l}{\textit{\textbf{SDE adaptations}}} \\
Target-only & 0.4070 & 0.4181 & 0.4315 & 0.4221 & 0.3819 & 0.4734 & 0.3683 \\
ASD & 0.4057 & 0.4179 & 0.4244 & 0.4190 & 0.3741 & 0.4771 & 0.3685 \\
Block verification & 0.4050 & 0.4203 & 0.4411 & 0.4212 & 0.3821 & 0.4808 & 0.3691 \\
Diffusion SD & 0.4052 & 0.4180 & 0.4231 & 0.4181 & 0.3743 & 0.4769 & 0.3671 \\
\bottomrule
\end{tabular}
\end{table*}

\begin{table*}[!tp]
\centering
\caption{Auxiliary analysis. SDE rows use their $h=8$ runs. ImageReward by topic ($\uparrow$), grouped by generation family and target model. Each row covers 200 prompts per topic; each group's target appears once. Values are recorded arm means.}
\label{tab:compact-topic-ir}
\label{tab:compact-topic-ar-ir}
\label{tab:compact-topic-diffusion-ir}
\scriptsize
\setlength{\tabcolsep}{4pt}
\renewcommand{\arraystretch}{1.10}
\begin{tabular}{@{}lrrrrrrr@{}}
\toprule
Method & Color & Text & Persp. & Count & Motion & Texture & Common. \\
\midrule
\multicolumn{8}{l}{\cellcolor{T2IGray}\textbf{Autoregressive}} \\
\multicolumn{8}{l}{\textit{Janus-Pro-7B, $384^2$}} \\
\multicolumn{8}{l}{\textit{\textbf{Training-free Methods}}} \\
Target-only & 0.9530 & 0.0690 & 0.7621 & 0.3643 & 0.8019 & 0.4464 & 1.1370 \\
CSD & 0.9499 & 0.0120 & 0.7651 & 0.3957 & 0.8444 & 0.4118 & 1.0550 \\
SJD & -1.0531 & -1.2116 & -1.2552 & -0.6553 & -0.3127 & -0.8959 & -0.4857 \\
\midrule
\multicolumn{8}{l}{\textit{LlamaGen Stage I, $256^2$}} \\
\multicolumn{8}{l}{\textit{\textbf{Training-based Methods}}} \\
Target-only & -2.0855 & -1.7684 & -2.1374 & -1.8323 & -1.3728 & -1.8735 & -1.8773 \\
LANTERN & -1.9380 & -1.6095 & -1.9793 & -1.7227 & -1.1908 & -1.7494 & -1.7305 \\
LANTERN++ & -2.0634 & -1.7076 & -2.0717 & -1.8943 & -1.5715 & -1.8339 & -1.9979 \\
VVS & -1.9741 & -1.6111 & -2.0084 & -1.7402 & -1.1811 & -1.7227 & -1.7637 \\
\midrule
\multicolumn{8}{l}{\textit{LlamaGen Stage II, $512^2$}} \\
\multicolumn{8}{l}{\textit{\textbf{Training-based Methods}}} \\
Target-only & -1.1024 & -1.1405 & -0.9199 & -0.3550 & -0.2712 & -1.1092 & -0.0595 \\
LANTERN & -1.2130 & -1.2538 & -1.0171 & -0.4621 & -0.3611 & -1.0115 & -0.2204 \\
LANTERN++ & -1.5341 & -1.4491 & -1.5289 & -0.8619 & -0.9191 & -1.4897 & -0.7648 \\
\multicolumn{8}{l}{\textit{\textbf{Training-free Methods}}} \\
Target-only & -1.1562 & -1.2337 & -0.9192 & -0.4494 & -0.3515 & -0.9625 & -0.1113 \\
ARsample filtering & -1.3273 & -1.2450 & -0.9675 & -0.6109 & -0.4931 & -0.8935 & -0.2393 \\
SJD & -1.1377 & -1.1523 & -0.9417 & -0.4051 & -0.2081 & -1.0952 & -0.0357 \\
\midrule
\multicolumn{8}{l}{\textit{Lumina-mGPT-7B, $768^2$}} \\
\multicolumn{8}{l}{\textit{\textbf{Training-free Methods}}} \\
Target-only & 0.0029 & -0.2616 & -0.2022 & 0.5834 & 0.6050 & 0.2353 & 0.5125 \\
GSD & -0.0211 & -0.2547 & -0.1693 & 0.5166 & 0.5025 & 0.3010 & 0.4460 \\
MC-SJD & 0.0524 & -0.3417 & -0.1294 & 0.5800 & 0.5937 & 0.3158 & 0.4560 \\
SCD & -0.0329 & -0.2862 & -0.1003 & 0.6084 & 0.5291 & 0.3227 & 0.4335 \\
SJD & -0.0274 & -0.3041 & -0.0890 & 0.6097 & 0.5967 & 0.3065 & 0.5049 \\
PathSpec & 0.0136 & -0.3186 & -0.0780 & 0.5819 & 0.6344 & 0.3320 & 0.4397 \\
SJD-PAC & 0.1101 & -0.2792 & -0.1338 & 0.6186 & 0.6185 & 0.3166 & 0.4819 \\
\midrule
\multicolumn{8}{l}{\textit{Anole-7B}} \\
\multicolumn{8}{l}{\textit{\textbf{Training-based Methods}}} \\
Target-only & -0.9075 & -1.1151 & -1.1089 & -0.6595 & -0.3110 & -0.6489 & -0.8265 \\
ADT-Tree + LANTERN & -1.0804 & -1.0515 & -1.2386 & -0.6763 & -0.4030 & -0.6143 & -0.9236 \\
\multicolumn{8}{l}{\textit{\textbf{Training-free Methods}}} \\
Target-only & -0.8463 & -0.8625 & -0.9248 & -0.4745 & -0.1427 & -0.6702 & -0.7225 \\
SJD & -0.7562 & -0.8673 & -0.9591 & -0.5051 & -0.0866 & -0.7538 & -0.7577 \\
\midrule
\multicolumn{8}{l}{\textit{Emu3}} \\
\multicolumn{8}{l}{\textit{\textbf{Training-free Methods}}} \\
Target-only & -0.1009 & -0.2867 & -0.1430 & 0.7134 & 0.4104 & 0.4305 & 0.4880 \\
SJD & -0.0531 & -0.2545 & -0.0832 & 0.7090 & 0.4616 & 0.3863 & 0.4730 \\
\midrule
\multicolumn{8}{l}{\cellcolor{T2IGray}\textbf{Diffusion/flow}} \\
\multicolumn{8}{l}{\textit{FLUX.1-schnell}} \\
\multicolumn{8}{l}{\textit{\textbf{Training-free Methods}}} \\
Target-only & 1.1455 & 1.0348 & 0.9611 & 1.2004 & 1.1411 & 0.4998 & 1.2270 \\
TaylorSeer & 1.1808 & 1.0676 & 0.9812 & 1.2152 & 1.1178 & 0.6078 & 1.2395 \\
\midrule
\multicolumn{8}{l}{\textit{FLUX.1-dev}} \\
\multicolumn{8}{l}{\textit{\textbf{Training-free Methods}}} \\
Target-only & 0.9895 & 0.8839 & 0.7536 & 0.9261 & 1.1276 & 0.4328 & 0.9850 \\
ClusCa & 0.9669 & 0.8091 & 0.7414 & 0.8884 & 1.0932 & 0.4509 & 0.9864 \\
SpeCa & 0.9922 & 0.7319 & 0.8018 & 0.9699 & 1.1322 & 0.5010 & 1.0216 \\
\multicolumn{8}{l}{\textit{\textbf{SDE adaptations}}} \\
Target-only & -2.2766 & -2.2777 & -2.2743 & -2.2813 & -2.2746 & -2.2781 & -2.2780 \\
ASD & -2.2765 & -2.2771 & -2.2739 & -2.2813 & -2.2751 & -2.2787 & -2.2777 \\
Block verification & -2.2763 & -2.2778 & -2.2731 & -2.2813 & -2.2746 & -2.2783 & -2.2774 \\
Diffusion SD & -2.2767 & -2.2775 & -2.2740 & -2.2812 & -2.2752 & -2.2788 & -2.2777 \\
\bottomrule
\end{tabular}
\end{table*}

\begin{table*}[!tp]
\centering
\caption{Auxiliary analysis. SDE rows use their $h=8$ runs. HPS (human preference score) by topic ($\uparrow$), grouped by generation family and target model. Each row covers 200 prompts per topic; each group's target appears once. Values are recorded arm means.}
\label{tab:compact-topic-hps}
\label{tab:compact-topic-ar-hps}
\label{tab:compact-topic-diffusion-hps}
\scriptsize
\setlength{\tabcolsep}{4pt}
\renewcommand{\arraystretch}{1.10}
\begin{tabular}{@{}lrrrrrrr@{}}
\toprule
Method & Color & Text & Persp. & Count & Motion & Texture & Common. \\
\midrule
\multicolumn{8}{l}{\cellcolor{T2IGray}\textbf{Autoregressive}} \\
\multicolumn{8}{l}{\textit{Janus-Pro-7B, $384^2$}} \\
\multicolumn{8}{l}{\textit{\textbf{Training-free Methods}}} \\
Target-only & 0.2811 & 0.2121 & 0.2508 & 0.2388 & 0.2499 & 0.2458 & 0.2856 \\
CSD & 0.2803 & 0.2095 & 0.2508 & 0.2390 & 0.2482 & 0.2437 & 0.2830 \\
SJD & 0.2093 & 0.1612 & 0.1903 & 0.1969 & 0.2046 & 0.1959 & 0.2241 \\
\midrule
\multicolumn{8}{l}{\textit{LlamaGen Stage I, $256^2$}} \\
\multicolumn{8}{l}{\textit{\textbf{Training-based Methods}}} \\
Target-only & 0.1470 & 0.1263 & 0.1347 & 0.1374 & 0.1511 & 0.1393 & 0.1434 \\
LANTERN & 0.1625 & 0.1354 & 0.1572 & 0.1422 & 0.1684 & 0.1573 & 0.1564 \\
LANTERN++ & 0.1443 & 0.1216 & 0.1362 & 0.1319 & 0.1331 & 0.1435 & 0.1300 \\
VVS & 0.1573 & 0.1345 & 0.1522 & 0.1412 & 0.1620 & 0.1521 & 0.1514 \\
\midrule
\multicolumn{8}{l}{\textit{LlamaGen Stage II, $512^2$}} \\
\multicolumn{8}{l}{\textit{\textbf{Training-based Methods}}} \\
Target-only & 0.2102 & 0.1717 & 0.2052 & 0.2173 & 0.2056 & 0.1889 & 0.2389 \\
LANTERN & 0.2121 & 0.1675 & 0.2040 & 0.2161 & 0.2057 & 0.1903 & 0.2378 \\
LANTERN++ & 0.1739 & 0.1411 & 0.1656 & 0.1777 & 0.1604 & 0.1511 & 0.1927 \\
\multicolumn{8}{l}{\textit{\textbf{Training-free Methods}}} \\
Target-only & 0.2114 & 0.1659 & 0.2049 & 0.2141 & 0.2033 & 0.1930 & 0.2426 \\
ARsample filtering & 0.2068 & 0.1662 & 0.2043 & 0.2102 & 0.1973 & 0.1950 & 0.2373 \\
SJD & 0.2108 & 0.1726 & 0.2055 & 0.2164 & 0.2066 & 0.1884 & 0.2401 \\
\midrule
\multicolumn{8}{l}{\textit{Lumina-mGPT-7B, $768^2$}} \\
\multicolumn{8}{l}{\textit{\textbf{Training-free Methods}}} \\
Target-only & 0.2832 & 0.2179 & 0.2569 & 0.2704 & 0.2670 & 0.2570 & 0.2840 \\
GSD & 0.2865 & 0.2229 & 0.2658 & 0.2733 & 0.2689 & 0.2651 & 0.2850 \\
MC-SJD & 0.2884 & 0.2240 & 0.2657 & 0.2757 & 0.2704 & 0.2635 & 0.2875 \\
SCD & 0.2890 & 0.2260 & 0.2668 & 0.2758 & 0.2704 & 0.2641 & 0.2849 \\
SJD & 0.2865 & 0.2253 & 0.2678 & 0.2757 & 0.2677 & 0.2662 & 0.2864 \\
PathSpec & 0.2879 & 0.2220 & 0.2654 & 0.2767 & 0.2742 & 0.2635 & 0.2861 \\
SJD-PAC & 0.2881 & 0.2247 & 0.2666 & 0.2759 & 0.2767 & 0.2654 & 0.2856 \\
\midrule
\multicolumn{8}{l}{\textit{Anole-7B}} \\
\multicolumn{8}{l}{\textit{\textbf{Training-based Methods}}} \\
Target-only & 0.2201 & 0.1708 & 0.2049 & 0.2045 & 0.2018 & 0.2009 & 0.2041 \\
ADT-Tree + LANTERN & 0.2228 & 0.1727 & 0.2057 & 0.2051 & 0.1979 & 0.2055 & 0.2082 \\
\multicolumn{8}{l}{\textit{\textbf{Training-free Methods}}} \\
Target-only & 0.2299 & 0.1869 & 0.2144 & 0.2131 & 0.2213 & 0.2149 & 0.2118 \\
SJD & 0.2330 & 0.1842 & 0.2138 & 0.2148 & 0.2239 & 0.2144 & 0.2098 \\
\midrule
\multicolumn{8}{l}{\textit{Emu3}} \\
\multicolumn{8}{l}{\textit{\textbf{Training-free Methods}}} \\
Target-only & 0.2873 & 0.2302 & 0.2709 & 0.2832 & 0.2620 & 0.2789 & 0.2854 \\
SJD & 0.2864 & 0.2313 & 0.2719 & 0.2820 & 0.2620 & 0.2786 & 0.2860 \\
\midrule
\multicolumn{8}{l}{\cellcolor{T2IGray}\textbf{Diffusion/flow}} \\
\multicolumn{8}{l}{\textit{FLUX.1-schnell}} \\
\multicolumn{8}{l}{\textit{\textbf{Training-free Methods}}} \\
Target-only & 0.3024 & 0.2714 & 0.2778 & 0.2808 & 0.2724 & 0.2579 & 0.3131 \\
TaylorSeer & 0.3035 & 0.2710 & 0.2766 & 0.2820 & 0.2751 & 0.2599 & 0.3121 \\
\midrule
\multicolumn{8}{l}{\textit{FLUX.1-dev}} \\
\multicolumn{8}{l}{\textit{\textbf{Training-free Methods}}} \\
Target-only & 0.3051 & 0.2695 & 0.2716 & 0.2852 & 0.2940 & 0.2691 & 0.3123 \\
ClusCa & 0.2990 & 0.2581 & 0.2691 & 0.2799 & 0.2916 & 0.2664 & 0.3114 \\
SpeCa & 0.3013 & 0.2581 & 0.2712 & 0.2807 & 0.2947 & 0.2687 & 0.3122 \\
\multicolumn{8}{l}{\textit{\textbf{SDE adaptations}}} \\
Target-only & 0.0537 & 0.0350 & 0.0356 & 0.0527 & 0.0406 & 0.0638 & 0.0420 \\
ASD & 0.0540 & 0.0346 & 0.0356 & 0.0537 & 0.0417 & 0.0669 & 0.0429 \\
Block verification & 0.0547 & 0.0352 & 0.0372 & 0.0544 & 0.0419 & 0.0654 & 0.0440 \\
Diffusion SD & 0.0540 & 0.0345 & 0.0356 & 0.0534 & 0.0417 & 0.0667 & 0.0429 \\
\bottomrule
\end{tabular}
\end{table*}

\begin{table*}[!tp]
\centering
\caption{Auxiliary analysis. SDE rows use their $h=8$ runs. Ancillary Overall quality over 1,400 prompts, with the stated quality references. SigLIP measures alignment, Aesth. is the aesthetic score, $Q_2$ the criterion score of the second topic evaluator, and Safety the recorded automatic status score. Higher values are better in every column.}
\label{tab:compact-ancillary}
\label{tab:compact-ancillary-core}
\label{tab:compact-ancillary-ar}
\label{tab:compact-ancillary-diffusion}
\scriptsize
\setlength{\tabcolsep}{4pt}
\renewcommand{\arraystretch}{1.10}
\begin{tabular}{@{}lrrrr@{}}
\toprule
Method & SigLIP$\uparrow$ & Aesth.$\uparrow$ & $Q_2\uparrow$ & Safety$\uparrow$ \\
\midrule
\multicolumn{5}{l}{\cellcolor{T2IGray}\textbf{Autoregressive}} \\
\multicolumn{5}{l}{\textit{Janus-Pro-7B, $384^2$}} \\
\multicolumn{5}{l}{\textit{\textbf{Training-free Methods}}} \\
Target-only & 0.9603 & 5.3618 & 0.8406 & 0.9879 \\
CSD & 0.9565 & 5.3552 & 0.8429 & 0.9864 \\
SJD & 0.5791 & 5.1897 & 0.6449 & 0.9814 \\
\midrule
\multicolumn{5}{l}{\textit{LlamaGen Stage I, $256^2$}} \\
\multicolumn{5}{l}{\textit{\textbf{Training-based Methods}}} \\
Target-only & 0.1691 & 4.1825 & 0.3631 & 0.9443 \\
LANTERN & 0.2160 & 4.1781 & 0.3985 & 0.9193 \\
LANTERN++ & 0.1353 & 4.0080 & 0.3386 & 0.9150 \\
VVS & 0.2216 & 4.0818 & 0.4074 & 0.9400 \\
\midrule
\multicolumn{5}{l}{\textit{LlamaGen Stage II, $512^2$}} \\
\multicolumn{5}{l}{\textit{\textbf{Training-based Methods}}} \\
Target-only & 0.7526 & 5.1328 & 0.7304 & 0.9886 \\
LANTERN & 0.7265 & 5.1508 & 0.7377 & 0.9921 \\
LANTERN++ & 0.5950 & 4.6181 & 0.6117 & 0.9736 \\
\multicolumn{5}{l}{\textit{\textbf{Training-free Methods}}} \\
Target-only & 0.7288 & 5.1696 & 0.7218 & 0.9957 \\
ARsample filtering & 0.6934 & 5.1953 & 0.7320 & 0.9964 \\
SJD & 0.7564 & 5.1561 & 0.7162 & 0.9836 \\
\midrule
\multicolumn{5}{l}{\textit{Lumina-mGPT-7B, $768^2$}} \\
\multicolumn{5}{l}{\textit{\textbf{Training-free Methods}}} \\
Target-only & 0.9236 & 5.5814 & 0.7796 & 0.9979 \\
GSD & 0.9019 & 5.6548 & 0.7870 & 0.9971 \\
MC-SJD & 0.9090 & 5.6445 & 0.7783 & 0.9971 \\
SCD & 0.9030 & 5.6480 & 0.7708 & 0.9993 \\
SJD & 0.9079 & 5.6516 & 0.7783 & 0.9979 \\
PathSpec & 0.9026 & 5.6642 & 0.7654 & 0.9993 \\
SJD-PAC & 0.9073 & 5.6703 & 0.7893 & 0.9993 \\
\midrule
\multicolumn{5}{l}{\textit{Anole-7B}} \\
\multicolumn{5}{l}{\textit{\textbf{Training-based Methods}}} \\
Target-only & 0.7058 & 4.8973 & 0.6436 & 0.9814 \\
ADT-Tree + LANTERN & 0.6619 & 4.7819 & 0.6325 & 0.9657 \\
\multicolumn{5}{l}{\textit{\textbf{Training-free Methods}}} \\
Target-only & 0.7728 & 4.9835 & 0.6875 & 0.9836 \\
SJD & 0.7666 & 4.9827 & 0.6661 & 0.9821 \\
\midrule
\multicolumn{5}{l}{\textit{Emu3}} \\
\multicolumn{5}{l}{\textit{\textbf{Training-free Methods}}} \\
Target-only & 0.9082 & 5.7033 & 0.8162 & 0.9993 \\
SJD & 0.9159 & 5.7129 & 0.8196 & 0.9986 \\
\midrule
\multicolumn{5}{l}{\cellcolor{T2IGray}\textbf{Diffusion/flow}} \\
\multicolumn{5}{l}{\textit{FLUX.1-schnell}} \\
\multicolumn{5}{l}{\textit{\textbf{Training-free Methods}}} \\
Target-only & 0.9507 & 5.6547 & 0.9292 & 0.9929 \\
TaylorSeer & 0.9659 & 5.5821 & 0.9252 & 0.9936 \\
\midrule
\multicolumn{5}{l}{\textit{FLUX.1-dev}} \\
\multicolumn{5}{l}{\textit{\textbf{Training-free Methods}}} \\
Target-only & 0.9161 & 5.7737 & 0.9094 & 0.9986 \\
ClusCa & 0.9263 & 5.7333 & 0.9086 & 0.9979 \\
SpeCa & 0.9175 & 5.7563 & 0.9104 & 0.9971 \\
\multicolumn{5}{l}{\textit{\textbf{SDE adaptations}}} \\
Target-only & 0.0001 & 4.2744 & 0.1518 & 0.3150 \\
ASD & 0.0001 & 4.2576 & 0.1530 & 0.3443 \\
Block verification & 0.0001 & 4.2565 & 0.1523 & 0.2914 \\
Diffusion SD & 0.0001 & 4.2552 & 0.1524 & 0.3464 \\
\bottomrule
\end{tabular}
\end{table*}

\begin{table*}[!tp]
\centering
\caption{Auxiliary analysis. SDE rows use their $h=8$ runs. Easy/Hard comparison, with 700 prompts per stratum. $Q$ is the topic criterion and $S$ the target/method latency ratio. Core rows use median paired ratios of warm times, and other rows use ratios of mean recorded times. Quality uses the stated references; timing retains each run's matched target.}
\label{tab:consolidated-difficulty}
\label{tab:consolidated-difficulty-ar}
\label{tab:consolidated-difficulty-diffusion}
\scriptsize
\setlength{\tabcolsep}{4pt}
\renewcommand{\arraystretch}{1.10}
\begin{tabular}{@{}lrrrrr@{}}
\toprule
Method / reference & $Q_E$ & $Q_H$ & $Q_H-Q_E$ & $S_E$ & $S_H$ \\
\midrule
\multicolumn{6}{l}{\cellcolor{T2IGray}\textbf{Autoregressive}} \\
\multicolumn{6}{l}{\textit{Janus-Pro-7B, $384^2$}} \\
\multicolumn{6}{l}{\textit{\textbf{Training-free Methods}}} \\
Target-only & 0.8236 & 0.8536 & +0.0300 & 1.000 & 1.000 \\
CSD & 0.8417 & 0.8610 & +0.0193 & 1.742 & 1.691 \\
SJD & 0.7360 & 0.7221 & -0.0138 & 1.356 & 1.346 \\
\midrule
\multicolumn{6}{l}{\textit{LlamaGen Stage I, $256^2$}} \\
\multicolumn{6}{l}{\textit{\textbf{Training-based Methods}}} \\
Target-only & 0.6193 & 0.5802 & -0.0390 & 1.000 & 1.000 \\
LANTERN & 0.6752 & 0.6240 & -0.0512 & 1.089 & 1.081 \\
LANTERN++ & 0.6214 & 0.5457 & -0.0757 & 3.698 & 3.521 \\
VVS & 0.6769 & 0.6540 & -0.0229 & 1.030 & 1.037 \\
\midrule
\multicolumn{6}{l}{\textit{LlamaGen Stage II, $512^2$}} \\
\multicolumn{6}{l}{\textit{\textbf{Training-based Methods}}} \\
Target-only & 0.7450 & 0.7445 & -0.0005 & 1.000 & 1.000 \\
LANTERN & 0.7286 & 0.7571 & +0.0286 & 1.001 & 1.002 \\
LANTERN++ & 0.6648 & 0.6674 & +0.0026 & 3.764 & 3.748 \\
\multicolumn{6}{l}{\textit{\textbf{Training-free Methods}}} \\
Target-only & 0.7088 & 0.7650 & +0.0562 & 1.000 & 1.000 \\
ARsample filtering & 0.7217 & 0.7521 & +0.0305 & 0.969 & 0.969 \\
SJD & 0.7219 & 0.7502 & +0.0283 & 1.582 & 1.554 \\
\midrule
\multicolumn{6}{l}{\textit{Lumina-mGPT-7B, $768^2$}} \\
\multicolumn{6}{l}{\textit{\textbf{Training-free Methods}}} \\
Target-only & 0.7460 & 0.7805 & +0.0345 & 1.000 & 1.000 \\
GSD & 0.7457 & 0.7802 & +0.0345 & 4.902 & 4.828 \\
MC-SJD & 0.7555 & 0.7907 & +0.0352 & 4.716 & 4.695 \\
SCD & 0.7367 & 0.7700 & +0.0333 & 5.179 & 5.226 \\
SJD & 0.7417 & 0.7767 & +0.0350 & 3.482 & 3.537 \\
PathSpec & 0.7450 & 0.7712 & +0.0262 & 3.346 & 3.324 \\
SJD-PAC & 0.7405 & 0.7793 & +0.0388 & 3.194 & 3.211 \\
\midrule
\multicolumn{6}{l}{\textit{Anole-7B}} \\
\multicolumn{6}{l}{\textit{\textbf{Training-based Methods}}} \\
Target-only & 0.6831 & 0.6619 & -0.0212 & 1.000 & 1.000 \\
ADT-Tree + LANTERN & 0.6786 & 0.6857 & +0.0071 & 1.099 & 1.079 \\
\multicolumn{6}{l}{\textit{\textbf{Training-free Methods}}} \\
Target-only & 0.6121 & 0.6974 & +0.0852 & 1.000 & 1.000 \\
SJD & 0.6102 & 0.7002 & +0.0900 & 1.012 & 0.985 \\
\midrule
\multicolumn{6}{l}{\textit{Emu3}} \\
\multicolumn{6}{l}{\textit{\textbf{Training-free Methods}}} \\
Target-only & 0.6990 & 0.7750 & +0.0760 & 1.000 & 1.000 \\
SJD & 0.7045 & 0.7781 & +0.0736 & 1.020 & 1.030 \\
\midrule
\multicolumn{6}{l}{\cellcolor{T2IGray}\textbf{Diffusion/flow}} \\
\multicolumn{6}{l}{\textit{FLUX.1-schnell}} \\
\multicolumn{6}{l}{\textit{\textbf{Training-free Methods}}} \\
Target-only & 0.8831 & 0.9014 & +0.0183 & 1.000 & 1.000 \\
TaylorSeer & 0.8960 & 0.9071 & +0.0112 & 1.474 & 1.479 \\
\midrule
\multicolumn{6}{l}{\textit{FLUX.1-dev}} \\
\multicolumn{6}{l}{\textit{\textbf{Training-free Methods}}} \\
Target-only & 0.8650 & 0.8850 & +0.0200 & 1.000 & 1.000 \\
ClusCa & 0.8598 & 0.8898 & +0.0300 & 3.550 & 3.598 \\
SpeCa & 0.8729 & 0.8845 & +0.0117 & 4.627 & 4.740 \\
\multicolumn{6}{l}{\textit{\textbf{SDE adaptations}}} \\
Target-only & 0.1052 & 0.1167 & +0.0114 & 1.000 & 1.000 \\
ASD & 0.0988 & 0.1076 & +0.0088 & 0.146 & 0.146 \\
Block verification & 0.0967 & 0.1055 & +0.0088 & 0.252 & 0.252 \\
Diffusion SD & 0.1050 & 0.1119 & +0.0069 & 0.148 & 0.147 \\
\bottomrule
\end{tabular}
\end{table*}

\begin{table*}[!tp]
\centering
\caption{Auxiliary analysis. Overall pointwise 95\% percentile interval bounds for the core AR rows. Quality changes are method minus target, and $S$ is the median paired ratio of warm times. The 2,000 bootstrap draws resample connected groups of prompts that share a scene or creative seed within each topic, retaining their prompts and matched target/method records.}
\label{tab:core-confidence}
\small
\setlength{\tabcolsep}{4pt}
\renewcommand{\arraystretch}{1.10}
\resizebox{\ifdim\width>\linewidth\linewidth\else\width\fi}{!}{%
\begin{tabular}{@{}lrrrrr@{}}
\toprule
Method & $S$ & $\Delta$CLIP & $\Delta$ImageReward & $\Delta$HPS & $\Delta Q$ \\
\midrule
\multicolumn{6}{l}{\textit{Janus-Pro-7B, $384^2$}} \\
CSD & [1.580, 1.855] & [-0.0032, 0.0039] & [-0.0434, 0.0159] & [-0.0024, -0.0004] & [-0.0027, 0.0276] \\
\multicolumn{6}{l}{\textit{LlamaGen Stage I, $256^2$}} \\
LANTERN & [1.078, 1.093] & [0.0146, 0.0243] & [0.1170, 0.1746] & [0.0126, 0.0160] & [0.0252, 0.0748] \\
LANTERN++ & [3.530, 3.688] & [-0.0078, 0.0022] & [-0.0581, 0.0018] & [-0.0073, -0.0038] & [-0.0432, 0.0105] \\
\multicolumn{6}{l}{\textit{LlamaGen Stage II, $512^2$}} \\
LANTERN & [0.998, 1.003] & [-0.0145, -0.0065] & [-0.1216, -0.0420] & [-0.0021, 0.0010] & [-0.0216, 0.0176] \\
LANTERN++ & [3.745, 3.767] & [-0.0399, -0.0306] & [-0.5517, -0.4715] & [-0.0409, -0.0376] & [-0.0989, -0.0582] \\
\multicolumn{6}{l}{\textit{Lumina-mGPT-7B, $768^2$}} \\
GSD & [4.776, 4.922] & [-0.0162, -0.0084] & [-0.0632, 0.0195] & [0.0030, 0.0059] & [-0.0161, 0.0166] \\
MC-SJD & [4.624, 4.765] & [-0.0106, -0.0033] & [-0.0314, 0.0475] & [0.0040, 0.0070] & [-0.0073, 0.0280] \\
SCD & [5.107, 5.283] & [-0.0109, -0.0034] & [-0.0392, 0.0381] & [0.0044, 0.0072] & [-0.0271, 0.0079] \\
SJD & [3.440, 3.563] & [-0.0111, -0.0035] & [-0.0178, 0.0557] & [0.0042, 0.0070] & [-0.0214, 0.0137] \\
\bottomrule
\end{tabular}%
}
\end{table*}

\begin{table*}[!tp]
\centering
\caption{Auxiliary analysis. Pointwise 95\% cluster interval bounds for (a) criterion changes, method minus target, and (b) median paired target/method latency ratios. Each topic contains 200 prompts; resampling preserves the groups of prompts that share a scene or creative seed, and their matched records.}
\label{tab:topic-confidence}\label{tab:topic-confidence-quality}\label{tab:topic-confidence-speed}
\scriptsize
\setlength{\tabcolsep}{4pt}
\renewcommand{\arraystretch}{1.10}
\begin{tabular}{@{}lrrrrrrr@{}}
\toprule
Method & Color & Text & Persp. & Count & Motion & Texture & Common. \\
\midrule
\multicolumn{8}{l}{\cellcolor{T2IGray}\textbf{(a) Topic criterion change $\Delta Q$}} \\
\multicolumn{8}{l}{\textit{Janus-Pro-7B, $384^2$}} \\
CSD & [-0.012, 0.005] & [-0.018, 0.101] & [-0.004, 0.011] & [-0.056, 0.090] & [-0.014, 0.053] & [-0.023, 0.003] & [-0.008, 0.039] \\
\multicolumn{8}{l}{\textit{LlamaGen Stage I, $256^2$}} \\
LANTERN & [-0.012, 0.078] & [-0.163, -0.024] & [0.143, 0.272] & [-0.079, 0.078] & [0.016, 0.168] & [0.023, 0.138] & [-0.016, 0.074] \\
LANTERN++ & [-0.184, -0.081] & [-0.114, 0.042] & [0.077, 0.222] & [-0.151, -0.010] & [-0.122, 0.033] & [0.014, 0.140] & [-0.097, -0.002] \\
\multicolumn{8}{l}{\textit{LlamaGen Stage II, $512^2$}} \\
LANTERN & [-0.042, 0.023] & [-0.157, -0.002] & [-0.030, 0.023] & [-0.093, 0.047] & [0.048, 0.176] & [-0.013, 0.033] & [-0.053, 0.013] \\
LANTERN++ & [-0.109, -0.039] & [-0.030, 0.110] & [-0.053, 0.000] & [-0.293, -0.147] & [-0.087, 0.042] & [-0.203, -0.115] & [-0.128, -0.061] \\
\multicolumn{8}{l}{\textit{Lumina-mGPT-7B, $768^2$}} \\
GSD & [-0.025, 0.015] & [-0.087, 0.048] & [-0.016, 0.008] & [-0.098, 0.038] & [0.032, 0.112] & [-0.005, 0.016] & [-0.053, 0.014] \\
MC-SJD & [-0.008, 0.022] & [-0.044, 0.113] & [-0.044, 0.002] & [-0.077, 0.065] & [-0.009, 0.072] & [-0.004, 0.016] & [-0.011, 0.052] \\
SCD & [-0.027, 0.011] & [-0.050, 0.086] & [-0.029, 0.003] & [-0.152, 0.017] & [-0.016, 0.066] & [-0.005, 0.017] & [-0.064, 0.005] \\
SJD & [-0.034, 0.007] & [-0.058, 0.085] & [-0.036, 0.003] & [-0.074, 0.085] & [-0.042, 0.042] & [-0.007, 0.012] & [-0.049, 0.012] \\
\midrule
\multicolumn{8}{l}{\cellcolor{T2IGray}\textbf{(b) Median paired latency ratio $S$}} \\
\multicolumn{8}{l}{\textit{Janus-Pro-7B, $384^2$}} \\
CSD & [1.339, 1.951] & [1.433, 2.014] & [1.473, 2.085] & [1.407, 2.055] & [1.585, 2.325] & [1.639, 2.463] & [1.237, 1.788] \\
\multicolumn{8}{l}{\textit{LlamaGen Stage I, $256^2$}} \\
LANTERN & [1.035, 1.086] & [1.076, 1.134] & [1.047, 1.088] & [1.083, 1.122] & [1.060, 1.101] & [1.066, 1.107] & [1.077, 1.112] \\
LANTERN++ & [3.404, 3.917] & [3.514, 5.082] & [3.404, 3.527] & [3.589, 4.351] & [3.439, 4.019] & [3.508, 4.464] & [3.491, 3.811] \\
\multicolumn{8}{l}{\textit{LlamaGen Stage II, $512^2$}} \\
LANTERN & [0.982, 0.998] & [1.010, 1.023] & [0.997, 1.012] & [0.975, 0.987] & [1.003, 1.018] & [1.012, 1.029] & [0.988, 1.000] \\
LANTERN++ & [3.706, 3.744] & [3.772, 3.835] & [3.753, 3.791] & [3.742, 3.790] & [3.693, 3.767] & [3.768, 3.823] & [3.682, 3.723] \\
\multicolumn{8}{l}{\textit{Lumina-mGPT-7B, $768^2$}} \\
GSD & [4.532, 4.839] & [4.478, 4.827] & [4.863, 5.127] & [4.466, 4.816] & [4.741, 5.321] & [5.048, 5.674] & [4.770, 5.037] \\
MC-SJD & [4.309, 4.834] & [4.430, 4.720] & [4.483, 4.826] & [4.268, 4.639] & [4.709, 5.238] & [4.731, 5.200] & [4.559, 5.014] \\
SCD & [4.744, 5.374] & [4.892, 5.312] & [4.961, 5.299] & [4.618, 5.130] & [5.197, 5.865] & [5.178, 5.675] & [5.189, 5.560] \\
SJD & [3.241, 3.610] & [3.324, 3.547] & [3.567, 3.806] & [3.085, 3.427] & [3.316, 3.674] & [3.434, 3.742] & [3.293, 3.663] \\
\bottomrule
\end{tabular}
\end{table*}

\begin{table*}[!tp]
\centering
\caption{Parameter sensitivity on FLUX.1-dev, with 1,400 prompt pairs per setting. $h$ is the speculative horizon; $\tau$ is SpeCa's base threshold or ClusCa's freshness threshold. The first three methods use the evaluated SDE adaptations; their $h=8$ rows are the runs in the main tables. $t_T$ and $t_M$ are mean target and method request times in each setting's run; $S=t_T/t_M$. $Q_T$ and $Q_M$ are the scores of the group target and the method, respectively. The group target is the ClusCa target for SpeCa and ClusCa and the target with a uniform schedule for the SDE adaptations; all quality changes subtract its scores. Panel (b) completes the eight scalar metrics and reports 700 pairs per construction stratum; $Q_2$ denotes the criterion score of the second topic evaluator. Panel (c) reports 200 pairs per topic.}
\label{tab:parameter-sweeps-overall}\label{tab:parameter-sweeps-topics}
\footnotesize
\setlength{\tabcolsep}{3pt}
\renewcommand{\arraystretch}{1.12}
\textbf{(a) Overall quality and request latency}\par\smallskip
\resizebox{\ifdim\width>\linewidth\linewidth\else\width\fi}{!}{%
\begin{tabular}{@{}llrrrrrrrrr@{}}
\toprule
Method & Setting & $t_T$ (s) & $t_M$ (s) & $S$ & $Q_T$ & $Q_M$ & $\Delta Q$ & $\Delta$CLIP & $\Delta$IR & $\Delta$HPS \\
\midrule
Diffusion SD & $h=2$ & 7.364 & 13.799 & 0.534 & 0.1110 & 0.1094 & -0.0015 & +0.0003 & -0.0001 & +0.0006 \\
 & $h=8$ & 6.439 & 43.662 & 0.147 & 0.1110 & 0.1085 & -0.0025 & -0.0028 & 0.0000 & +0.0008 \\
\addlinespace[3pt]
ASD & $h=2$ & 12.608 & 23.721 & 0.531 & 0.1110 & 0.1081 & -0.0029 & +0.0002 & -0.0001 & +0.0006 \\
 & $h=8$ & 12.605 & 86.320 & 0.146 & 0.1110 & 0.1032 & -0.0077 & -0.0022 & +0.0001 & +0.0008 \\
\addlinespace[3pt]
Block verification & $h=2$ & 12.610 & 18.433 & 0.684 & 0.1110 & 0.1088 & -0.0021 & +0.0036 & +0.0006 & +0.0007 \\
 & $h=8$ & 12.607 & 49.989 & 0.252 & 0.1110 & 0.1011 & -0.0099 & +0.0025 & +0.0003 & +0.0014 \\
\addlinespace[3pt]
SpeCa & $\tau=5$ & 12.732 & 2.972 & 4.284 & 0.8750 & 0.8793 & +0.0043 & +0.0029 & +0.0195 & -0.0006 \\
 & $\tau=9$ & 12.734 & 2.481 & 5.134 & 0.8750 & 0.8655 & -0.0095 & +0.0066 & -0.0450 & -0.0069 \\
\addlinespace[3pt]
ClusCa & $\tau=3$ & 12.729 & 4.464 & 2.852 & 0.8750 & 0.8802 & +0.0052 & +0.0014 & +0.0048 & -0.0007 \\
 & $\tau=7$ & 12.727 & 3.102 & 4.103 & 0.8750 & 0.8698 & -0.0052 & +0.0072 & -0.1075 & -0.0115 \\
\bottomrule
\end{tabular}%
}
\par\medskip\textbf{(b) Secondary metrics and construction strata}\par\smallskip
\resizebox{\ifdim\width>\linewidth\linewidth\else\width\fi}{!}{%
\begin{tabular}{@{}llrrrrrrrr@{}}
\toprule
\multicolumn{6}{c}{} & \multicolumn{2}{c}{Easy} & \multicolumn{2}{c}{Hard} \\
\cmidrule(lr){7-8}\cmidrule(lr){9-10}
Method & Setting & $\Delta$SigLIP & $\Delta$Aesth. & $\Delta Q_2$ & $\Delta$Safety & $\Delta Q$ & $S$ & $\Delta Q$ & $S$ \\
\midrule
Diffusion SD & $h=2$ & 0.0000 & -0.0491 & -0.0008 & -0.0379 & -0.0002 & 0.534 & -0.0029 & 0.534 \\
 & $h=8$ & 0.0000 & -0.0192 & +0.0006 & +0.0314 & -0.0002 & 0.148 & -0.0048 & 0.147 \\
\addlinespace[3pt]
ASD & $h=2$ & 0.0000 & -0.0468 & -0.0014 & -0.0336 & 0.0000 & 0.532 & -0.0057 & 0.531 \\
 & $h=8$ & 0.0000 & -0.0167 & +0.0012 & +0.0293 & -0.0064 & 0.146 & -0.0090 & 0.146 \\
\addlinespace[3pt]
Block verification & $h=2$ & 0.0000 & -0.0040 & +0.0004 & +0.0200 & -0.0040 & 0.684 & -0.0002 & 0.684 \\
 & $h=8$ & 0.0000 & -0.0179 & +0.0005 & -0.0236 & -0.0086 & 0.252 & -0.0112 & 0.252 \\
\addlinespace[3pt]
SpeCa & $\tau=5$ & -0.0008 & -0.0051 & +0.0004 & -0.0014 & +0.0074 & 4.199 & +0.0012 & 4.372 \\
 & $\tau=9$ & +0.0015 & -0.0578 & -0.0008 & -0.0043 & -0.0093 & 5.019 & -0.0098 & 5.254 \\
\addlinespace[3pt]
ClusCa & $\tau=3$ & +0.0019 & 0.0000 & -0.0023 & -0.0014 & -0.0010 & 2.823 & +0.0114 & 2.881 \\
 & $\tau=7$ & +0.0037 & -0.1221 & -0.0127 & -0.0014 & -0.0098 & 4.044 & -0.0007 & 4.163 \\
\bottomrule
\end{tabular}%
}
\par\medskip\textbf{(c) Quality and request latency across seven topics}\par\smallskip
\resizebox{\ifdim\width>\linewidth\linewidth\else\width\fi}{!}{%
\begin{tabular}{@{}llrrrrrrrrrrrrrr@{}}
\toprule
\multicolumn{2}{c}{} & \multicolumn{2}{c}{Color} & \multicolumn{2}{c}{Text} & \multicolumn{2}{c}{Persp.} & \multicolumn{2}{c}{Count} & \multicolumn{2}{c}{Motion} & \multicolumn{2}{c}{Texture} & \multicolumn{2}{c}{Common.} \\
\cmidrule(lr){3-4}\cmidrule(lr){5-6}\cmidrule(lr){7-8}\cmidrule(lr){9-10}\cmidrule(lr){11-12}\cmidrule(lr){13-14}\cmidrule(lr){15-16}
Method & Setting & $\Delta Q$ & $S$ & $\Delta Q$ & $S$ & $\Delta Q$ & $S$ & $\Delta Q$ & $S$ & $\Delta Q$ & $S$ & $\Delta Q$ & $S$ & $\Delta Q$ & $S$ \\
\midrule
Diffusion SD & $h=2$ & +0.0025 & 0.535 & 0.0000 & 0.533 & -0.0067 & 0.533 & -0.0125 & 0.533 & -0.0025 & 0.533 & +0.0117 & 0.536 & -0.0033 & 0.533 \\
 & $h=8$ & -0.0042 & 0.148 & 0.0000 & 0.148 & +0.0033 & 0.147 & -0.0217 & 0.147 & -0.0042 & 0.148 & +0.0092 & 0.148 & 0.0000 & 0.148 \\
\addlinespace[3pt]
ASD & $h=2$ & +0.0108 & 0.531 & 0.0000 & 0.531 & -0.0017 & 0.532 & -0.0258 & 0.533 & -0.0025 & 0.532 & +0.0050 & 0.532 & -0.0058 & 0.531 \\
 & $h=8$ & -0.0008 & 0.146 & 0.0000 & 0.146 & +0.0017 & 0.146 & -0.0267 & 0.146 & -0.0017 & 0.146 & -0.0267 & 0.146 & 0.0000 & 0.146 \\
\addlinespace[3pt]
Block verification & $h=2$ & -0.0058 & 0.684 & 0.0000 & 0.684 & +0.0017 & 0.684 & -0.0342 & 0.684 & -0.0025 & 0.684 & +0.0317 & 0.684 & -0.0058 & 0.684 \\
 & $h=8$ & -0.0067 & 0.252 & 0.0000 & 0.252 & 0.0000 & 0.252 & -0.0350 & 0.252 & -0.0067 & 0.252 & -0.0208 & 0.252 & 0.0000 & 0.253 \\
\addlinespace[3pt]
SpeCa & $\tau=5$ & +0.0033 & 4.185 & -0.0167 & 4.292 & -0.0042 & 4.290 & +0.0275 & 4.277 & +0.0158 & 4.308 & -0.0058 & 4.327 & +0.0100 & 4.309 \\
 & $\tau=9$ & +0.0008 & 5.012 & -0.0717 & 5.132 & -0.0050 & 5.207 & -0.0108 & 5.103 & +0.0217 & 5.131 & -0.0117 & 5.138 & +0.0100 & 5.215 \\
\addlinespace[3pt]
ClusCa & $\tau=3$ & +0.0008 & 2.801 & +0.0383 & 2.848 & 0.0000 & 2.888 & -0.0167 & 2.846 & +0.0242 & 2.843 & -0.0033 & 2.888 & -0.0067 & 2.849 \\
 & $\tau=7$ & +0.0017 & 4.091 & -0.0517 & 4.078 & -0.0058 & 4.084 & -0.0200 & 4.089 & +0.0442 & 4.228 & -0.0033 & 4.056 & -0.0017 & 4.097 \\
\bottomrule
\end{tabular}%
}
\end{table*}

\onecolumn\twocolumn

\subsection{Absolute Latency and Target Work}
\label{app:latency-details}

\paragraph{Latency profiles by topic.}
Figure~\ref{fig:topic-speed-radar} uses the same runs as the main
comparisons. The figure shows eight AR configurations and all six diffusion
configurations,
and Table~\ref{tab:topic-spread-selected} lists all 24 configurations.
Every topic point and Overall value uses the reported run and its matched
reference. The radar plots the percentage deviation of each topic ratio from
the method's own Overall ratio, whereas the tables report absolute
target/method ratios. Quality in the main tables instead uses the fixed
references described above.

Table~\ref{tab:compact-absolute-latency} collects the Overall mean and median
times of the auxiliary runs, grouped by generation family and target model.
Each method occupies one row, with target-only and method times in adjacent
columns; the SDE rows use their $h=8$ runs. The scope column
distinguishes warm time, request time, and elapsed runner time. Each core warm measurement
is the median of three timed repeats. ADT-Tree and ARsample produce 1,024 output tokens in every
pair, whereas PathSpec's method outputs contain 2,357--2,506 units, with
equal target and method lengths in six of 1,400 pairs.

\paragraph{Timing semantics and tail behavior.}
Component fields in Table~\ref{tab:latency-components} come from the
representative repeat whose warm elapsed
time is the median of the three recorded repeats. Prefill can include the
first target forward, which is also counted in generation or verification,
and the Lumina interval for generating visual tokens and its speculative
timers for the full generation use different endpoints.
The residual controller interval includes unassigned runtime work. Time for the whole image can also fall while token
availability becomes less regular: Janus CSD reduces the mean number of
target forward passes from 576 to 192.07 but raises the 95th percentile of
inter-token latency (ITL), averaged over outputs, from 26.48 to 47.05~ms.

\begin{table*}[t]
\centering
\caption{Latency ratios by topic and relative spread. Ratios $S_g$ divide mean target time by mean method time over 200 topic prompts; Overall pools 1,400 pairs. Spread is $100(\max_g S_g/\min_g S_g-1)$; bold marks each row's minimum and maximum topic ratio. Evaluation settings follow Section~\ref{sec:setup}.}
\label{tab:topic-spread-selected}
\setlength{\tabcolsep}{3.2pt}
\renewcommand{\arraystretch}{1.08}
\scriptsize
\resizebox{\textwidth}{!}{%
\begin{tabular}{llrrrrrrrrr}
\toprule
Method & Target & Color & Text & Persp. & Count & Motion & Texture & Common. & Overall $S$ & Spread (\%) \\
\midrule
\multicolumn{11}{l}{\textit{Autoregressive}} \\
CSD & Janus-Pro-7B $384^2$ & \textbf{1.5617} & 1.6966 & 1.7178 & 1.6677 & 1.8441 & \textbf{1.9846} & 1.5629 & 1.7101 & 27.08 \\
SJD & Janus-Pro-7B $384^2$ & \textbf{1.3194} & 1.3643 & 1.3587 & 1.3239 & \textbf{1.4011} & 1.3654 & 1.3207 & 1.3508 & 6.19 \\
LANTERN & LlamaGen $256^2$ & 1.0789 & \textbf{1.1719} & \textbf{1.0630} & 1.1540 & 1.0880 & 1.1223 & 1.0692 & 1.1063 & 10.25 \\
LANTERN++ & LlamaGen $256^2$ & 4.4682 & 4.5694 & \textbf{4.2519} & \textbf{4.8069} & 4.4900 & 4.6403 & 4.4377 & 4.5219 & 13.05 \\
VVS & LlamaGen $256^2$ & \textbf{0.9686} & 1.0441 & 1.0460 & \textbf{1.0573} & 1.0391 & 1.0414 & 1.0474 & 1.0337 & 9.16 \\
LANTERN & LlamaGen $512^2$ & \textbf{0.8729} & 0.9437 & 0.8823 & 0.8843 & \textbf{0.9474} & 0.9043 & 0.9050 & 0.9052 & 8.54 \\
LANTERN++ & LlamaGen $512^2$ & 3.1827 & \textbf{3.4168} & 3.2569 & 3.1763 & \textbf{3.0980} & 3.2127 & 3.1889 & 3.2163 & 10.29 \\
ARsample filtering & LlamaGen $512^2$ & 0.9936 & 0.9939 & 0.9954 & \textbf{0.9925} & 0.9932 & 0.9942 & \textbf{1.0018} & 0.9949 & 0.94 \\
SJD & LlamaGen $512^2$ & 1.5743 & 1.5920 & 1.5781 & 1.5231 & 1.5527 & \textbf{1.6435} & \textbf{1.5212} & 1.5682 & 8.04 \\
GSD & Lumina $768^2$ & 4.9045 & 4.8684 & 5.2521 & \textbf{4.7079} & 5.1133 & \textbf{5.4098} & 4.9343 & 5.0195 & 14.91 \\
MC-SJD & Lumina $768^2$ & 4.6784 & 4.5152 & 4.6543 & \textbf{4.4612} & 4.9853 & \textbf{4.9982} & 4.8180 & 4.7234 & 12.04 \\
SCD & Lumina $768^2$ & 5.2877 & 5.1965 & 5.1719 & \textbf{4.9875} & \textbf{5.5490} & 5.4593 & 5.4135 & 5.2901 & 11.26 \\
SJD & Lumina $768^2$ & 3.5779 & 3.5406 & \textbf{3.7742} & \textbf{3.4338} & 3.6028 & 3.6373 & 3.6093 & 3.5953 & 9.91 \\
PathSpec & Lumina $768^2$ & \textbf{2.8283} & 2.8415 & 2.8957 & 3.0729 & 3.4465 & 3.8002 & \textbf{3.8270} & 3.3348 & 35.31 \\
SJD-PAC & Lumina $768^2$ & \textbf{3.0006} & 3.2114 & 3.2752 & 3.1647 & 3.2062 & \textbf{3.3030} & 3.2057 & 3.2031 & 10.08 \\
ADT-Tree + LANTERN & Anole-7B & 1.1368 & \textbf{1.1585} & 1.1375 & \textbf{0.9011} & 1.0264 & 1.1202 & 1.0284 & 1.0889 & 28.56 \\
SJD & Anole-7B & 1.0630 & 1.0718 & 1.0745 & \textbf{1.0794} & \textbf{0.8633} & 0.8708 & 0.9631 & 0.9983 & 25.03 \\
SJD & Emu3 & 1.0161 & 1.0035 & 1.0478 & \textbf{1.0531} & \textbf{0.9539} & 1.0500 & 1.0497 & 1.0249 & 10.41 \\
\midrule
\multicolumn{11}{l}{\textit{Diffusion/flow}} \\
TaylorSeer & FLUX.1-schnell & \textbf{1.6279} & 1.6181 & 1.6117 & 1.6112 & 1.6128 & \textbf{1.6105} & 1.6121 & 1.6149 & 1.09 \\
ClusCa & FLUX.1-dev & \textbf{3.6948} & 3.5110 & 3.6808 & 3.4866 & 3.4975 & \textbf{3.4741} & 3.6852 & 3.5739 & 6.35 \\
SpeCa & FLUX.1-dev & 4.6075 & \textbf{4.8079} & \textbf{4.5786} & 4.6157 & 4.6395 & 4.7813 & 4.7533 & 4.6825 & 5.01 \\
ASD & FLUX.1-dev & \textbf{0.1458} & 0.1458 & \textbf{0.1462} & 0.1461 & 0.1459 & 0.1460 & 0.1462 & 0.1460 & 0.28 \\
Block verification & FLUX.1-dev & 0.2520 & \textbf{0.2518} & 0.2524 & 0.2523 & 0.2520 & 0.2521 & \textbf{0.2527} & 0.2522 & 0.35 \\
Diffusion SD & FLUX.1-dev & \textbf{0.1480} & 0.1476 & \textbf{0.1467} & 0.1473 & 0.1476 & 0.1478 & 0.1478 & 0.1475 & 0.89 \\
\bottomrule
\end{tabular}%
}
\vspace{2pt}
\begin{minipage}{\textwidth}\scriptsize SDE adaptations use their $h=8$ runs from Table~\ref{tab:parameter-sweeps-overall}. Core rows use warm times; other rows retain their recorded request/runner timers. PathSpec has unequal output lengths in 1,394/1,400 pairs.\end{minipage}
\end{table*}

\begin{table*}[!tp]
\centering
\caption{Auxiliary analysis. SDE rows use their $h=8$ runs. Absolute times in seconds per output over 1,400 prompts. Each method row lists its paired target-only and method measurements. Scope identifies the recorded timer: warm time, request time, or elapsed runner time. Diffusion/flow rows use request time.}
\label{tab:compact-absolute-latency}
\label{tab:compact-absolute-ar}
\label{tab:compact-absolute-diffusion}
\footnotesize
\setlength{\tabcolsep}{6pt}
\renewcommand{\arraystretch}{1.10}
\begin{tabularx}{\linewidth}{@{}>{\raggedright\arraybackslash}Xlrrrr@{}}
\toprule
Method & Scope & \multicolumn{2}{c}{Target (s)} & \multicolumn{2}{c}{Method (s)} \\
\cmidrule(lr){3-4}\cmidrule(lr){5-6}
 &  & Mean & Median & Mean & Median \\
\midrule
\multicolumn{6}{l}{\cellcolor{T2IGray}\textbf{Autoregressive}} \\
\multicolumn{6}{l}{\textit{Janus-Pro-7B, $384^2$}} \\
CSD & warm & 14.138 & 12.748 & 8.268 & 8.270 \\
SJD & request & 8.810 & 8.028 & 6.522 & 6.119 \\
\midrule
\multicolumn{6}{l}{\textit{LlamaGen Stage I, $256^2$}} \\
LANTERN & warm & 6.366 & 4.843 & 5.745 & 4.555 \\
LANTERN++ & warm & 6.366 & 4.843 & 1.412 & 1.421 \\
VVS & request & 3.889 & 3.872 & 3.762 & 3.698 \\
\midrule
\multicolumn{6}{l}{\textit{LlamaGen Stage II, $512^2$}} \\
LANTERN & warm & 17.456 & 16.665 & 19.283 & 16.676 \\
LANTERN++ & warm & 17.456 & 16.665 & 5.423 & 4.436 \\
ARsample filtering & request & 4.963 & 4.920 & 5.123 & 5.062 \\
SJD & request & 15.354 & 15.043 & 9.791 & 9.560 \\
\midrule
\multicolumn{6}{l}{\textit{Lumina-mGPT-7B, $768^2$}} \\
GSD & warm & 141.110 & 145.783 & 28.122 & 30.037 \\
MC-SJD & warm & 141.110 & 145.783 & 29.786 & 31.337 \\
SCD & warm & 141.110 & 145.783 & 26.684 & 28.028 \\
SJD & warm & 141.110 & 145.783 & 39.233 & 41.862 \\
PathSpec & request & 121.301 & 83.506 & 36.374 & 30.052 \\
SJD-PAC & request & 149.548 & 150.563 & 46.689 & 52.397 \\
\midrule
\multicolumn{6}{l}{\textit{Anole-7B}} \\
ADT-Tree + LANTERN & request & 30.846 & 36.962 & 28.327 & 29.692 \\
SJD & runner & 34.842 & 37.286 & 34.902 & 37.377 \\
\midrule
\multicolumn{6}{l}{\textit{Emu3}} \\
SJD & runner & 421.113 & 431.097 & 410.897 & 425.363 \\
\midrule
\multicolumn{6}{l}{\cellcolor{T2IGray}\textbf{Diffusion/flow}} \\
\multicolumn{6}{l}{\textit{FLUX.1-schnell}} \\
TaylorSeer & request & 0.826 & 0.826 & 0.559 & 0.559 \\
\midrule
\multicolumn{6}{l}{\textit{FLUX.1-dev}} \\
ClusCa & request & 12.607 & 12.574 & 3.528 & 3.545 \\
SpeCa & request & 12.690 & 12.717 & 2.710 & 2.663 \\
ASD & request & 12.605 & 12.605 & 86.320 & 86.472 \\
Block verification & request & 12.607 & 12.611 & 49.989 & 50.061 \\
Diffusion SD & request & 6.439 & 5.972 & 43.662 & 40.519 \\
\bottomrule
\end{tabularx}
\end{table*}

\begin{table*}[!tp]
\centering
\caption{Auxiliary analysis. Mean recorded AR component times in seconds per output, over the 1,400 benchmark prompts. Prefill overlaps generation; the speculative component timers use generation boundaries specific to each adapter. Controller is residual time. Zero proposal time denotes no separately timed drafter; blank baseline timers are unavailable.}
\label{tab:latency-components}
\scriptsize
\setlength{\tabcolsep}{4pt}
\renewcommand{\arraystretch}{1.10}
\resizebox{\ifdim\width>\linewidth\linewidth\else\width\fi}{!}{%
\begin{tabular}{@{}lrrrrrrrr@{}}
\toprule
Method & Warm time & Prefill & Token gen. & Decode & Postproc. & Proposal & Verify & Controller \\
\midrule
\multicolumn{9}{l}{\textit{Janus-Pro-7B, $384^2$}} \\
Target-only & 14.1384 & 0.0368 & 14.1081 & 0.0271 & 0.0004 &  &  &  \\
CSD & 8.2677 & 0.0642 & 8.2297 & 0.0359 & 0.0001 & 0.0000 & 7.8764 & 0.3534 \\
\multicolumn{9}{l}{\textit{LlamaGen Stage I, $256^2$}} \\
Target-only & 6.3657 & 0.0437 & 6.3276 & 0.0203 & 0.0021 &  &  &  \\
LANTERN & 5.7446 & 0.0337 & 5.7223 & 0.0086 & 0.0011 & 0.8871 & 3.6131 & 1.2221 \\
LANTERN++ & 1.4119 & 0.0363 & 1.3706 & 0.0230 & 0.0048 & 0.2014 & 0.8928 & 0.2764 \\
\multicolumn{9}{l}{\textit{LlamaGen Stage II, $512^2$}} \\
Target-only & 17.4555 & 0.0276 & 17.4184 & 0.0237 & 0.0035 &  &  &  \\
LANTERN & 19.2826 & 0.0307 & 19.2369 & 0.0292 & 0.0043 & 3.0994 & 12.0673 & 4.0702 \\
LANTERN++ & 5.4228 & 0.0346 & 5.3760 & 0.0269 & 0.0054 & 0.7985 & 3.5171 & 1.0605 \\
\multicolumn{9}{l}{\textit{Lumina-mGPT-7B, $768^2$}} \\
Target-only & 141.1097 & 0.0356 & 140.6633 & 0.1917 & 0.0000 &  &  &  \\
GSD & 28.1218 & 0.0448 & 27.7029 & 0.1809 & 0.0000 & 0.0000 & 23.6700 & 4.2710 \\
MC-SJD & 29.7859 & 0.0434 & 29.3874 & 0.1770 & 0.0000 & 0.0000 & 23.2160 & 6.3930 \\
SCD & 26.6838 & 0.0413 & 26.3038 & 0.1690 & 0.0000 & 0.0000 & 21.9246 & 4.5902 \\
SJD & 39.2334 & 0.0426 & 38.8227 & 0.1928 & 0.0000 & 0.0000 & 35.0409 & 3.9997 \\
\bottomrule
\end{tabular}%
}
\end{table*}

\begin{table*}[!tp]
\centering
\caption{Auxiliary analysis. Target work and inter-token latency (ITL) over the 1,400 benchmark prompts. Panel A reports the mean number of target forward passes and the mean 95th percentile ITL of each output. Forward reduction uses each target's mean count. ITL averages the percentiles computed within each output; accepted token groups can produce zero intervals within a group. Panel B reports mean evaluation counts and the number of pairs with lower method latency.}
\label{tab:compact-target-work}
\small
\setlength{\tabcolsep}{4pt}
\renewcommand{\arraystretch}{1.10}
\textit{A. Core target work and ITL}\\[2pt]
\resizebox{\ifdim\width>\linewidth\linewidth\else\width\fi}{!}{%
\begin{tabular}{@{}lrrr@{}}
\toprule
Method & Target forwards & Forward reduction & Mean ITL P95 (ms) \\
\midrule
\multicolumn{4}{l}{\textit{Janus-Pro-7B, $384^2$}} \\
Target-only & 576.00 & 0.0\% & 26.48 \\
CSD & 192.07 & 66.7\% & 47.05 \\
\multicolumn{4}{l}{\textit{LlamaGen Stage I, $256^2$}} \\
Target-only & 256.00 & 0.0\% & 24.83 \\
LANTERN & 173.74 & 32.1\% & 33.60 \\
LANTERN++ & 48.36 & 81.1\% & 28.71 \\
\multicolumn{4}{l}{\textit{LlamaGen Stage II, $512^2$}} \\
Target-only & 1024.00 & 0.0\% & 17.16 \\
LANTERN & 653.23 & 36.2\% & 30.69 \\
LANTERN++ & 179.76 & 82.4\% & 29.89 \\
\multicolumn{4}{l}{\textit{Lumina-mGPT-7B, $768^2$}} \\
Target-only & 4710.00 & 0.0\% & 60.18 \\
GSD & 742.26 & 84.2\% & 41.05 \\
MC-SJD & 692.65 & 85.3\% & 45.80 \\
SCD & 690.11 & 85.3\% & 41.75 \\
SJD & 1076.59 & 77.1\% & 39.39 \\
\bottomrule
\end{tabular}%
}
\par\medskip
\textit{B. Paired target work}\\[2pt]
\resizebox{\ifdim\width>\linewidth\linewidth\else\width\fi}{!}{%
\begin{tabular}{@{}lrrr@{}}
\toprule
Method & Target evals. & Method evals. & Faster pairs \\
\midrule
ADT-Tree + LANTERN & 1024.00 & 594.13 & 912/1,400 \\
ARsample filtering & 1024.00 & 1024.00 & 0/1,400 \\
PathSpec & 4710.00 & 843.83 & 1,400/1,400 \\
\bottomrule
\end{tabular}%
}
\end{table*}

\section{Plugin Implementations and Reference Contracts}
\label{sec:method}

The integrations that generate complete images in Section~\ref{sec:plugin} wrap existing
generation methods, and the bounded diagnostics use a canonical transaction
framework; this section specifies both.

\subsection{Shared Online Estimator}

At iteration $t$, the remaining work $n_t^{\mathrm{rem}}$, backend capabilities,
structural boundaries, and backend signals restrict the configured horizons:
\begin{equation}
 \begin{aligned}
 h_{\max,t}&=\min(n_t^{\mathrm{rem}},h_{\mathrm{runtime}},h_{\mathrm{boundary}},h_{\mathrm{signal}}),\\
 \mathcal H_t&=\{h\in\mathcal H:1\le h\le h_{\max,t}\}.
 \end{aligned}
 \label{eq:horizons}
\end{equation}
Only the available caps enter the minimum. The router keeps
exponentially weighted location and scale estimates with Huber clipping,
update rate $0.2$, clipping multiplier $3$, and a configured envelope multiplier
$z$. LANTERN uses $z=0.25$ and Lumina uses $z=0$. For a nonnegative quantity $X$,
\begin{equation}
 \begin{aligned}
 X_t^L&=\max(0,\mu_{X,t}-z\sigma_{X,t}),\\
 X_t^U&=\mu_{X,t}+z\sigma_{X,t}.
 \end{aligned}
 \label{eq:bands}
\end{equation}
These envelopes are routing heuristics.
Unobserved horizons use priors on component costs and the acceptance fraction
estimate $\alpha_t^L$:
\begin{equation}
 P_t^L(h)=\max(1,h\alpha_t^L).
 \label{eq:progress}
\end{equation}
The initial cost model includes proposal, verification, signal collection,
controller, checkpoint, and expected recovery costs. After observing a
horizon for a model family (autoregressive or diffusion/flow), the router uses its measured progress and total
attempt time:
\begin{equation}
 G_t^L(h)=P_{\mathrm{emp},t}^L(h)C_{T,t}^L
          -C_{\mathrm{attempt},t}^U(h).
 \label{eq:gain}
\end{equation}
The selected LANTERN and Lumina policies use gain margin $m=0$, whereas
Anole uses $m=25\,\mu$s. Acceleration is measured by the actual request
latency. A policy without the gain screen overrides
only the rejection for nonpositive gain. It selects the longest horizon that
passes the same risk and budget tests and still rejects horizons during a cold
start, a cooldown, or with an exhausted budget.

\subsection{Policies for the Six Evaluated Integrations}

\paragraph{Prompt policy.}
A multinomial Naive Bayes model predicts one of the seven topics from
lowercase word unigram and bigram counts. Unnormalized class log scores sum
the log prior and the feature log probabilities weighted by frequency;
unseen features use the stored probability for unknown features. Two frozen
versions are fitted on construction candidates that share no prompt or scene
family with the benchmark or a development set, with smoothing $\alpha=0.5$
and features seen in at least two training prompts. Janus CSD and LANTERN use
the version that also excludes 196 more development prompts; it has 1,526
training prompts and labels all 1,400 benchmark prompts correctly. SpeCa,
Lumina, and Anole use the earlier version. Janus and Anole
also compute an integer bucket $b\in\{0,\ldots,255\}$ from the first byte of
a SHA-256 digest computed from the prompt. The classifier is fitted, but the generators and
draft models are not retrained. The final LANTERN policy applies one
configuration to every prompt.

\paragraph{Janus CSD.}
The Janus-Pro-7B integration generates $384^2$ images with CFG 5 and
temperature 1. Its native speculative window is 16. Here $b$ comes from the
hash of the policy's fixed salt, a newline, and the prompt. For $b<160$, the frozen
policy uses window 20 on predicted color, perspective, and commonsense
prompts, and window 32 on counting prompts; otherwise it retains window 16.
This policy changes the native CSD window without instantiating the online
cost controller. Prompt classification runs inside the request boundary
for all four arms.

\paragraph{LANTERN.}
The LlamaGen Stage~I integration uses $256^2$, CFG 7.5, top-$k$ 1,000,
top-$p$ 1, temperature 1, BF16, and native relaxation $\delta=0.1$.
The online candidate set is $\{2\}$, with one unit of warmup, zero gain
margin, acceptance prior 0.85, and envelope multiplier 0.25.
Native LANTERN supplies the fallback, and the last 64 visual tokens always
use that fallback. The configured risk limit is 1.5 and no diffusion
proxy budget applies.

\paragraph{Lumina SJD.}
The Lumina-mGPT integration uses $768^2$, CFG 3, BF16, image top-$k$ 2,000,
text top-$k$ 10, and native horizon 16. Its candidate horizons are
$\{16,32,64\}$, with one unit of warmup, acceptance prior 0.98, zero gain
margin, and zero envelope multiplier. Predicted color prompts permit a
decision every 192 visual units, commonsense prompts every 256, and
text, texture, and perspective prompts every 64. Counting and motion
prompts stay on native SJD. After at least one attempted intervention,
negative cumulative realized gain disables further interventions.
Native intervals feed measured cost per advanced unit back into the
estimator without counting their progress twice. These AR controls do not
debit the diffusion proxy budget.

\paragraph{Anole SJD.}
The Anole-7B integration uses $512^2$, CFG 7, image top-$k$ 2,000, and
native proposal length 8. Here $b$ uses no salt.
The prompt policy permits length 16 for counting
or commonsense when $b<96$, length 10 for text when $b<64$, and length 10
for color when $b<24$; all remaining prompts retain length 8. Selected
prompts also activate the online controller, whose horizon candidates
$\{2,4,8\}$ differ from these native proposal lengths. The controller uses
a $25\,\mu$s margin, risk limit 1, one unit of warmup, and decisions every
64 visual units.
Structural handling distinguishes visual progress from begin/end markers
and respects raster boundaries.

\paragraph{TaylorSeer.}
FLUX.1-schnell generates $512^2$ images in four steps. The plugin reduces
the initial phase of full computation from three steps to two and, at the
final step, refreshes the last 24 of 38 single-stream blocks and 12 of
19 double-stream blocks. It uses cache age as a proxy with budget 0.15.
This configuration changes the effective schedule and the generated images.

\paragraph{SpeCa.}
FLUX.1-dev generates $512^2$ images in $T=50$ steps with guidance 3.5, base threshold
7, decay 0.3, and native reuse limits of 2 and 10. Predicted text,
perspective, counting, and commonsense prompts use the guarded policy;
color, motion, and texture retain native SpeCa. The guarded policy preserves
the native floor and allows extension of the cache cap, with candidate horizons
$\{1,2,4,8,16\}$, zero gain margin, risk limit 1, and proxy budget 0.25.
It bypasses rejection for nonpositive gain and multiplies the native
threshold by 1.3 only at steps 25--49. The residual proxy $r_t$ tests
\begin{equation}
 \begin{aligned}
 R_t(h)&=\min(1,r_t\sqrt h)\le\tau_{\mathrm q},\\
 q_t(h)&=\min(1,r_th/T)\le B_t.
 \end{aligned}
 \label{eq:cache-proxy}
\end{equation}
The clipped risk test is nonbinding at $\tau_{\mathrm q}=1$; the remaining budget
$B_t$ and native guards apply. In the evaluated prompts, 800 take the guarded route and 600
take the native route.

\subsection{Canonical Transaction Reference}

Figure~\ref{fig:transaction} describes the separate reference contract used
in the bounded diagnostics of Appendix~\ref{sec:diagnostics}. Each attempt
starts from the authoritative state and a recoverable checkpoint. A verifier
certifies the committed prefix or transition, including any correction, and
rejected suffixes remain provisional. Recovery discards the provisional
state, restores the checkpoint, and advances one authoritative target unit. A
failed restoration or an invalid state contract terminates execution.

\begin{figure*}[!tp]
\centering
\resizebox{\linewidth}{!}{%
\begin{tikzpicture}[
  node distance=5mm and 8mm,
  every node/.style={font=\small},
  box/.style={draw=T2IDark, rounded corners=2pt, thick, align=center,
    minimum height=8mm, inner sep=4pt},
  arrow/.style={-{Latex[length=2mm]}, thick, draw=T2IDark},
  dashedarrow/.style={-{Latex[length=2mm]}, thick, dashed, draw=T2IRed}
]
\node[box, fill=T2IBlue!12] (auth) {authoritative state $A_t$};
\node[box, fill=T2IGray, right=of auth] (ckpt) {checkpoint $C_t$};
\node[box, fill=T2IGold!15, right=of ckpt] (prov) {provisional proposal\\and verifier state $P_t$};
\node[diamond, aspect=2.1, draw=T2IDark, thick, fill=white, right=of prov,
  align=center, inner sep=1pt] (cert) {certified?};
\node[box, fill=T2ITeal!15, above right=4mm and 9mm of cert] (commit) {validate receipt\\commit $A_{t+\Delta}$};
\node[box, fill=T2IRed!10, below right=4mm and 9mm of cert] (restore) {discard $P_t$\\restore $C_t$};
\node[box, fill=T2IBlue!12, right=of restore] (fallback) {one authoritative\\target unit $A_{t+1}$};
\draw[arrow] (auth) -- (ckpt);
\draw[arrow] (ckpt) -- (prov);
\draw[arrow] (prov) -- (cert);
\draw[arrow] (cert) -- node[above left, font=\scriptsize] {yes} (commit);
\draw[dashedarrow] (cert) -- node[below left, font=\scriptsize, align=center]
  {reject /\\recoverable error} (restore);
\draw[arrow] (restore) -- (fallback);
\end{tikzpicture}%
}
\caption{Authoritative transaction semantics. Proposals remain provisional
until a validated state transition is committed. Rejection or a recoverable
error restores the checkpoint before target fallback. Invalid state contracts
or failed restoration terminate the run.}
\label{fig:transaction}
\end{figure*}
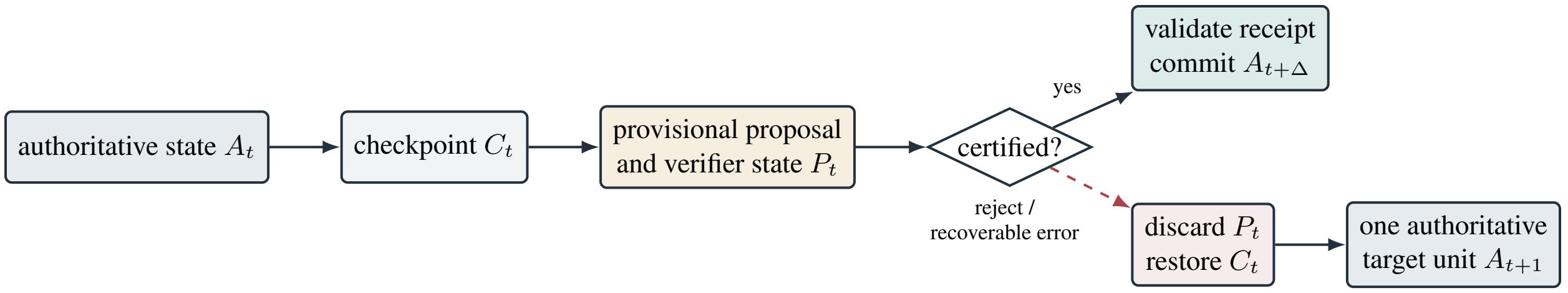

The canonical AR reference admits horizons satisfying
\begin{equation}
 \begin{aligned}
 h&\in\mathcal H_t,& E_{\mathrm{exact}}(h)&=1,\\
 R_{\mathrm{op},t}^U(h)&\le\tau_{\mathrm{op}},&G_t^L(h)&>m,
 \end{aligned}
 \label{eq:ar-feasible}
\end{equation}
where $E_{\mathrm{exact}}$ is the indicator of complete capability eligibility
and $R_{\mathrm{op}}$ is operational risk with limit $\tau_{\mathrm{op}}$. An eligible proposal is then verified: the verifier checks raw
tokens and the continuation state, including CFG and key-value (KV) caches, masks, positions,
and random number state. The canonical Lumina implementation verifies one row
at a time and uses a sequential authoritative commit to avoid numerical
differences caused by tensor shape.

The reference diffusion transaction instead requires
\begin{equation}
 \begin{aligned}
 h&\in\mathcal H_t,& R_{\mathrm{op},t}^U(h)&\le\tau_{\mathrm{op}},\\
 R_{\mathrm q,t}^U(h)&\le\tau_{\mathrm q},&
 Q_t^U(h)&\le B_t^{\mathrm{rem}},\\
 B_t^{\mathrm{rem}}&>0,&G_t^L(h)&>m,
 \end{aligned}
 \label{eq:diff-feasible}
\end{equation}
where $R_{\mathrm q}$ is quality risk with limit $\tau_{\mathrm q}$ and $Q_t^U$ is the proxy
charge against the remaining budget $B_t^{\mathrm{rem}}$. Successful transitions debit the proxy ledger.
The earlier FLUX trajectory pilot used two exact warmup transitions,
followed by a calibrated fixed horizon with a possible fallback for the entire output.
Appendix~\ref{app:budget-diagnostics} reports that diagnostic, whereas the
SpeCa and TaylorSeer integrations evaluated on 1,400 prompts act at each step.

\subsection{Evaluation on the Complete Benchmark}
\label{app:plugin-evaluation}

\paragraph{Population and matched arms.}
The final study contains six model--method pairs: Janus CSD, LlamaGen
LANTERN, Lumina SJD, Anole SJD, TaylorSeer, and SpeCa. Each covers all 1,400
canonical prompts, with 200 per topic, 700 per difficulty, and 100 per
topic/difficulty cell. Target-only, native, plugin, and observation-only
arms give 33,600 logical records, one per arm and prompt. All plugin comparisons use
NVIDIA H200 GPUs. Generation controls and prompt seeds are matched within
a group. Each group's own references remain separate from the benchmark
tables in Section~\ref{sec:overall-comparison}.

\paragraph{Measures and pass criterion.}
For method arm $a$ and topic $g$, define
\begin{equation}
 \begin{aligned}
 S_g^a&=\bar t_g^{\rm target}/\bar t_g^a,&
 D^a&=\max_g S_g^a/\min_g S_g^a-1,\\
 U_g&=\bar t_g^{\rm native}/\bar t_g^{\rm plugin},&
 W&=\sum_g\mathbf{1}[U_g>1].
 \end{aligned}
 \label{eq:plugin-speed-objective}
\end{equation}
Overall ratios pool all 1,400 prompts. For eight quality metrics with higher values preferred,
\begin{equation}
 \begin{aligned}
 \Delta Q&=\tfrac17\sum_g(Q_g^{\rm plugin}-Q_g^{\rm native}),\\
 K&=\sum_{j=1}^{8}\mathbf{1}[\bar M_j^{\rm plugin}\ge\bar M_j^{\rm native}].
 \end{aligned}
 \label{eq:plugin-topic-objective}
\end{equation}
A group passes when $U>1$, $S^{\rm plugin}>1$, $W\ge4$,
$D^{\rm plugin}<D^{\rm native}$, and $K\ge5$. The campaign criterion is at
least four passing groups and a geometric mean $U>1$ across all six.
These checks use observed means.
Topic counts and pass decisions use unrounded values. Topics left on the
native path also count toward $W$; their ratios stay within 0.2\% of one.
Four groups pass and the geometric mean of $U$ across all six groups is
$1.0308\times$, which meets the campaign criterion.
Five groups reduce $D$, but TaylorSeer fails the quality check, and Lumina
fails the Overall speed, disparity, and quality checks.
Table~\ref{tab:plugin-details} retains every group's topic results,
difficulty results, eight quality differences, and the timing of all four arms.

\paragraph{Timing and scoring.}
Recorded request latency includes generation and the routing work applied
by the respective runner; loading and warmup are outside the request.
SpeCa warms all four arms, unlike the earlier study below.
Janus and Anole include prompt classification in every arm; Lumina adds
its measured classification cost to the request. LANTERN synchronizes
around generation and image saving, so its boundary includes saving while
the other runners stop at image generation or decode. These boundaries
are kept fixed within each group.

Where an arm has repeated request measurements, we average within each
prompt before averaging over prompts. LANTERN quality uses the first
measured image, while its native/plugin latency averages three
measurements. For 613 plugin prompts, a later measurement produces a
different image; the quality scores therefore refer to the first
output of each prompt. All eight metrics cover every arm and prompt,
and both topic evaluators use the protocol in Appendix~\ref{app:evaluation-protocol}
with batch size one.

\begin{figure}[t]
\centering
\includegraphics[width=\columnwidth]{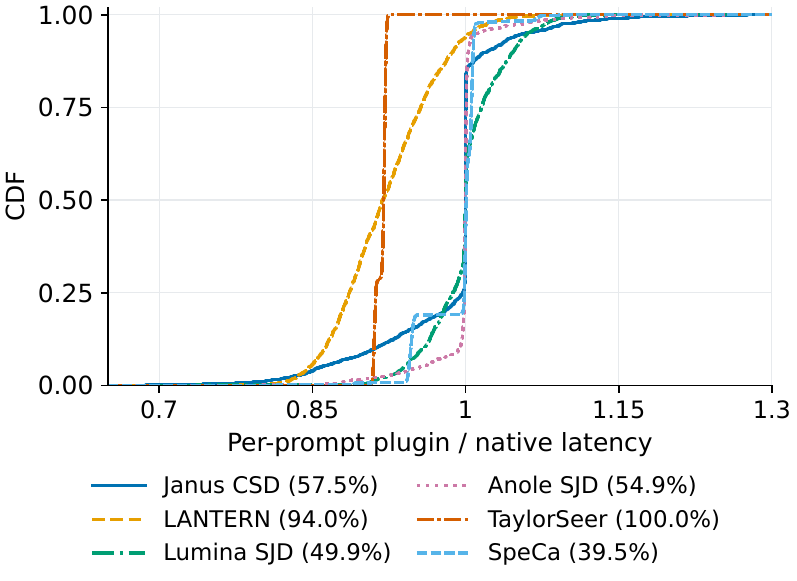}
\caption{CDFs of the plugin/native latency ratio over all 1,400 prompts per method. Ratios below one indicate faster requests; legend percentages give the fraction of faster prompts.}
\label{fig:plugin-latency-cdf}
\end{figure}

\paragraph{Latency distributions across prompts.}
Figure~\ref{fig:plugin-latency-cdf} plots the empirical CDF of
$t_i^{\rm plugin}/t_i^{\rm native}$ for all 1,400 prompts in each of the six
comparisons. Repeated measurements are averaged within each prompt as
above; ratios below one indicate faster execution with \topicRoute.
LANTERN lowers latency for 94.0\% of prompts, compared with 39.5\% for
SpeCa despite the latter's Overall gain. The curves distinguish consistent
improvements across prompts from gains in the mean; the legend reports the
fraction of prompts with a ratio strictly below one.

\paragraph{Observation-only control.}
All six observation-only arms reproduce their native arm's first image on
1,400 of 1,400 prompts. Their mean times differ from native by
$-0.0283\%$ to $+0.5494\%$. This comparison measures each runner's
observation path: some native arms already execute classification, and the observer's work
differs across backends.

\paragraph{Scope of the frozen policies.}
The final evaluations follow development and earlier benchmark feedback.
For example, Janus's hash policy was selected using an earlier run on the
same prompt set and then regenerated with a different seed.
Appendix~\ref{app:plugin-followup} adds matched ablations of these final
policies and an external prompt set;
Appendix~\ref{app:plugin-legacy-evaluation} documents earlier controller
development.

\begin{table*}[t]
\centering
\caption{Detailed results for the six \topicRoute groups: topic speed $U$, eight quality changes, difficulty strata, topic $\Delta Q$, and request latency. Quality changes are plugin minus native. Each topic contains 200 prompts; Easy and Hard contain 700 each. Observer overhead compares complete requests.}
\label{tab:plugin-details}
\begingroup
\setlength{\tabcolsep}{3pt}
\renewcommand{\arraystretch}{1.06}
\scriptsize
\begin{tabular}{lrrrrrrrr}
\toprule
\multicolumn{9}{l}{\textbf{(a) Speed relative to the native method by topic and Overall}} \\
Method & Color & Text & Persp. & Count & Motion & Texture & Common. & Overall \\
\midrule
Janus / CSD & 1.0123 & 1.0020 & 1.0250 & 1.0258 & 0.9999 & 1.0000 & 1.0262 & 1.0134 \\
LANTERN & 1.0542 & 1.0984 & 1.0849 & 1.0860 & 1.0797 & 1.0889 & 1.0491 & 1.0739 \\
Lumina / SJD & 0.9959 & 0.9939 & 0.9992 & 1.0001 & 1.0002 & 1.0069 & 1.0010 & 0.9995 \\
Anole / SJD & 1.0003 & 1.0031 & 1.0001 & 1.0042 & 1.0001 & 1.0001 & 1.0117 & 1.0029 \\
TaylorSeer & 1.0862 & 1.0936 & 1.0943 & 1.0866 & 1.0945 & 1.0940 & 1.0861 & 1.0916 \\
SpeCa & 0.9998 & 1.0128 & 1.0151 & 1.0166 & 0.9996 & 0.9989 & 1.0090 & 1.0074 \\
\midrule
\multicolumn{9}{l}{\textbf{(b) Overall quality changes relative to the native method}} \\
Method & $\Delta$CLIP & $\Delta$IR & $\Delta$HPS & $\Delta Q$ & $\Delta$Aes. & $\Delta$SigLIP & $\Delta Q_2$ & $\Delta$Safe \\
\midrule
Janus / CSD & +0.0005 & -0.0044 & +0.0001 & +0.0025 & +0.0001 & -0.0023 & +0.0010 & +0.0007 \\
LANTERN & -0.0004 & +0.0082 & -0.0001 & -0.0063 & +0.0050 & +0.0017 & +0.0218 & +0.0271 \\
Lumina / SJD & -0.0015 & -0.0235 & -0.0011 & +0.0104 & -0.0117 & -0.0039 & +0.0071 & +0.0021 \\
Anole / SJD & -0.0019 & +0.0139 & -0.0003 & +0.0015 & +0.0006 & +0.0027 & +0.0101 & +0.0050 \\
TaylorSeer & +0.0202 & -0.0786 & -0.0183 & -0.0154 & -0.1768 & +0.0145 & -0.0019 & -0.0150 \\
SpeCa & +0.0005 & -0.0002 & -0.0002 & +0.0063 & +0.0006 & -0.0005 & +0.0030 & +0.0014 \\
\midrule
\multicolumn{9}{l}{\textbf{(c) Easy/Hard speed and criterion quality}} \\
 & \multicolumn{4}{c}{Easy} & \multicolumn{4}{c}{Hard} \\
Method & $U$ & $Q_{\rm native}$ & $Q_{\rm plugin}$ & $\Delta Q$ & $U$ & $Q_{\rm native}$ & $Q_{\rm plugin}$ & $\Delta Q$ \\
\midrule
Janus / CSD & 1.0146 & 0.8307 & 0.8364 & +0.0057 & 1.0123 & 0.8486 & 0.8479 & -0.0007 \\
LANTERN & 1.0735 & 0.6643 & 0.6619 & -0.0024 & 1.0744 & 0.6336 & 0.6233 & -0.0102 \\
Lumina / SJD & 0.9992 & 0.7471 & 0.7481 & +0.0010 & 0.9999 & 0.7631 & 0.7829 & +0.0198 \\
Anole / SJD & 1.0021 & 0.7195 & 0.7205 & +0.0010 & 1.0037 & 0.7507 & 0.7529 & +0.0021 \\
TaylorSeer & 1.0915 & 0.8738 & 0.8605 & -0.0133 & 1.0916 & 0.9176 & 0.9002 & -0.0174 \\
SpeCa & 1.0078 & 0.8767 & 0.8845 & +0.0079 & 1.0069 & 0.8881 & 0.8929 & +0.0048 \\
\midrule
\multicolumn{9}{l}{\textbf{(d) Changes in the primary criterion by topic and Overall}} \\
Method & Color & Text & Persp. & Count & Motion & Texture & Common. & Overall \\
\midrule
Janus / CSD & +0.0017 & +0.0000 & +0.0042 & -0.0050 & +0.0000 & +0.0000 & +0.0167 & +0.0025 \\
LANTERN & +0.0183 & -0.0342 & -0.0175 & -0.0492 & +0.0583 & -0.0042 & -0.0158 & -0.0063 \\
Lumina / SJD & -0.0008 & +0.0658 & +0.0108 & +0.0000 & +0.0000 & +0.0067 & -0.0100 & +0.0104 \\
Anole / SJD & -0.0058 & +0.0275 & +0.0000 & -0.0042 & +0.0000 & +0.0000 & -0.0067 & +0.0015 \\
TaylorSeer & +0.0017 & -0.0242 & +0.0000 & -0.0850 & +0.0700 & +0.0008 & -0.0708 & -0.0154 \\
SpeCa & +0.0000 & +0.0192 & +0.0025 & +0.0075 & +0.0000 & +0.0000 & +0.0150 & +0.0063 \\
\midrule
\multicolumn{9}{l}{\textbf{(e) Absolute request latency (seconds) and observer control}} \\
Method & Target & Native & Plugin & Observer & \multicolumn{2}{c}{Observer overhead (\%)} & \multicolumn{2}{c}{Identical images} \\
\midrule
Janus / CSD & 7.1458 & 3.0042 & 2.9644 & 3.0034 & \multicolumn{2}{r}{-0.0283} & \multicolumn{2}{r}{1,400/1,400} \\
LANTERN & 6.0467 & 6.4025 & 5.9617 & 6.4110 & \multicolumn{2}{r}{+0.1337} & \multicolumn{2}{r}{1,400/1,400} \\
Lumina / SJD & 56.6571 & 31.5445 & 31.5588 & 31.5517 & \multicolumn{2}{r}{+0.0226} & \multicolumn{2}{r}{1,400/1,400} \\
Anole / SJD & 22.9410 & 16.6474 & 16.5995 & 16.7067 & \multicolumn{2}{r}{+0.3562} & \multicolumn{2}{r}{1,400/1,400} \\
TaylorSeer & 0.5681 & 0.3477 & 0.3185 & 0.3485 & \multicolumn{2}{r}{+0.2385} & \multicolumn{2}{r}{1,400/1,400} \\
SpeCa & 6.0019 & 1.2789 & 1.2695 & 1.2859 & \multicolumn{2}{r}{+0.5494} & \multicolumn{2}{r}{1,400/1,400} \\
\midrule
\multicolumn{9}{l}{Identical images compares observer and native image hashes for the same prompt.} \\
\bottomrule
\end{tabular}
\endgroup
\end{table*}

\subsection{Uncertainty, Ablations, and External Prompts}
\label{app:plugin-followup}

\paragraph{Paired intervals.}
Table~\ref{tab:plugin-intervals} applies the paired resampling of
Appendix~\ref{app:resampling-details} to the six groups, using the same
indices. The interval for $U$ excludes one for every group except Lumina,
whose interval is $[0.9976, 1.0015]$. For $\Delta Q$, only TaylorSeer's
interval, $[-0.0262, -0.0040]$, excludes zero. Of the 48 quality
differences (six groups and eight metrics), 12 have intervals that exclude
zero, seven of them for TaylorSeer. Of the 27 nondecreasing means counted
by $K$, only four have intervals above zero, so $K$ mostly counts
differences that cannot be distinguished from zero.

\begin{table}[t]
\centering
\caption{Paired 95\% percentile intervals for the six \topicRoute groups. The 2,000 replicates reuse the resampling indices of the core intervals: 100 connected Easy/Hard groups per topic, with all arms of a selected prompt sharing its multiplicity. $U$ is a ratio of resampled mean latencies and $\Delta Q$ a mean paired difference.}
\label{tab:plugin-intervals}
\begingroup
\setlength{\tabcolsep}{3pt}
\renewcommand{\arraystretch}{1.06}
\scriptsize
\begin{tabular}{lrcrc}
\toprule
Method & $U$ & 95\% interval & $\Delta Q$ & 95\% interval \\
\midrule
Janus / CSD & 1.0134 & [1.0098, 1.0171] & +0.0025 & [-0.0060, +0.0107] \\
LANTERN & 1.0739 & [1.0707, 1.0772] & -0.0063 & [-0.0286, +0.0148] \\
Lumina / SJD & 0.9995 & [0.9976, 1.0015] & +0.0104 & [-0.0001, +0.0207] \\
Anole / SJD & 1.0029 & [1.0015, 1.0043] & +0.0015 & [-0.0073, +0.0105] \\
TaylorSeer & 1.0916 & [1.0912, 1.0920] & -0.0154 & [-0.0262, -0.0040] \\
SpeCa & 1.0074 & [1.0060, 1.0088] & +0.0063 & [-0.0019, +0.0140] \\
\bottomrule
\end{tabular}
\endgroup
\end{table}

\paragraph{Reruns and rescoring.}
The ablation studies rerun the final policies of every group except
TaylorSeer with their frozen settings (Table~\ref{tab:plugin-ablations}).
The reruns reproduce the images of Table~\ref{tab:plugin-main} bitwise
for 248--1,000 of 1,400 prompts. Every rerun of a policy with $U>1$ keeps
$U>1$, with values from 1.0025 to 1.0825, and Lumina again shows no gain
($U=0.9990$). The quality count changes much more. $K$ falls from 6 to
between 1 and 3 in three Janus reruns, from 5 to 2 for LANTERN, and from 5
to between 2 and 4 in four SpeCa reruns. It rises from 6 to 8 for Anole
and from 3 to 4 for Lumina. Of the four groups that pass in
Table~\ref{tab:plugin-main}, only Anole passes again. The studies of
LANTERN's terminal guard and TaylorSeer's single-stream refresh instead
reuse the images of Table~\ref{tab:plugin-main} and rescore them with the
same frozen evaluator binding and batch size. The primary criterion score
changes for 127--167 of 1,400 images per arm. $\Delta Q$ moves from
$-0.0063$ to $-0.0018$ for LANTERN and from $-0.0154$ to $-0.0093$ for
TaylorSeer, while $K$ is unchanged. The speed gains therefore recur across
runs, whereas $K$ changes between reruns and $\Delta Q$ changes on
rescoring.

\paragraph{Ablations of the final policies.}
Table~\ref{tab:plugin-ablations} varies components of the final policies
of all six groups.
Anole's variants share all control images. Removing its hash gates raises
$U$ from 1.0025 to 1.0095 and lowers $D_{\rm plugin}$ from 0.0627 to 0.0517.
A second variant assigns proposal length 16 to all 221 accelerated prompts. It
changes only 69 Color and Text prompts and keeps topic and hash
eligibility, so it removes the proposal length chosen for each topic but
keeps topic routing. This variant raises $U$ to 1.0041 and, like the rerun
and the variant without hash gates, keeps $K=8$. Removing online control
leaves $U$ at 1.0026 and lowers $\Delta Q$ from $+0.0074$ to $+0.0006$ and
$K$ to 5. In Janus's paired runs, removing the
hash gate admits 299 more prompts, raises $U$ from 1.0163 to 1.0214, and
lowers $D_{\rm plugin}$ from 0.2488 to 0.2303. Shuffling 716 of the 1,400
route assignments raises $U$ from 1.0159 to 1.0198, but $D_{\rm plugin}$
rises from 0.2670 to 0.3399, above its native value of 0.3021. Janus's
topic routing therefore improves balance, not speed.

LANTERN's gain screen, ablated with shared control images, trades speed for
balance. Removing it raises $U$ from 1.0825 to
1.1225 and $\Delta Q$ from $-0.0181$ to $-0.0043$, whereas $D_{\rm plugin}$
rises from 0.0274 to 0.0517. In SpeCa, threshold relaxation produces the
gain. Without it the plugin is slower than native ($U=0.9971$ and $0.9979$,
with one faster topic), and a proxy budget of 0.15 instead of 0.25 lowers
$U$ from 1.0132 to 1.0006. Removing the budget, enabling the gain screen,
or shuffling 670 of the 1,400 route assignments changes $U$ by at most
0.11\% relative to the final policy of the same run. All four Lumina arms
stay within 0.1\% of native speed ($U$ from 0.9990 to 1.0003). Removing
its gain screen or its stop after a negative cumulative gain, or shuffling
its route, changes $U$ by at most 0.12\% relative to the final policy and
leaves $K=4$.

Three comparisons come from separate runs whose control images match the
final run on only 708--800 prompts. Without LANTERN's terminal guard, which
returns the last 64 visual tokens to native LANTERN, $U$ rises to 1.1279. Without
TaylorSeer's refresh of 24 single-stream blocks at the final step, with
the 12 double-stream refreshes kept, $U$ rises to 1.2225 and $\Delta Q$
falls from $-0.0093$ to $-0.0524$. This loss far exceeds the change in
$\Delta Q$ between Table~\ref{tab:plugin-main} and the reruns of the final
policies (at most 0.0118), whereas
LANTERN's change from $-0.0018$ to $-0.0098$ lies within that range. The
Janus run with a uniform window removes topic routing, and its $D_{\rm plugin}$
of 0.1929 exceeds its $D_{\rm native}$ of 0.1742, consistent with the paired
shuffle.

\begin{table*}[t]
\centering
\caption{Ablations and reruns of the final \topicRoute policies on the 1,400 benchmark prompts. Each row compares a variant with the native arm of its own run, and $S$ and $D$ use that run's target-only arm, which the SpeCa runs lack. Rows with the same run number share control images, or one native arm for SpeCa. Ctrl.\ counts prompts whose target and native images equal those of the first row of the group. Differences between rows with different run numbers also reflect changes in their control images. $^\dagger$Images of Table~\ref{tab:plugin-main}, rescored together with the ablation.}
\label{tab:plugin-ablations}
\begingroup
\setlength{\tabcolsep}{3pt}
\renewcommand{\arraystretch}{1.06}
\scriptsize
\begin{tabular}{llcrrrrrrrr}
\toprule
Group & Variant & Run & $U\uparrow$ & $S\uparrow$ & $W$/7 & $D_{\rm native}$ & $D_{\rm plugin}$ & $\Delta Q$ & $K$/8 & Ctrl. \\
\midrule
Janus / CSD & Final policy, rerun & 1 & 1.0172 & 2.4595 & 5 & 0.2548 & 0.2375 & -0.0060 & 1 & -- \\
 & Uniform window, no topic routing & 2 & 1.0299 & 2.5123 & 7 & 0.1742 & 0.1929 & -0.0115 & 6 & 749 \\
 & Final policy, rerun & 3 & 1.0163 & 2.4680 & 7 & 0.2658 & 0.2488 & -0.0026 & 3 & 554 \\
 & No hash gate & 3 & 1.0214 & 2.4883 & 5 & 0.2676 & 0.2303 & -0.0035 & 2 & 554 \\
 & Final policy, rerun & 4 & 1.0159 & 2.4880 & 5 & 0.2981 & 0.2670 & -0.0035 & 2 & 754 \\
 & Shuffled route & 4 & 1.0198 & 2.4944 & 7 & 0.3021 & 0.3399 & +0.0076 & 5 & 754 \\
\midrule
LANTERN & Final policy$^\dagger$ & 1 & 1.0739 & 1.0143 & 7 & 0.0659 & 0.0226 & -0.0018 & 5 & -- \\
 & No terminal guard & 2 & 1.1279 & 1.0508 & 7 & 0.0591 & 0.0157 & -0.0098 & 5 & 708 \\
 & Final policy, rerun & 3 & 1.0825 & 1.0061 & 7 & 0.0811 & 0.0274 & -0.0181 & 2 & 400 \\
 & No gain screen & 3 & 1.1225 & 1.0418 & 7 & 0.0756 & 0.0517 & -0.0043 & 3 & 400 \\
\midrule
Lumina / SJD & Final policy, rerun & 1 & 0.9990 & 1.8246 & 3 & 0.0474 & 0.0430 & +0.0023 & 4 & -- \\
 & Shuffled route & 1 & 0.9990 & 1.8245 & 3 & 0.0474 & 0.0433 & -0.0055 & 4 & 1,400 \\
 & No gain screen & 1 & 0.9999 & 1.8261 & 3 & 0.0474 & 0.0389 & +0.0044 & 4 & 1,400 \\
 & No stop after negative gain & 1 & 1.0003 & 1.8269 & 5 & 0.0474 & 0.0450 & -0.0015 & 4 & 1,400 \\
\midrule
Anole / SJD & Final policy, rerun & 1 & 1.0025 & 1.3816 & 6 & 0.0711 & 0.0627 & +0.0074 & 8 & -- \\
 & No hash gates & 1 & 1.0095 & 1.3905 & 5 & 0.0726 & 0.0517 & +0.0065 & 8 & 1,400 \\
 & No online control & 1 & 1.0026 & 1.3821 & 6 & 0.0723 & 0.0645 & +0.0006 & 5 & 1,400 \\
 & One proposal length for selected prompts & 1 & 1.0041 & 1.3844 & 7 & 0.0721 & 0.0617 & +0.0085 & 8 & 1,400 \\
\midrule
TaylorSeer & Final policy$^\dagger$ & 1 & 1.0916 & 1.7838 & 7 & 0.0197 & 0.0123 & -0.0093 & 2 & -- \\
 & No single-stream refresh & 2 & 1.2225 & 2.0000 & 7 & 0.0189 & 0.0075 & -0.0524 & 2 & 800 \\
\midrule
SpeCa & Final policy, rerun & 1 & 1.0100 & -- & 6 & -- & -- & +0.0063 & 2 & -- \\
 & With gain screen & 1 & 1.0111 & -- & 6 & -- & -- & +0.0054 & 2 & -- \\
 & Final policy, rerun & 2 & 1.0132 & -- & 7 & -- & -- & +0.0015 & 4 & -- \\
 & Proxy budget 0.15 & 2 & 1.0006 & -- & 4 & -- & -- & +0.0012 & 8 & -- \\
 & No threshold relaxation & 2 & 0.9971 & -- & 1 & -- & -- & +0.0002 & 6 & -- \\
 & Budget 0.15, no relaxation & 2 & 0.9979 & -- & 1 & -- & -- & +0.0000 & 8 & -- \\
 & Final policy, rerun & 3 & 1.0090 & -- & 4 & -- & -- & +0.0015 & 2 & -- \\
 & No proxy budget & 3 & 1.0080 & -- & 5 & -- & -- & +0.0008 & 3 & -- \\
 & Final policy, rerun & 4 & 1.0078 & -- & 6 & -- & -- & -0.0011 & 2 & -- \\
 & Shuffled route & 4 & 1.0069 & -- & 7 & -- & -- & +0.0015 & 3 & -- \\
\bottomrule
\end{tabular}
\endgroup
\end{table*}

\paragraph{External DrawBench prompts.}
We also run the frozen policies without retuning on 200 DrawBench
prompts~\cite{saharia2022photorealistic}. An audit of prompt hashes and
normalized text finds no overlap with the benchmark or with eleven
development and holdout prompt sets. Table~\ref{tab:plugin-holdout} shows
that five policies remain faster than native with intervals above one,
from $U=1.0142$ for Anole to 1.0972 for TaylorSeer, and the geometric mean
of $U$ is $1.0375\times$. Lumina again shows no gain ($U=0.9984$, interval
$[0.9936, 1.0030]$). TaylorSeer again loses quality, with five of six
common metrics lower and an aesthetic change of $-0.2216$, while the other
groups keep three to five nondecreasing metrics. Without a topic rubric,
these prompts test speed and generic quality but not $Q$ or topic
disparity.

\begin{table*}[t]
\centering
\caption{Frozen \topicRoute policies on 200 external DrawBench prompts that share no prompt with the benchmark or the development sets. $U$ intervals resample prompts with 2,000 paired replicates. Quality changes are plugin minus native for six common metrics, and $K$ counts nondecreasing means. The prompts have no topic rubric, and SpeCa has no target-only arm.}
\label{tab:plugin-holdout}
\begingroup
\setlength{\tabcolsep}{3pt}
\renewcommand{\arraystretch}{1.06}
\scriptsize
\begin{tabular}{lrcrrrrrrrr}
\toprule
Method & $U\uparrow$ & 95\% interval & $S\uparrow$ & $\Delta$CLIP & $\Delta$IR & $\Delta$HPS & $\Delta$Aes. & $\Delta$SigLIP & $\Delta$Safe & $K$/6 \\
\midrule
Janus / CSD & 1.0566 & [1.0429, 1.0706] & 3.0898 & +0.0037 & -0.0106 & +0.0006 & +0.0195 & +0.0164 & -0.0100 & 4 \\
LANTERN & 1.0474 & [1.0408, 1.0543] & 1.0286 & +0.0102 & +0.1204 & +0.0014 & +0.0434 & +0.0408 & -0.0200 & 5 \\
Lumina / SJD & 0.9984 & [0.9936, 1.0030] & 1.8887 & +0.0032 & -0.0265 & -0.0000 & +0.0341 & +0.0129 & -0.0050 & 3 \\
Anole / SJD & 1.0142 & [1.0077, 1.0222] & 1.4076 & +0.0005 & +0.0302 & -0.0001 & -0.0020 & +0.0060 & -0.0100 & 3 \\
TaylorSeer & 1.0972 & [1.0970, 1.0975] & 1.7690 & +0.0050 & -0.0294 & -0.0167 & -0.2216 & -0.0104 & -0.0400 & 1 \\
SpeCa & 1.0143 & [1.0100, 1.0188] & -- & +0.0017 & +0.0010 & -0.0001 & -0.0012 & +0.0015 & +0.0000 & 4 \\
\midrule
Geometric mean & 1.0375 & [1.0345, 1.0407] &  &  &  &  &  &  &  &  \\
\bottomrule
\end{tabular}
\endgroup
\end{table*}

\subsection{Earlier Controller Studies}
\label{app:plugin-legacy-evaluation}

The studies below evaluate earlier controller versions. Four initial
integrations each cover all 1,400 prompts with target-only, original, and
plugin arms. Four later SpeCa variants and thirteen ablation configurations
also cover the complete benchmark, giving 21 distinct configurations. The
relaxed LANTERN configuration is counted once. Each
comparison uses its own paired reference and quality scores.

\paragraph{Earlier outcomes and timing boundaries.}
The initial LANTERN composition improves recorded request time by a factor
of 1.064 and accelerates all seven topics, but its target/plugin ratio is
only 0.8609. Its criterion gain of 0.1402 accompanies declines in six of
the other seven quality metrics and a change in verification policy.
Lumina SJD takes 27.319 seconds with the plugin versus 27.132 seconds for
the original. This earlier controller includes passive native rollbacks in
its cooldown. TaylorSeer takes 0.5612 versus 0.5590 seconds and
preserves all 1,400 original images byte for byte. Identical images reuse the
same scores, so all eight quality changes are zero for this pair.
The default SpeCa composition slows every topic, with an Overall ratio of
0.344.

Request times include controller and generation work but exclude loading
and explicit compilation. The earlier SpeCa plugin workers warm up their
cache kernel outside this window, whereas the original timings include
first requests without that explicit warmup. TaylorSeer's target time
covers generation alone, while both original and plugin times include
preprocessing. Its original/plugin comparison shares a request boundary,
but its ratios relative to the target retain this boundary difference.

SpeCa's guarded budget, guarded refresh, and guarded prefix variants
accelerate all seven topics and reduce the speed spread. Early
relaxation has an Overall ratio of 1.005 but improves only three topics.
Guarded refresh records 1.0936, with a criterion decrease of 0.0645 Overall
and 0.1750 on perspective. Including worker startup and compilation
amortization changes its speed ratio to 0.7388.

\paragraph{Ablations and a fixed refresh control.}
Table~\ref{tab:plugin-variants} retains all thirteen earlier ablations and
the four later SpeCa variants.
The SpeCa ablations use different native-method scores from the initial
study, so quality changes remain paired within each study.
Relaxed LANTERN improves five of eight aggregate quality metrics
but is slower than original LANTERN; it was added after the quality gate
results were seen and changes several settings together.

A SpeCa control with a fixed refresh interval takes 7.358 seconds per
image, versus 7.869 seconds for the initial plugin composition. The control
takes 6.49\% less time than the plugin, while the plugin's criterion exceeds
the control's by 0.0201.

\paragraph{Development sensitivity.}
Table~\ref{tab:plugin-development} retains 23 settings from five development
studies, each evaluated on 96 prompts with six quality measures, four
shown in the table. The studies vary horizons, gain screening, proxy
risk, budgets, guarded thresholds, prefix protection, and early
relaxation.

\begin{table*}[t]
\centering
\caption{Earlier \topicRoute controller variants and ablations (1,400 paired prompts each). $U$ is the speed ratio relative to the original method, $W$ counts faster topics, and $D$ is the spread in speed relative to the target; IR denotes ImageReward. Quality differences use the paired arm for the original method. The relaxed LANTERN row uses $\delta=0.05$.}
\label{tab:plugin-variants}
\begingroup
\setlength{\tabcolsep}{3pt}
\renewcommand{\arraystretch}{1.06}
\scriptsize
\begin{tabular}{llrrrrrr}
\toprule
Family & Configuration & $U\uparrow$ & $W$/7 & $D_{\rm orig}$ & $D_{\rm plugin}$ & $\Delta Q\uparrow$ & $\Delta$IR$\uparrow$ \\
\midrule
\multicolumn{8}{l}{\textbf{Later complete variants}} \\
SpeCa & guarded budget 0.25 & 1.059 & 7 & 0.050 & 0.036 & -0.0605 & -0.0578 \\
SpeCa & guarded refresh & 1.094 & 7 & 0.050 & 0.036 & -0.0645 & -0.0581 \\
SpeCa & guarded prefix & 1.018 & 7 & 0.050 & 0.047 & -0.0552 & -0.0066 \\
SpeCa & early relaxation & 1.005 & 3 & 0.050 & 0.038 & -0.0593 & -0.0075 \\
\midrule
\multicolumn{8}{l}{\textbf{Component ablations}} \\
LANTERN & Fixed horizon 4 & 1.116 & 7 & 0.070 & 0.036 & -0.0105 & -0.0479 \\
LANTERN & Relaxed drafter / budget / risk & 0.941 & 0 & 0.070 & 0.010 & +0.0454 & +0.0428 \\
LANTERN & Maximum horizon 2 & 1.132 & 7 & 0.070 & 0.008 & -0.0150 & -0.0457 \\
LANTERN & Minimum gain 50 ms & 1.079 & 7 & 0.070 & 0.034 & -0.0201 & -0.0626 \\
LANTERN & Without gain screen & 1.065 & 7 & 0.070 & 0.055 & -0.0182 & -0.0569 \\
LANTERN & Observe only & 1.055 & 7 & 0.070 & 0.057 & -0.0244 & -0.0576 \\
SpeCa & Fixed horizon 4 & 0.339 & 0 & 0.050 & 0.046 & +0.0076 & +0.0131 \\
SpeCa & Maximum horizon 2 & 0.339 & 0 & 0.050 & 0.048 & +0.0101 & +0.0137 \\
SpeCa & Minimum gain 50 ms & 0.328 & 0 & 0.050 & 0.026 & -0.0007 & +0.0047 \\
SpeCa & Without gain screen & 0.536 & 0 & 0.050 & 0.026 & +0.0074 & +0.0137 \\
SpeCa & Observe only & 1.012 & 5 & 0.050 & 0.032 & -0.0076 & -0.0096 \\
SpeCa & Proxy budget 0.10 & 0.345 & 0 & 0.050 & 0.039 & +0.0023 & +0.0131 \\
SpeCa & Risk limit 0.35 & 0.345 & 0 & 0.050 & 0.027 & +0.0058 & +0.0132 \\
\midrule
\multicolumn{8}{l}{\textbf{Cost control with a fixed refresh interval for the original SpeCa integration}} \\
\multicolumn{8}{l}{Control: 7.358 s; plugin: 7.869 s; control/plugin time ratio: 0.935; plugin minus control $\Delta Q=+0.0201$.} \\
\bottomrule
\end{tabular}
\endgroup
\end{table*}

\begin{table*}[t]
\centering
\caption{Earlier SpeCa controller development sensitivity on 96 prompts per configuration. Speed $U$ uses the native SpeCa reference of each development study; the quality columns are absolute means, and IR denotes ImageReward. These prompts are separate from the benchmark of 1,400 prompts.}
\label{tab:plugin-development}
\begingroup
\setlength{\tabcolsep}{3pt}
\renewcommand{\arraystretch}{1.06}
\scriptsize
\begin{tabular}{lrrrrr}
\toprule
Configuration & $U\uparrow$ & CLIP$\uparrow$ & IR$\uparrow$ & HPS$\uparrow$ & Aesthetic$\uparrow$ \\
\midrule
\multicolumn{6}{l}{\textbf{Initial sensitivity scan}} \\
Adaptive, budget 1, minimum gain 0 & 0.344 & 0.6631 & 1.1143 & 0.3078 & 5.8789 \\
Original adaptive controller & 0.348 & 0.6628 & 1.0944 & 0.3078 & 5.8727 \\
Adaptive, maximum horizon 2 & 0.338 & 0.6634 & 1.1170 & 0.3080 & 5.8691 \\
Adaptive, risk 1, budget 0.5, horizon 4 & 0.344 & 0.6631 & 1.1155 & 0.3078 & 5.8799 \\
Adaptive, risk 1, budget 1 & 0.344 & 0.6631 & 1.1145 & 0.3078 & 5.8757 \\
Guarded floor, threshold 1.30, budget 0.25 & 1.053 & 0.6606 & 1.0973 & 0.3042 & 5.8965 \\
Guarded floor, threshold 1.30, budget 1 & 1.123 & 0.6592 & 1.0701 & 0.3045 & 5.8688 \\
Guarded refresh, threshold 1.15, budget 1 & 1.035 & 0.6629 & 1.0283 & 0.3056 & 5.8743 \\
Guarded refresh, residual cap 1.10 & 1.018 & 0.6657 & 1.0614 & 0.3063 & 5.8724 \\
Guarded refresh, risk 0.35 & 0.962 & 0.6595 & 1.0949 & 0.3069 & 5.9118 \\
Guarded refresh, risk 0.65 & 0.956 & 0.6595 & 1.0949 & 0.3069 & 5.9118 \\
Guarded refresh, risk 0.85 & 0.977 & 0.6595 & 1.0949 & 0.3069 & 5.9118 \\
Guarded refresh, threshold 1.30, budget 1 & 1.110 & 0.6627 & 1.1284 & 0.3042 & 5.8887 \\
Native cap extension, horizon 16 & 0.922 & 0.6573 & 1.0893 & 0.3078 & 5.8880 \\
Native floor, horizon 16, threshold 1.15 & 1.054 & 0.6577 & 1.0041 & 0.3050 & 5.8545 \\
Native floor, horizon 16, threshold 1.30 & 1.123 & 0.6602 & 1.0969 & 0.3040 & 5.8988 \\
Native floor, horizon 16 & 1.014 & 0.6576 & 1.0859 & 0.3069 & 5.8799 \\
\midrule
\multicolumn{6}{l}{\textbf{Prefix guard development}} \\
Prefix guard: first 25 steps, risk 0.93 & 1.013 & 0.6594 & 1.1142 & 0.3069 & 5.8906 \\
\midrule
\multicolumn{6}{l}{\textbf{Early relaxation development}} \\
Early relaxation: threshold 1.30, risk 0.93 & 0.963 & 0.6595 & 1.0949 & 0.3069 & 5.9118 \\
Early relaxation: threshold 1.50, risk 0.93 & 0.964 & 0.6588 & 1.0924 & 0.3054 & 5.9069 \\
Early relaxation: threshold 1.50, no refresh & 1.012 & 0.6576 & 1.0859 & 0.3069 & 5.8799 \\
\midrule
\multicolumn{6}{l}{\textbf{No refresh development}} \\
Early relaxation: threshold 1.30, no refresh & 1.026 & 0.6600 & 1.0801 & 0.3080 & 5.9007 \\
\midrule
\multicolumn{6}{l}{\textbf{Risk threshold development}} \\
Guarded refresh: threshold 1.30, risk 0.93 & 0.964 & 0.6595 & 1.0949 & 0.3069 & 5.9118 \\
\bottomrule
\end{tabular}
\endgroup
\end{table*}

\subsection{Bounded Autoregressive Evaluation}
\label{sec:diagnostics}
\label{app:diagnostic-details}

The Lumina-mGPT-7B-768~\citep{liu2024lumina} diagnostics use BF16, greedy
decoding, and the target's native CFG, tokenizer, and decoder. Four target-only parity
cases have zero mismatches. The continuation tests fix one prompt and seed,
horizons two and four, and 64 of the target's 2,304 visual units; every
comparison fixes the target identity and the measurement boundary.

\paragraph{Packed numerical paths can change token decisions.}
The packed verifier diverged at row zero for both horizons, whereas the
reference that processes one row at a time matched. Switching to the math
attention backend did not remove the mismatch. With a short prefix, the earliest difference
occurred in the MLP gate projection of the first layer, consistent with
rounding that depends on tensor shape. Computing MLP projections row by row recovered token
agreement for 2/2 and 4/4 tested rows, with exact checkpoint restore. After
two excluded warmup pairs, the repaired target forward pass had ratios of
median sequential to median packed execution time of $1.610087\times$ and
$2.783761\times$ at horizons two and four, and was faster in 11/11 measured pairs per
horizon. These ratios measure the target forward pass, not end-to-end
generation.

\paragraph{Continuation state diagnostics.}
A repaired SJD transaction accepted three draft units, discarded one, and
committed four units that matched clean sequential replay in raw tokens,
token types, and the cached next-token identity. A longer continuation
nevertheless diverged at transaction index eight (counting from zero), after 33 of 64
visual units had matched. Restoring token agreement did not restore the full
continuation state: CFG/KV caches, masks, positions, and the ownership of
corrections and suffixes must also agree at each authoritative boundary.

\paragraph{Sequential authority restores bounded exactness at a cost.}
The conservative variant executes both target verification and
authoritative replay through canonical transitions that process one row at
a time. Independent speculative and clean sessions then matched all 64
visual units at both horizons, including raw token identities and types,
transaction and runtime state, cached tokens and logits for the next target
step, checkpoint/restore, and session isolation. The tested transactions
included accepted prefixes, corrections,
and structural boundary caps.

\begin{table*}[!tp]
\centering
\small
\begin{tabular}{@{}rrrr@{}}
\toprule
Horizon & Target-only p50 (ms) & Sequential speculative p50 (ms) & Median paired ratio \\
\midrule
2 & 19,320.727 & 21,509.036 & 0.8983 \\
4 & 19,313.023 & 23,767.021 & 0.8126 \\
\bottomrule
\end{tabular}
\caption{Paired timing of the generation loop over 64 units for the variant with
sequential authority. Each horizon uses two warmup pairs and three measured pairs in
alternating execution order, with synchronized timing boundaries. The ratio
is the median of target-only/speculative time ratios across pairs. Values below one indicate slower speculative
execution. All declared exactness checks passed.}
\label{tab:diagnostics}
\end{table*}

Table~\ref{tab:diagnostics} shows that this exact variant was slower than
target-only decoding at both horizons. Its target path is sequential,
whereas the forward pass timed above uses the repaired packed path.

\subsection{Evaluation under an RGB Deviation Budget}
\label{app:budget-diagnostics}

In a separate diagnostic on FLUX.1-dev~\citep{flux2024} at $1024^2$ with 28
target steps, we evaluate the mechanism under a budget on the RGB deviation of
each output from its target-only image. Four target-only parity
cases have zero latent or scheduler mismatches over all 28 steps.
Split conformal calibration used 56 fit prompts, 56 disjoint calibration
prompts, and 84 holdout prompts. The causal proxy achieved a 95\% lower
confidence bound of 0.8468 on its Spearman correlation, and its calibrated
bound covered the realized RGB deviation at all four horizons (1, 2, 4, and
8) on 84/84 holdout prompts.

These diagnostic prompts are rejected construction candidates that share no
prompts or scenes with the 1,400 benchmark prompts but share upstream creative
references with them. The fit, calibration, and holdout sets partition this
diagnostic pool.

The initial adaptive pilot proposed no speculative work. After inspecting
that result, we ran an exploratory study of fallback behavior. In that study,
the adaptive policy selected target-only decoding in all 28 cases, whereas a forced
aggressive proposal reduced runtime but exceeded the budget in all 28. A
separate study with fixed horizons yielded median paired speedups of
$1.826\times$, $2.280\times$, and $2.724\times$ in end-to-end time over
target-only decoding for horizons one, two, and four, respectively, and 0/14
cases met the budget at each horizon.
These results support treating the budget in Equation~\ref{eq:diff-feasible}
as a hard constraint.

\section{Layered System Design}
\label{app:layered-system-design}

\begin{figure*}[!tp]
\centering
\begin{tikzpicture}[
  x=1mm,
  y=1mm,
  every node/.style={font=\small},
  box/.style={draw=T2IDark, rounded corners=2pt, thick, align=center,
    minimum height=9mm, inner sep=3pt},
  shared/.style={box, fill=T2IBlue!10},
  control/.style={box, fill=T2IGold!16},
  iface/.style={box, fill=T2IGray},
  ar/.style={box, fill=T2ITeal!14},
  diff/.style={box, fill=T2IRed!9},
  evidence/.style={box, fill=T2IGray, minimum height=8mm},
  layer/.style={font=\small\bfseries, text=T2IDark, anchor=east},
  dep/.style={-{Latex[length=2mm]}, thick, draw=T2IDark},
  receipt/.style={-{Latex[length=2mm]}, dashed, thick, draw=T2IBlue},
  feedback/.style={-{Latex[length=2mm]}, dashed, thick, draw=T2IGold!80!black}
]

\node[layer] at (13,0) {L1};
\node[shared, text width=130mm] (plan) at (87,0)
  {\textbf{Frozen evaluation contract}\\
   benchmark population \quad | \quad paired prompts and seeds \quad | \quad
   target and measurement scope};

\node[layer] at (13,-15) {L2};
\node[shared, text width=60mm] (bound) at (52,-15)
  {\textbf{Reproducible run binding}\\
   sample, policy, target identity, and physical attempt};
\node[shared, text width=60mm] (binding) at (122,-15)
  {\textbf{Immutable trial record}\\
   seed, permitted scope, timing boundary, and attempt identity};

\node[layer] at (13,-32) {L3};
\node[control, text width=33mm] (neg) at (38,-32)
  {\textbf{Prompt routing}\\
   predicted topic and optional prompt hash};
\node[control, text width=37mm] (runtime) at (86,-32)
  {\textbf{Backend policy}\\
   native execution or a permitted intervention};
\node[control, text width=33mm] (router) at (135,-32)
  {\textbf{Optional online control}\\
   progress, cost, and causal guards};

\node[layer] at (13,-49) {L4};
\node[iface, text width=130mm] (roles) at (87,-49)
  {\textbf{Native execution interface}\\
   token verification \quad | \quad cache reuse or refresh
   \quad | \quad native fallback and measured feedback};

\node[layer] at (13,-67) {L5};
\node[ar, text width=60mm, minimum height=13mm] (arcontract) at (52,-67)
  {\textbf{Autoregressive acceptance}\\
   configured native verification, accepted prefix, and cache updates};
\node[diff, text width=60mm, minimum height=13mm] (diffcontract) at (122,-67)
  {\textbf{Diffusion/flow intervention guards}\\
   residual or cache age, partial refresh, and proxy budget};

\node[layer] at (13,-87) {L6};
\node[ar, text width=60mm, minimum height=14mm] (lumina) at (52,-87)
  {\textbf{CSD + LANTERN + SJD}\\
   speculative window, tree depth, or block length\\
   Janus, LlamaGen, Lumina, and Anole};
\node[diff, text width=60mm, minimum height=14mm] (flux) at (122,-87)
  {\textbf{SpeCa + TaylorSeer}\\
   threshold control or partial block refresh\\
   FLUX.1-dev and FLUX.1-schnell};

\node[layer] at (13,-108) {L7};
\node[evidence, text width=31mm] (telemetry) at (32,-108)
  {run telemetry};
\node[evidence, text width=24mm] (result) at (67,-108)
  {paired result};
\node[evidence, text width=25mm] (trace) at (101,-108)
  {state and output evidence};
\node[evidence, text width=34mm] (envelope) at (142,-108)
  {immutable result archive};
\node[font=\small, text=T2IDark, text width=140mm, align=center]
  at (87,-117)
  {Measurement and evidence layer};

\draw[dep] (plan) -- (bound);
\draw[dep] (plan) -- (binding);
\draw[dep] (bound) -- (neg);
\draw[dep] (binding) -- (runtime);
\draw[dep] (neg) -- (runtime);
\draw[dep] (runtime) -- (router);
\draw[dep] (runtime) -- (roles);
\draw[dep] (router.south) -- (router.south |- roles.north);
\draw[dep] (roles) -- (arcontract);
\draw[dep] (roles) -- (diffcontract);
\draw[dep] (arcontract) -- (lumina);
\draw[dep] (diffcontract) -- (flux);

\draw[receipt] ([xshift=5mm]roles.north) --
  node[right, font=\small] {execution feedback}
  ([xshift=5mm]runtime.south);
\draw[dep] (lumina) -- (telemetry);
\draw[dep] (flux) -- (telemetry);
\draw[dep] (telemetry) -- (result);
\draw[dep] (result) -- (trace);
\draw[dep] (trace) -- (envelope);
\draw[feedback] (telemetry.north) -- (32,-100) -- (165,-100) -- (165,-32) --
  (router.east);
\node[font=\small, rotate=90, fill=white, inner sep=1pt, text=T2IDark]
  at (165,-70) {cost / progress / reuse};

\begin{scope}[on background layer]
  \node[draw=T2IBlue!55, dashed, rounded corners=4pt,
    fit=(plan)(bound)(binding)(neg)(runtime)(router)(roles), inner sep=4pt] {};
\end{scope}
\node[anchor=west, font=\small, text=T2IBlue, fill=white, inner sep=1.5pt]
  at (16,-57) {shared design across models};

\end{tikzpicture}
\caption{Layered evaluation and \topicRoute plugin design.
Prompt routing selects backend policies, with optional online execution
control. The evidence layer records execution and returns feedback local
to each sample. Appendix~\ref{sec:method} specifies the canonical
transactions on tokens and state.}
\label{fig:layered-system-design}
\end{figure*}
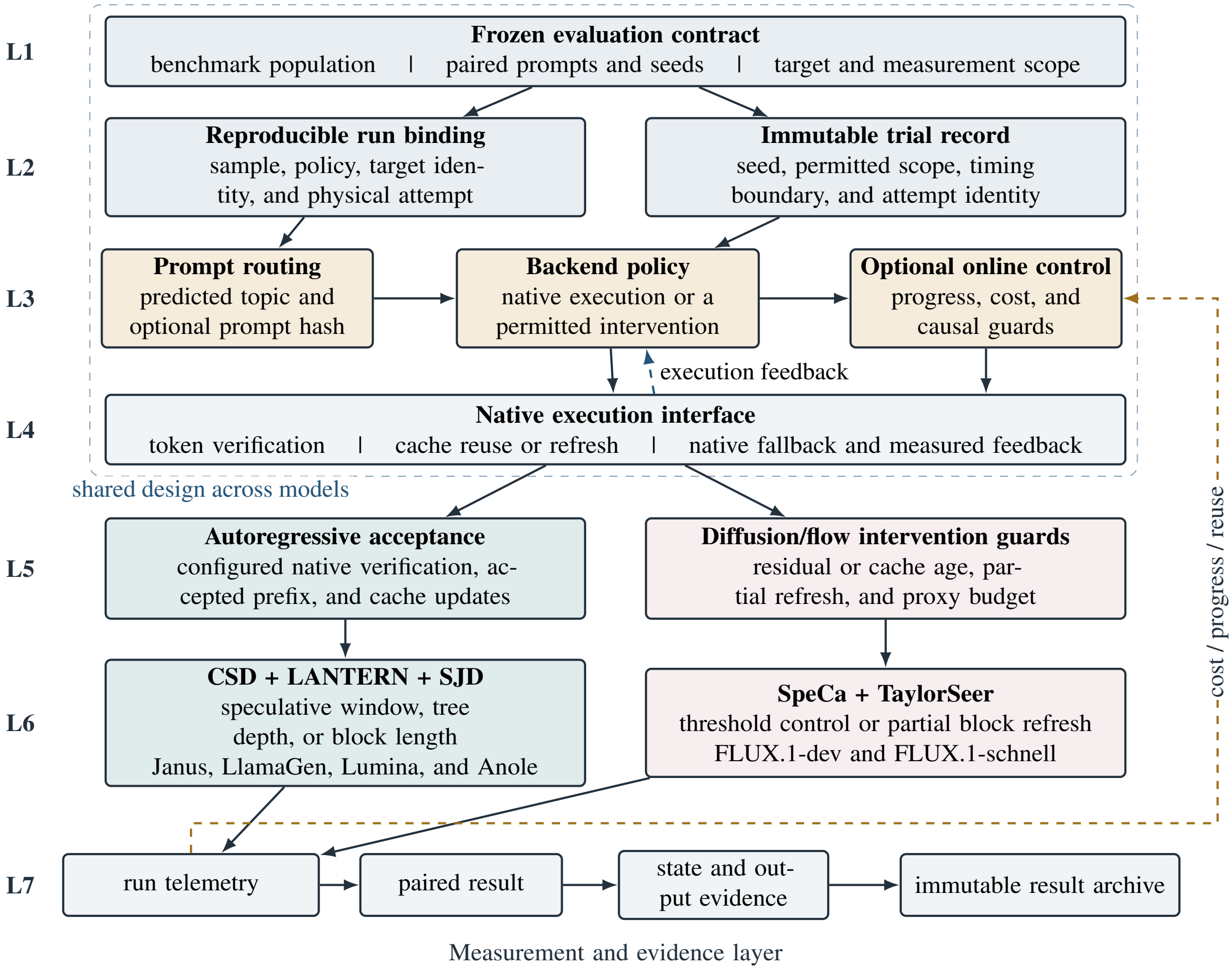

Figure~\ref{fig:layered-system-design} separates evaluation identity,
prompt routing, optional online control, native execution, and measurement.
AR integrations retain their configured verification rules, whereas cache
integrations combine causal proxy guards with native reuse or refresh
operations. Appendix~\ref{sec:diagnostics} evaluates the canonical
transaction of Figure~\ref{fig:transaction} for exact state preservation, and
the experiments on complete images evaluate the backend integrations of
Appendix~\ref{sec:method}.

\section{Discussion and Limitations}
\label{sec:discussion}

\paragraph{Why evaluation by topic matters.}
Comparisons by topic expose quality and latency variation. LANTERN on LlamaGen at
$256^2$ improves alignment and preference scores in all seven topics yet
lowers the criterion score for text and symbols. LANTERN++ at $512^2$ has
lower recorded latency alongside broad reductions in quality scores. These
observations support reporting by topic with complementary quality measures,
and ADT-Tree shows that reducing target work need not yield proportional
gains in request latency.

\paragraph{Dataset scope.}
The 1,400 English prompts balance seven topics and easy/hard construction
labels. The topics follow exploratory tests and source taxonomies without an
independent validation, and the labels do not measure empirically calibrated
model difficulty. The 35
descriptive categories are unequal in size, and five are unused. Shared
scenes and construction references introduce dependence, so the paired
bootstrap resamples groups of prompts that share a scene or creative seed.
The prompt workload does not
include dedicated evaluations of safety, demographic representation,
memorization, or multilingual typography.

\paragraph{Measurement limitations.}
The shared prompts and metric definitions support method comparisons, with
quality changes measured against the stated fixed references.
Appendix~\ref{app:evaluation-protocol} specifies the timing boundaries and
evaluator settings. The core AR intervals quantify pointwise
sampling uncertainty under the stated grouping by scene and creative seed.

\paragraph{Implications for method design.}
The findings motivate reporting quality changes by topic alongside latency
and target work, with each method compared against its stated reference.
The plugin design in Section~\ref{sec:plugin} combines predicted prompt
topics, frozen backend policies, and optional online cost feedback. Four
of six groups meet the group criterion; TaylorSeer trades quality for the
largest speed gain, while Lumina fails to improve Overall latency.
LANTERN passes the quality check despite a small decline in the primary
criterion. The controls differ by backend. Janus uses static window routing,
and the selected SpeCa policy bypasses the online gain screen. In the paired
ablations of the final policies (Appendix~\ref{app:plugin-followup}), the
hash gates of Janus and Anole add neither speed nor balance, and neither do
Anole's proposal lengths for each topic. Janus's topic routing and LANTERN's gain
screen improve balance at some cost in speed. Threshold relaxation produces
SpeCa's gain, and Lumina's components have little effect on its speed. In
separate runs, the variants without LANTERN's terminal guard or TaylorSeer's
single-stream refresh are faster than the final policies, and the variant
without the refresh also has a larger loss in $Q$. Speed gains recur in the reruns and on 200 external
DrawBench prompts. The number of nondecreasing metrics and the pass
decisions change between reruns; of the four groups that pass in
Table~\ref{tab:plugin-main}, only Anole passes again.

\section{Qualitative Examples and Prompts}
\label{app:qualitative-comparison}

\begin{figure*}[!tp]
\centering
\resizebox{\linewidth}{!}{\input{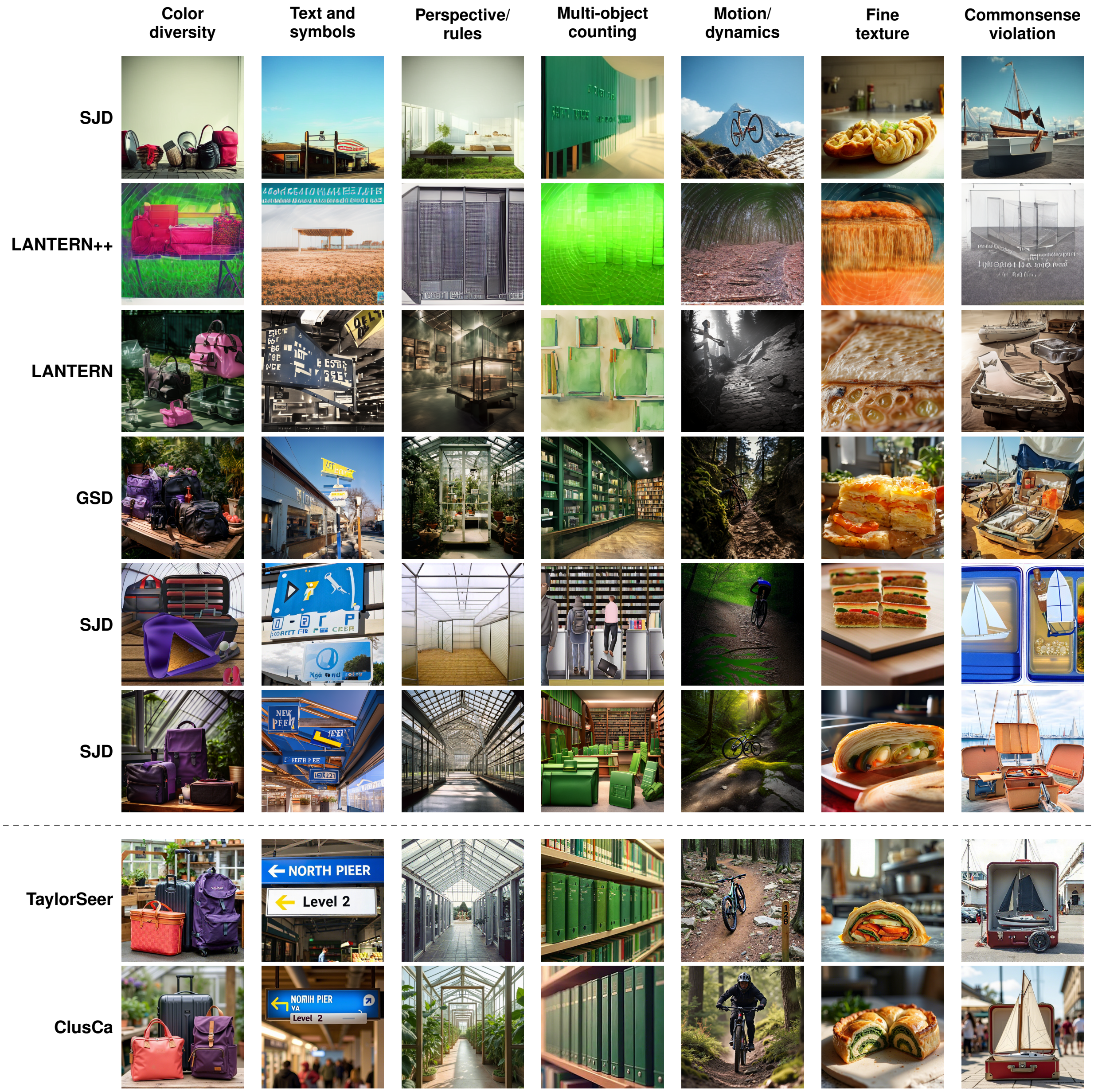}}
\caption{Selected success and failure cases across eight evaluated configurations and seven topics. All images in a column share one prompt. The dashed line separates autoregressive (top) from diffusion/flow (bottom) methods. Full prompts appear in Appendix~\ref{app:qualitative-prompts}.}
\label{fig:topic-comparison-full}
\end{figure*}

Figure~\ref{fig:topic-comparison-full} compares eight configurations on seven
examples, one per topic. All rows share the prompt within each column.
From top to bottom, the three SJD rows correspond to Janus, Anole, and Emu3.
The figure uses original generated images; their evaluation batches are
tracked separately from those used for the tables. LANTERN images
use the strict-512 evaluation, and LANTERN++ images use LlamaGen Stage~I at
$256^2$.

\paragraph{Selection procedure.}
For each backbone group, the grid shows one accelerated configuration;
these rows were fixed before the shared prompts were chosen. Candidate
prompts are screened using the recorded topic scores and criterion
descriptions, and the original images are then inspected against the full
prompts. The seven prompts are listed below.

\paragraph{Reading the examples.}
Several configurations retain recognizable pastry texture and architectural
depth while failing exact text or isolated object counts. TaylorSeer's text
example adds an extra letter in ``NORTH PIEER'', and its bicycle appears
stationary despite the requested descent. ClusCa renders recognizable scenes
in all seven topics yet misspells the sign as ``NOMIH PIER'', fills the
shelf with many more than seven green binders where the prompt asks for seven
separated notebooks, and places a model sailboat in an open suitcase instead
of a full-size sailboat in a lunchbox. LANTERN++ replaces the sailboat scene with empty display boxes and
unreadable text. These failures range from violated constraints to missing
scenes.

\subsection{Prompts for the Examples}
\label{app:qualitative-prompts}

The following prompts correspond, from left to right, to the columns in Figure~\ref{fig:topic-comparison-full}.

\paragraph{Color diversity.}
A coral picnic basket, a charcoal-black travel suitcase, and a violet canvas backpack form a triangular arrangement beside a greenhouse workbench. Realistic photograph. Controlled illumination keeps all colors visually distinct.

\paragraph{Text and symbols.}
At a neighborhood food hall, a blue overhead directional sign reads ``NORTH PIER'', with ``LEVEL 2'' on a smaller white panel and a yellow arrow pointing left. Documentary photograph. No object blocks the important printed or symbolic marks in the scene.

\paragraph{Perspective/rules.}
A modern greenhouse interior contains a sequence of square doorframes converging toward one distant vanishing region, with a display cabinet establishing scale. Documentary photograph. Object boundaries remain readable wherever the subjects overlap.

\paragraph{Multi-object counting.}
Exactly 7 green notebooks are evenly separated along a library display wall, with every item fully visible. Realistic digital painting. Clear gaps make the exact quantity directly verifiable.

\paragraph{Motion/dynamics.}
A mountain bicycle descends a rocky woodland trail, with a sharp leading edge and a directional trail behind it. A fixed trail marker remains sharp at the lower-right edge under low angled light. Realistic photograph. The action remains inside the frame with a visible path of movement.

\paragraph{Fine texture.}
A sliced stuffed vegetable pastry sits inside a working kitchen, revealing flaky crust around distinct vegetable layers. Cool diffuse daylight separates a bright highlight and a softly shaded region. Realistic photograph. Fine material detail remains visible along the nearest edges and broad surfaces.

\paragraph{Commonsense violation.}
A full-size sailboat fits inside an open lunchbox at an outdoor nautical exhibition. An ordinary folding chair and tool case retain familiar dimensions. Realistic photograph. Foreground and background comparisons confirm the same impossible size or containment relation. Contact, perspective, and surrounding architecture make the reversal measurable.

\end{document}